\documentclass[accepted]{uai2026} 
                        
\usepackage[american]{babel}

\usepackage{natbib} 
\usepackage{mathtools} 

\usepackage{amsmath,amsfonts,bm,dsfont,amsthm,nicefrac}

\newcommand{\W}{\mathbb{W}}

\renewcommand{\min}[1]{\underset{#1}{\text{min}}\,}
\renewcommand{\max}[1]{\underset{#1}{\text{max}}\,}
\renewcommand{\inf}[1]{\underset{#1}{\text{inf}}\,}
\renewcommand{\sup}[1]{\underset{#1}{\text{sup}}\,}

\newcommand{\argmin}[1]{\underset{#1}{\text{arg min}}\,}

\newcommand{\arginf}[1]{\underset{#1}{\text{arg inf}}\,}

\newtheorem{theorem}{Theorem}[section]
\newtheorem{proposition}{Proposition}[section]

\newtheorem{lemma}{Lemma}[section]

\newtheorem{definition}{Definition}[section]
\newtheorem{remark}{Remark}[section]

\newcommand{\gradW}{\nabla_{W}}
\usepackage{pifont}
\usepackage{booktabs}
\usepackage{hyperref}
\usepackage{url}
\usepackage[acronym]{glossaries}
\usepackage{nicefrac}
\usepackage{graphicx, subcaption, wrapfig}
\usepackage{amsmath, amsfonts, amsthm, dsfont, nicefrac}
\usepackage[table, dvipsnames]{xcolor}
\usepackage{wrapfig}
\usepackage{algorithm}
\usepackage{algpseudocode}
\usepackage{multirow}
\usepackage{tabularx}

\glsdisablehyper 



\newcommand{\cmark}{\ding{51}}%
\newcommand{\xmark}{\ding{55}}%

\usepackage{tikz}

\newacronym{ot}{OT}{Optimal Transport}
\newacronym{otdd}{OTDD}{OT Dataset Distance}
\newacronym{icnn}{ICNN}{Input Convex Neural Network}
\newacronym{jko}{JKO}{Jordan, Kinderlehrer, Otto}
\newacronym{pl}{PL}{Polyak Łojasiewicz}
\newacronym{wgf}{WGF}{Wasserstein Gradient Flow}
\newacronym{ml}{ML}{Machine Learning}
\newacronym{gmm}{GMM}{Gaussian Mixture Model}
\newacronym{eeg}{EEG}{electroencephalogram}
\newacronym{erm}{ERM}{Empirical Risk Minimization}
\newacronym{npgd}{NPGD}{Noisy Particle Gradient Descent}

\newacronym{ode}{ODE}{Ordinary Differential Equation}
\newacronym{da}{DA}{Domain Adaptation}

\usepackage{placeins}  
\usepackage{stfloats}  

\title{Wasserstein Gradient Flows for Scalable and Regularized Barycenter Computation}

\author[1]{Eduardo Fernandes Montesuma}
\author[1]{Yassir Bendou}
\author[2]{Mike Gartrell\thanks{Work done while at Sigma Nova.}}
\affil[1]{%
    Sigma Nova\\
    Paris, France
}
\affil[2]{%
    Rhizome Labs\\
    Paris, France
}
  
\begin{document}
\maketitle

\begin{abstract}
Wasserstein barycenters provide a principled approach for aggregating probability measures, while preserving the geometry of their ambient space. Existing discrete methods are not because as they assume access to the complete set of samples from the input measures. Meanwhile, neural network approaches do scale well, but rely on complex optimization problems and cannot easily incorporate label information. We address these limitations through gradient flows in the space of probability measures. Through time discretization, we achieve a scalable algorithm that i) relies on mini-batch optimal transport, ii) accepts modular regularization through task-aware functions, and iii) seamlessly integrates supervised information into the ground-cost. We empirically validate our approach on domain adaptation benchmarks that span computer vision, neuroscience, and chemical engineering. Our method establishes a new state-of-the-art Wasserstein barycenter solver, with labeled barycenters consistently outperforming unlabeled ones. Our code at \url{https://github.com/SigmaNova/barycentric-gradient-flows}.
\end{abstract}
\section{Introduction}\label{sec:intro}

Defining the mean or center of a set of probability measures is a fundamental primitive in geometric probability theory~\citep{nielsen2020}. Given $K \in \mathbb{N}$ probability measures $\mathcal{Q} = \{Q_{k}\}_{k=1}^{K}$ on a metric space $(\Omega, d)$, their \emph{Wasserstein barycenter}~\citep{agueh2011barycenters} is given by,
\begin{align}
    P^{\star} = \argmin{P \in \mathcal{P}_{2}(\Omega)}\biggl{\{}  \mathbb{B}_{2}(P) =  \sum_{k=1}^{K}\lambda_{k}\mathbb{W}_{2}(P, Q_{k})^{2}  \biggr{\}},\label{eq:w2_barycenter}
\end{align}
where $\mathcal{P}_{2}(\Omega)$ is the set of measures over $\Omega$ with finite second moments, and $\lambda = (\lambda_{1}, \cdots \lambda_{k})$, $\lambda_{k} \geq 0$, $ \forall k$, $\sum_{k}\lambda_{k} = 1$ is an array of \emph{barycentric coordinates}. Equation~\ref{eq:w2_barycenter} defines a notion of the average of the input measures in $\mathcal{Q}$. The main advantage of these barycenters stems from \gls{ot}~\citep{villani2008optimal}, that is, they \emph{lift} the geometry of the space $(\Omega, d)$ to the space $(\mathcal{P}_{2}(\Omega), \mathbb{W}_{2})$~\citep{kloeckner2010geometric}.

\begin{figure*}[ht]
    \centering
    \begin{subfigure}{0.23\linewidth}
        \includegraphics[width=\linewidth]{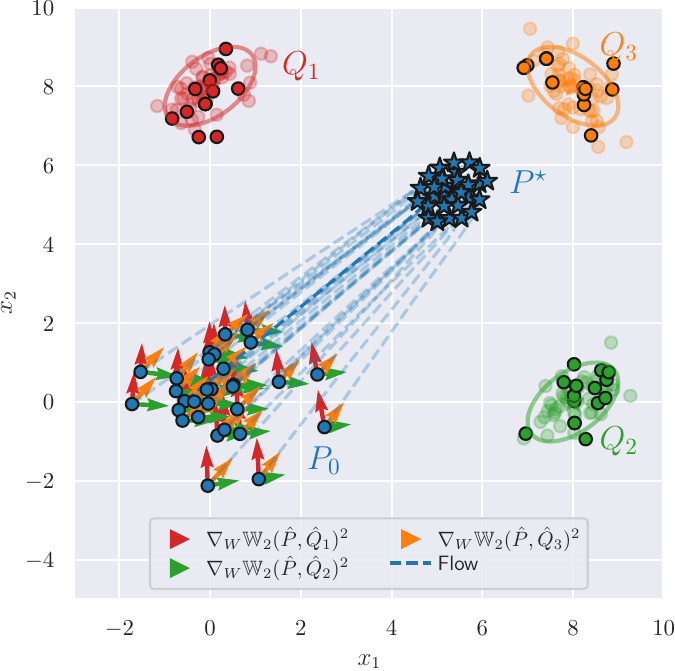}
        \caption{Conceptual Illustration.}
    \end{subfigure}
    \hspace{0.005\linewidth}\vrule\hspace{0.005\linewidth}
    \begin{subfigure}{0.23\linewidth}
        \includegraphics[width=\linewidth]{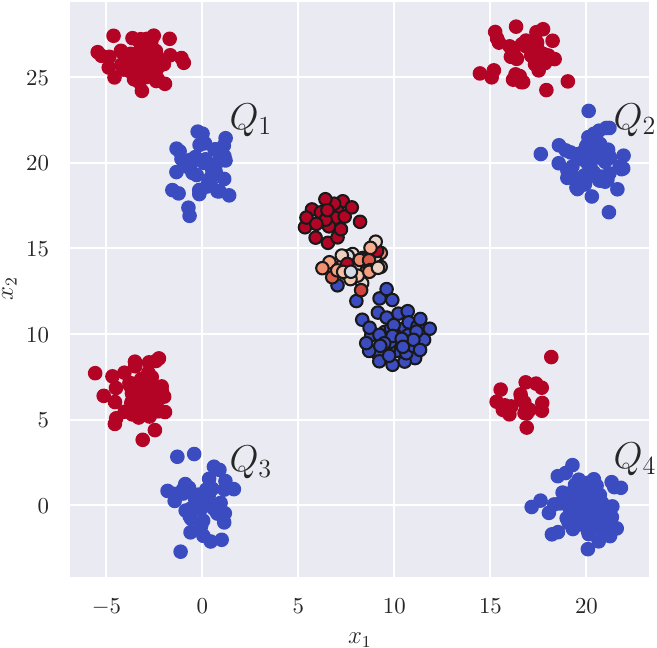}
        \caption{$\mathbb{B}$.}
    \end{subfigure}
    \begin{subfigure}{0.23\linewidth}
        \includegraphics[width=\linewidth]{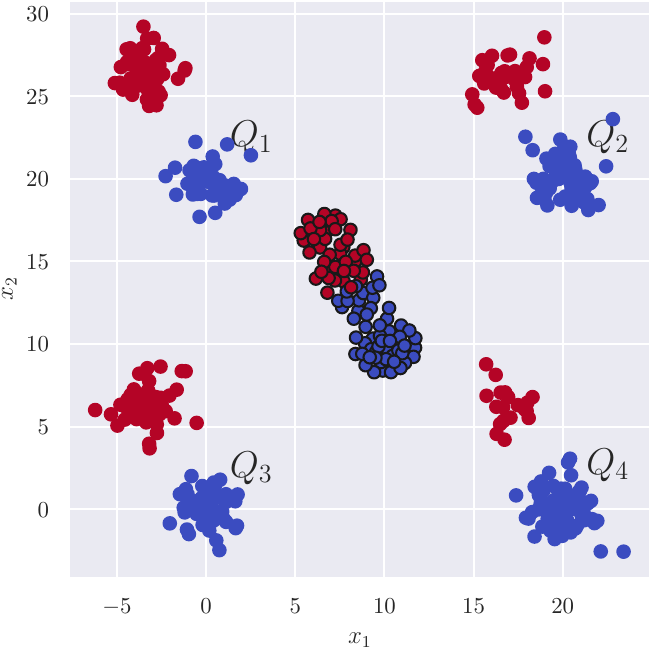}
        \caption{$\mathbb{B} + \mathbb{V}$.}
    \end{subfigure}
    \begin{subfigure}{0.23\linewidth}
        \includegraphics[width=\linewidth]{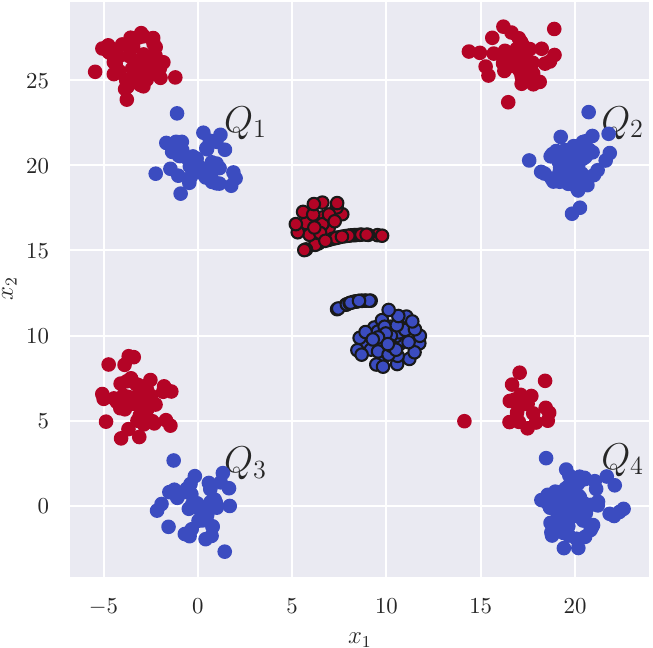}
        \caption{$\mathbb{B} + \mathbb{V} + \mathbb{U}$.}
    \end{subfigure}
    \caption{In (a), we show an illustration of our flow. We show the initial barycenter measure $P_{0}$ which flow towards the solution $P^{\star}$. In red, green, and orange, we show the input measures' mini-batches as solid dots. The flow is given by the combination of gradients $v_{\tau,i} = \sum_{k=1}^{K}\lambda_{k}\nabla\mathbb{W}_{2}(P_{\tau}, Q_{k})$. In parallel, (b) shows a pathological barycenter calculation, with fuzzy labels and unclear class boundaries. Adding regularizers $\mathbb{V}$ (b) and $\mathbb{U}$ (c) progressively resolves this issue.}
    \label{fig:conceptual-illustration}
\end{figure*}

Wasserstein barycenters have wide applications to machine learning, such as posterior aggregation in Bayesian inference~\citep{srivastava2018scalable}, model fusion~\citep{singh2020model}, ensembling in time series forecasting~\citep{coz2023barycenter}, fairness~\citep{gordaliza2019obtaining,visentin2025computing}, and \gls{da}~\citep{montesuma2021wasserstein,montesuma2023multi,montesuma2024lighter}. In each application, the main advantage is geometric fidelity, that is, the Wasserstein barycenter captures and preserves the underlying structure of samples. Figure~\ref{fig:conceptual-illustration} shows a conceptual illustration.

Despite their advantages, Wasserstein barycenters remain challenging beyond the Gaussian setting~\citep{altschuler2022wasserstein}. We have identified three gaps in the current literature. First, the seminal algorithm of~\cite{cuturi2014fast} requires having access to the complete set of input measures' samples at once, making it intractable for large datasets. Neural network methods~\citep{fan2020scalable,korotin2021continuous,kolesov2024estimating,gazdieva2024robust} alleviate this issue by operating on mini-batches, but parametrize the barycentric measure with $\mathcal{O}(K)$ neural networks, coupling model complexity with the number of input measures. Second, neural methods cannot seamlessly incorporate label information into their ground-cost, limiting their performance in supervised tasks (cf. Table~\ref{tab:results-msda}). Third, the objective function in Equation~\ref{eq:w2_barycenter} only accounts for distributional fit. In practice, the barycentric measure must satisfy additional structural properties, such as class separation. Current \textbf{barycenter solvers} do not offer a principled way of enforcing these properties.

Recent methods study the barycenter problem through the gradient flow perspective. First,~\cite{chewi2020gradient} established a gradient flow algorithm on the Bures-Wasserstein manifold, i.e., on Gaussian measures. Second,~\cite{chizat2025doubly} describes the \emph{doubly regularized} barycenter problem through the same lens. Here, in addition to \emph{entropic regularization} at the level of \gls{ot}, they proposed an \emph{outer regularization} that penalizes the entropy of the barycentric measure, which leads to a \gls{npgd} algorithm for empirical measures. Although this algorithm incorporates outer regularization, it has the same limitation as~\cite{cuturi2014fast}, as it assumes \emph{full-batch} access to the samples of input measures.

We address the aforementioned challenges through the lens of \emph{gradient flows} in the space of probability measures~\citep{ambrosio2008gradient, santambrogio2017euclidean}. We therefore conceptualize the barycenter problem as the flow from an initial measure $P_0 = \mathcal{N}(0,\text{Id})$ following the Wasserstein gradient of the functional,
\begin{align}
    \mathbb{F}(P) = \mathbb{B}(P) + \mathbb{R}(P),\label{eq:functional}
\end{align}
where $\mathbb{B}$ denotes the barycenter functional (cf. Equation~\ref{eq:w2_barycenter}), and $\mathbb{R}$ denotes additional regularizing functionals (see Sections~\ref{sec:gfs} and~\ref{sec:functionals}, and Appendix~\ref{appx:gradients}).

Following the established gradient flow literature~\citep{santambrogio2017euclidean}, we decompose $\mathbb{R}$ into internal, potential, and interaction energies, denoted $\mathbb{G}$, $\mathbb{V}$, and $\mathbb{U}$, respectively. This decomposition generalizes the double regularization strategy of~\cite{chizat2025doubly}. Indeed, setting $\mathbb{V} = \mathbb{U} = 0$ and $\mathbb{G}(P) = \eta_{G} H(P)$, where $H(P)$ denotes the entropy of $P$, corresponds to their exact framework. Similarly,~\cite{alvarez2021dataset} devised an approach for flowing a single dataset towards a known target. Our setting differs structurally from that. First, we go beyond the single measure objective, optimizing the \emph{multi-measure} barycentric objective (cf. Equation~\ref{eq:w2_barycenter}). Second, while~\cite{alvarez2021dataset} \emph{transforms} an existing dataset towards a known target, ours \emph{synthesizes} a measure (i.e., the barycentric measure) from noise.

Our contributions are as follows. (i) \textbf{A mini-batch, time discretized gradient flow algorithm for regularized empirical Wasserstein barycenters (Algorithm~\ref{alg:empirical_flow}).} By randomly sampling the input measures, we reduce the computational complexity of traditional barycenter methods~\citep{cuturi2014fast,montesuma2023multi,chizat2025doubly} achieving $2\times$ to $50\times$ speedups with respect to discrete solvers (Figure~\ref{fig:running-time-analysis}). (ii) \textbf{Modular, task-aware regularizing functionals for Wasserstein barycenters (Section~\ref{sec:functionals}).} Following previous gradient flow literature~\citep{santambrogio2017euclidean,alvarez2021dataset}, we decompose $\mathbb{R}$ into internal, potential, and interaction energies, allowing incorporation of plug-and-play regularizers into the barycenter problem. This goes beyond~\cite{chizat2025doubly}, who considered a particular internal energy as an outer regularization strategy. (iii) \textbf{Extensive \gls{da} benchmarking (Section~\ref{sec:experiments}).} We empirically show that our algorithm produces barycenters that respect class structure, by incorporating both appropriate functionals (Section~\ref{sec:functionals}), and labels into the ground-cost of optimal transport (Section~\ref{sec:joint-measures}). On five \gls{da} benchmarks, spanning computer vision, neuroscience, and chemical engineering, we establish a comparison between empirical and neural barycenter solvers, showing that structured costs are essential for \gls{da} performance.

\textbf{Paper organization.} Section~\ref{sec:background} reviews \gls{ot}, Wasserstein barycenters and gradient flows. Section~\ref{sec:wgfs} develops our empirical, time-discretized gradient flow framework and algorithms. Section~\ref{sec:experiments} presents our experiments. Finally, Section~\ref{sec:conclusion} concludes this paper.
\section{Background}\label{sec:background}

Throughout this paper $(\Omega, d)$ is a Polish metric space, and its elements are denoted $z \in \Omega$. We denote $\mathcal{P}_{2}(\Omega)$ as the set of measures with finite second moments~\cite[Definition 6.4]{villani2008optimal}. We work with empirical measures,
\begin{align}
\text{Emp}_{n}(\Omega) = \biggr\{ \hat{P} \in \mathcal{P}_{2}(\Omega): \hat{P} = \frac{1}{n}\sum_{i=1}^{n}\delta_{z_{i}^{(P)}} \biggr\}, \label{eq:approximations}
\end{align}
where $\delta_{z_0}(z) = \delta(z-z_0)$ denotes the Dirac measure centered at $z_0 \in \Omega$. Given a measure $P \in \mathcal{P}_{2}(\Omega)$, we denote by $\hat{P} \in \text{Emp}_{n}(\Omega)$ its empirical approximation based on i.i.d. samples $z_{1}^{(P)},\cdots,z_{n}^{(P)} \sim P$. The superscript $\cdot^{(P)}$ indicates the measure from which the samples originate.

\subsection{Optimal Transport}\label{sec:ot}

\gls{ot} is a field of mathematics concerned with mass transportation at least effort~\citep{villani2008optimal,peyre2019computational,montesuma2024recent}. It was originally founded by~\cite{monge1781memoire}, and seeks a mapping such that,
\begin{align}
    T_{P\rightarrow Q}^{\star} = \arginf{T_{\sharp}P = Q} \mathbb{E}_{z\sim P}[d(z,T(z))^{2}].\label{eq:monge}
\end{align}
$T_{P\rightarrow Q}^{\star}$ is an \gls{ot} mapping between $P$ and $Q$, and $T_{\sharp}$ is the pushforward mapping $P$ to $Q$. Alternatively,~\cite{kantorovich1942transfer} proposed a formulation in terms of a transport plan $\gamma \in \Gamma(P, Q)$, where $\Gamma(P, Q)$ is the set of all joint measures with marginals $P$ and $Q$. In this case,
\begin{align}
    \gamma^{\star} = \arginf{\gamma \in \Gamma(P, Q)}\mathbb{E}_{(z,z')\sim\gamma}[d(z,z')^{2}],\label{eq:ot-kantorovich}
\end{align}
where $\gamma^{\star}$ denotes an \gls{ot} plan. \gls{ot} defines the so-called $2-$Wasserstein distance, through the infimum value of the optimization problem in Equations~\ref{eq:monge} and~\ref{eq:ot-kantorovich},
\begin{equation}
    \W_{2}(P, Q)^{2} = \inf{\gamma \in \Gamma(P, Q)}\mathbb{E}_{(z,z')\sim \gamma}[d(z,z')^{2}].\label{eq:w2}
\end{equation}
Equation~\ref{eq:w2} defines a metric in $\mathcal{P}_{2}(\Omega)$ which bridges the notions of Euclidean (i.e., center of mass) and Wasserstein barycenters. An alternative consists of regularizing the \gls{ot} problem with the entropy of $\gamma$, resulting in the Sinkhorn divergence~\cite{cuturi2013sinkhorn},
\begin{align}
    \mathbb{W}_{2,\epsilon}(P, Q)^{2} = \inf{\gamma \in \Gamma(P, Q)}\mathbb{E}[d(z,z')^{2}] + \lambda H(\gamma),\label{eq:sinkhorn}
\end{align}
where $H(\gamma) = \text{KL}(\gamma|P\otimes Q)$, for $(P\otimes Q)(z) = P(z)Q(z)$, and $\text{KL}$ is the Kullback Leibler divergence. Equation~\ref{eq:sinkhorn} enjoys many desirable properties. First, by relying on the matrix scaling algorithm of~\cite{sinkhorn1964relationship}, it has complexity $\mathcal{O}(n^{2})$ \emph{per iteration}. Second, it can be vectorized on modern GPUs, leading to dramatic speed ups. Third, given that the objective in Equation~\ref{eq:sinkhorn} is strictly convex with respect $\gamma$, the \gls{ot} plan $\gamma_{\epsilon}^{\star}$ is unique.

\subsection{Wasserstein Barycenter}

In the metric setting, the barycenter problem is known as~\cite{frechet1948elements} or~\cite{karcher1977riemannian} means. In our case, given a finite family of measures $\mathcal{Q} = \{Q_{k}\}_{k=1}^{K}$ and a set of barycentric coordinates $\lambda \in \Delta_{K} = \{a \in \mathbb{R}^{K}_{\geq 0}:\sum_{k}a_{k}=1\}$, we define the Wasserstein barycenter over $(\mathcal{P}_{2}(\Omega), \mathbb{W}_{2})$ through the following optimization problem,
\begin{align}
    \hat{P}^{\star} \in \argmin{\hat{P} \in \text{Emp}_{n}(\Omega)}  \underbrace{\sum_{k=1}^{K}\lambda_{k}\mathbb{W}_{2,\epsilon}(\hat{P}, \hat{Q}_{k})^{2}}_{\mathbb{B}_{2,\epsilon}(\hat{P}|\mathcal{Q})}.\label{eq:wasserstein_barycenter}
\end{align}
Here, $\mathbb{B}_{2,\epsilon}(\cdot|\mathcal{Q})$ is a functional mapping $\text{Emp}_{n}(\Omega)$ to $\mathbb{R}_{\geq 0}$, i.e., the set of non-negative real numbers. Henceforth, we denote $\mathbb{B}_{\epsilon}(\hat{P})$ for simplicity. We provide a further description of barycenter solvers in Appendix~\ref{appx:survey}.

\noindent\textbf{Existence and Uniqueness of Barycenters.} The problem in Equation~\ref{eq:wasserstein_barycenter} is a minimization over empirical measures. As such, its objective has 3 levels that must be separated. First, each $\mathbb{W}_{2,\epsilon}$ has an associated transport plan, $\gamma_{\epsilon}^{\star}$, which is unique by the strict convexity of entropic \gls{ot} (c.f. Section~\ref{sec:ot}). Second, the continuous barycenter (i.e., the minimizer of Equation~\ref{eq:w2_barycenter}), denoted $P^{\star} \in \mathcal{P}_{2}(\Omega)$, always exists on $\Omega = B(0, R)$ by the direct method of calculus of variations~\citep[Box 1.1]{santambrogio2015optimal}, and~\cite{agueh2011barycenters} shows that uniqueness holds when at least one input measure vanishes on small sets. Third, the empirical barycenter $\hat{P}^{\star} \in \text{Emp}_{n}(\Omega)$ exists, since the objective is continuous on the compact $\Omega^{n}$. However, the \textbf{empirical barycenter} is not unique, as Problem~\ref{eq:wasserstein_barycenter} is not convex.

\subsection{Gradient Flows}\label{sec:gfs}

Our method relies on the gradient flow interpretation of the minimization in Equation~\ref{eq:wasserstein_barycenter}. Starting from the Euclidean setting, let $F:\mathbb{R}^{d}\rightarrow\mathbb{R}$ be a functional. A gradient flow is the solution to the \gls{ode},
\begin{equation}
    \dot{x}(t) = -\nabla F(x(t))\text{, subject to }x(0) = x_0.\label{eq:EVI}
\end{equation}
When discretized with the \emph{forward Euler discretization}, one has gradient descent,
\begin{align*}
    \dot{x}(t) = \frac{x_{t+1} - x_{t}}{\alpha} \implies x_{t+1} = x_{t} - \alpha\nabla F(x_{t})
\end{align*}
We now extend this framework to probability measures. Let $\mathbb{F}:\mathcal{P}_{2}(\Omega)\rightarrow\mathbb{R}$ be a functional. The evolution of a curve $\{ P_{t} \}_{t\geq 0}$ in $\mathcal{P}_{2}(\Omega)$ is given by the continuity equation
\begin{equation}
    \partial_{t} P_{t} = -\text{div}(P_{t}v_{t}),\label{eq:continuity}
\end{equation}
where $v_{t}:\Omega\rightarrow\Omega$ is a velocity field, and $\text{div}$ is the divergence operator, understood in a measure theoretic sense. Choosing $v_{t} = -\gradW \mathbb{F}(P_{t})$ yields the gradient flow of $\mathbb{F}$, that is, the measure theoretic analogue of Equation~\ref{eq:EVI}. Here, $\gradW$ denotes the Wasserstein gradient of the functional $\mathbb{F}$ (c.f.~\cite[Section 5.4]{chewi2024statistical} and Appendix~\ref{appx:calculus}).

The structure of $\mathbb{F}$ determines the dynamics of $P_{t}$. Following the classical \gls{ot} theory~\citep[Chapter 7]{santambrogio2015optimal}, we consider the functional $\mathbb{F}(P) = \mathbb{B}(P) + \mathbb{R}(P)$, where,
\begin{equation}
\begin{aligned}
    \mathbb{R}(P) = &\underbrace{\int G(p(z))dz}_{\text{Internal energy }\mathbb{G}(P)} + \underbrace{\int V(z)dP(z)}_{\text{Potential energy }\mathbb{V}(P)} +\\ &\underbrace{\int\int U(z,z')dP(z)dP(z')}_{\text{Interaction energy }\mathbb{U}(P)}
\end{aligned}\label{eq:functional-measure}
\end{equation}
where $p:\Omega\rightarrow\mathbb{R}$ denotes the density of $P$, $G:\mathbb{R}\rightarrow\mathbb{R}$, $V:\Omega\rightarrow\mathbb{R}$, and $U:\Omega\times\Omega\rightarrow\mathbb{R}$. For example, when $G(s) = s\log s$, $\mathbb{G}$ is the \emph{entropy functional} of~\citep{chizat2025doubly}, associated with diffusion dynamics. $\mathbb{G}$, $\mathbb{V}$, and $\mathbb{U}$ introduce diffusion, drift, and pairwise behaviors in the flow, respectively. See Section~\ref{sec:functionals} for further examples.
\section{Wasserstein Barycenters as Gradient Flows}\label{sec:wgfs}

In this section, we describe a new method for computing empirical \gls{ot} barycenters. We use, $\mathbb{F}(\hat{P}) = \mathbb{B}_{\epsilon}(\hat{P}) + \mathbb{R}(\hat{P})$, where $\mathbb{B}_{\epsilon}$ is the barycenter objective defined in Equation~\ref{eq:wasserstein_barycenter}, and $\mathbb{R}$ is a combination of different energies defined in Equation~\ref{eq:functional-measure}. We implement a flow from a prior measure $P_{0} = \mathcal{N}(0, \text{Id})$ to an empirical minimizer of $\mathbb{F}(\hat{P})$ through a time-discretized version of the continuity equation. We proceed in two steps. \textbf{First}, we derive the empirical version of the continuity equation (c.f., Equation~\ref{eq:continuity}). For $\hat{P}_{t} $,
\begin{align}
    \partial_{t} \hat{P}_{t} &= \frac{1}{n}\sum_{i=1}^{n}\partial_{t}\delta_{z_{t,i}^{(P)}} = -\dfrac{1}{n}\sum_{i=1}^{n}\text{div}(\dot{z}_{t,i}^{(P)}\delta_{z_{t,i}^{(P)}}),\label{eq:continuity_eq_discrete}\\
    &= -\text{div}(\hat{P}_{t}v_{t}),\nonumber
\end{align}
where we identified $v_{t}(z_{t,i}^{(P)}) = \dot{z}_{t,i}^{(P)}$, i.e., the velocity field displacing samples of $\hat{P}_{t}$. \textbf{Second}, we discretize the gradient flow in time, through an index $\tau = \alpha t$, $\alpha > 0$. Again, using the forward Euler scheme,
\begin{equation}
    \dot{z}_{t,i}^{(P)} = \frac{z_{\tau+1,i}^{(P)} -z_{\tau,i}^{(P)}}{\alpha}\implies  z_{\tau+1,i}^{(P)} = z_{\tau,i}^{(P)} + \alpha v_{\tau,i},\label{eq:discrete_gradient_flow}
\end{equation}
where $v_{\tau,i} = v_{\tau}(z_{\tau,i}^{(P)})$. We can obtain different update strategies based on the velocity field,
\begin{align}
    v_{\tau,i} &= - \nabla_{z_{\tau,i}^{(P)}}\mathbb{F}(\hat{P}_{\tau}),\label{eq:standrd}\\
    v_{\tau,i} &= - \nabla_{z_{\tau,i}^{(P)}}\mathbb{F}(\hat{P}_{\tau}) + \sqrt{2\alpha\eta_{G}}\xi_{\tau,i},\label{eq:langevin}\\
    v_{\tau,i} &= u_{\tau,i}, \quad  u_{\tau+1,i} = \beta u_{\tau,i} - \alpha\nabla_{z_{\tau,i}^{(P)}}\mathbb{F}(\hat{P}_{\tau}).\label{eq:momentum}
\end{align}
where $\xi_{\tau,i}\sim\mathcal{N}(0,\text{Id})$. Equations~\ref{eq:standrd},~\ref{eq:langevin}, and~\ref{eq:momentum} correspond to the steepest descent, Langevin dynamics, and momentum descent strategies, respectively. $\eta_{G} \geq 0$ denotes the diffusion coefficient. The Langevin dynamics version of the flow corresponds to the \gls{npgd} dynamics of~\cite{chizat2025doubly}. Furthermore, for empirical measures, the functionals in $\mathbb{F}$ are tractable, and take the following form,
\begin{equation}
    \begin{aligned}
        \mathbb{B}_{\epsilon}(\hat{P}) &= \sum_{k=1}^{K}\lambda_{k}\sum_{i=1}^{n}\sum_{j=1}^{n_{k}}\gamma_{\epsilon,k,i,j}^{\star}d(z_{i}^{(P)}, z_{j}^{(Q_{k})})^{2},\\
        \mathbb{G}(\hat{P}) &= \dfrac{1}{n}\sum_{i=1}^{n}G(\hat{p}(z_{i}^{(P)})),\\
        \mathbb{V}(\hat{P}) &= \dfrac{1}{n}\sum_{i=1}^{n}V(z_{i}^{(P)}),\\
        \mathbb{U}(\hat{P}) &= \dfrac{1}{n^{2}}\sum_{i=1}^{n}\sum_{j=1}^{n}U(z_{i}^{(P)},z_{j}^{(P)}),
    \end{aligned}\label{eq:functionals}
\end{equation}
where $\hat{p}$ denotes the \emph{estimated} density of $\hat{P}$. Since $\hat{P}$ is empirical it is, in general, difficult to compute $\mathbb{G}$. A notable exception is the entropy functional (i.e., $G(s) = s\log s$), which adds a diffusion term to the gradient flow dynamics. This corresponds to Langevin dynamics~\citep{chizat2025doubly} (Equation~\ref{eq:langevin}). For general choices of $G$, one needs to estimate the density $\hat{p}$ at each iteration of the empirical flow.   

\begin{algorithm}[ht]
\caption{Empirical Time Discretized Wasserstein Gradient Flow}\label{alg:empirical_flow}
\begin{algorithmic}[1]
\Require $\lambda \in \Delta_{K}$, $n \in \mathbb{N}$, $m \in \mathbb{N}$, $\alpha \in \mathbb{R}_{+}$, $\epsilon \in [0,+\infty)$
\Ensure Barycenter support $\{ z_{i}^{(P)} \}_{i=1}^{n}$
\State $\{z_{0,i}^{(P)}\}_{i=1}^{n}$, such that $z_{0,i}^{(P)} \sim \mathcal{N}(0, \text{Id})$.
\For{$\tau \gets 1$ to $n_{\text{iter}}$}
   \State $\mathbb{F}(\hat{P}_{\tau}) \leftarrow \mathbb{V}(\hat{P}_{\tau}) + \mathbb{U}(\hat{P}_{\tau}) + \mathbb{G}(\hat{P}_{\tau})$
   \For{$k\leftarrow 1$ to $K$}
       \State Sample $\{z_{i}^{(Q_{k})}\}_{i=1}^{m}$ i  .i.d. from $Q_{k}$.
       \State $\mathbb{F}(\hat{P}_{\tau}) \leftarrow \mathbb{F}(\hat{P}_{\tau}) + \lambda_{k}\mathbb{W}_{2,\epsilon}(\hat{P}_{\tau}, \hat{Q}_{k})^{2}$
   \EndFor
   \State $z_{\tau+1,i}^{(P)} \leftarrow z_{\tau,i}^{(P)} + \alpha v_{\tau,i}$
\EndFor
\end{algorithmic}
\end{algorithm}

An immediate consequence of Equation~\ref{eq:functionals} is that our gradient flow formulation encompasses several previous works. For instance, with $G = U = V = 0$, we have the classical free-support barycenters of~\cite{cuturi2014fast}. Furthermore, for the entropy functional and $\epsilon > 0$, we have the \gls{npgd} algorithm of~\cite{chizat2025doubly}. Finally, as we show in Section~\ref{sec:joint-measures} below, with a specific metric on $\Omega = \mathcal{X}\times\mathcal{Y}$, we retrieve the algorithm of~\cite{montesuma2023multi}.

\subsection{Scaling with Mini-Batch OT}\label{sec:minibatch}

The computation of $\mathbb{B}_{\epsilon}$ in the discrete approach~\citep{cuturi2014fast} involves an \gls{ot} problem between the complete set of samples in the input measures. Therefore, when the input measures are large scale, the barycenter problem becomes infeasible. We circumvent this issue with mini-batch \gls{ot}~\citep{fatras2021minibatch}.

The main idea behind our scalable barycenter algorithm is accessing $\hat{Q}_{k} \in \mathcal{Q}$ through sampling. At each iteration of Algorithm~\ref{alg:empirical_flow}, we sample a mini-batch $\{z_{i}^{(Q_{k})}\}_{i=1}^{m}$ of $m$ samples from each input measure (cf. line 5 in Algorithm~\ref{alg:empirical_flow}). Since we sample the same number of samples from each measure, we can actually vectorize the computation of the $K$ \gls{ot} problems in the evaluation of $\mathbb{W}_{2,\epsilon}$.

Let $\{z_{\tau,i}^{(P)}\}_{i=1}^{n}$ be the barycenter support, and $\{z_{j}^{(Q_{k})}\}_{j=1}^{m}$ be a mini-batch from $Q_{k}$. The Gibbs kernel $E_{k,i,j} = e^{-(C_{k,i,j}/\epsilon)}$, and $C_{k,i,j} = d(z_{\tau,i}^{(P)},z_{j}^{(Q_{k})})^{2}$ are arrays with shape $(K,n,m)$. The Sinkhorn iterations become,
\begin{equation}
    \psi_{k,j}^{(\ell+1)} = \frac{b_{k,j}}{\sum_{i=1}^{n} E_{k,i,j}\, \phi_{k,i}^{(\ell)}},\phi_{k,i}^{(\ell+1)} = \frac{a_{k,i}}{\sum_{j=1}^{m} E_{k,i,j}\, \psi_{k,j}^{(\ell+1)}},\label{eq:sinkhorn_iterations}
\end{equation}
where the division operation should be understood coordinate-wise. Furthermore, $b_{j} = n^{-1}$ and $a_{k,i} = m^{-1}$ is the importance of each sample in the barycenter, and input measure support, respectively. For simplicity, we assume uniform weights. For $L \in \mathbb{N}$ iterations, we can retrieve $\gamma_{\epsilon,k}^{\star} = \text{diag}(\phi_{k}^{(L)})E_{k}\text{diag}(\psi_{k}^{(L)})$. Since each $\gamma_{\epsilon,k}^{\star}$ has shape $(n, m)$, we can actually stack them in an array of shape $(K,n,m)$, similarly to $E$ and $C$. 

The key insight is that the contractions $\sum_{i}E_{k,i,j}\phi_{k,i}^{(\ell)}$ and $\sum_{j}E_{k,i,j}\psi_{k,j}$ are batched matrix-vector products across the leading $K$ axis, meaning that \textbf{the $K$ \gls{ot} problems can be vectorized}. This is possible because we draw the same number of samples from each input measure, making it possible to vectorize the Sinkhorn iterations. For instance, a similar strategy is possible for measures over a fixed grid~\cite[Remark 4.16]{peyre2019computational}. The \gls{ot} plan $\gamma_{\epsilon}$, obtained at the end of the Sinkhorn iterations, is used to compute $\mathbb{B}_{\epsilon}$, in lines 4 through 7 in Algorithm~\ref{alg:empirical_flow}.

\subsection{Flows over Joint Measures}\label{sec:joint-measures}

An interesting advantage of the gradient flow formalism is that the only necessary ingredient is a differentiable metric over $\Omega$. Therefore, we generalize the setting of~\cite{cuturi2014fast}, who assumed $\Omega = \mathbb{R}^{d}$, and $d(z,z') = \lVert z - z' \rVert_{2}$. As we show in our experiments, in machine learning applications one has $\Omega = \mathcal{X}\times\mathcal{Y}$, i.e., measures over the joint space of features and labels. Let $z = (x, y),x\in\mathcal{X},y\in\mathcal{Y}$,
\begin{align}
    d(z,z') = \sqrt{\lVert x - x' \rVert_{2}^{2} + \beta\lVert y - y' \rVert_{2}^{2}},\label{eq:joint_metric}
\end{align}
where $\beta \geq 0$ is a parameter that balances the feature distance terms and the label distance terms.

For regression applications (i.e., $\mathcal{Y} = \mathbb{R}$), this distance is quite natural. However, for classification, $\mathcal{Y}$ is categorical (e.g., $\mathcal{Y} = \{1,\cdots,n_{\text{classes}}\}$). One possible strategy, used in~\cite{montesuma2021wasserstein} and~\cite{alvarez2021dataset}, is fixing the labels and flowing only the features. In contrast to these works, we embed $\mathcal{Y}$ into the compact continuous space $\Delta_{n_{\text{classes}}}$, through a one-hot encoding operation. For our flow, we parametrize labels through a change of variables,
\begin{align*}
    y_{i,c}^{(P)} = \text{softmax}(\ell_{i,1}^{(P)}, \cdots, \ell_{i,n_{c}}^{(P)}) = \dfrac{\exp(\ell_{i,c}^{(P)})}{\sum_{c=1}^{n_{c}}\exp(\ell_{i,c}^{(P)})},
\end{align*}
thus, instead of optimizing over $z=(x,y)$, we optimize over $z=(x,\ell)$, $x \in \mathbb{R}^{d}$, $\ell \in \mathbb{R}^{n_{\text{classes}}}$. From the soft probabilities, we can retrieve the actual discrete labels with an argmax, $y_{\text{hard},i}^{(P)} = \text{argmax}_{c=1,\cdots,n_{c}}y_{i,c}^{(P)}$.

\subsection{Task-Aware Regularizing Functionals}\label{sec:functionals}

One of the advantages of our proposed method is regularizing the barycenter calculation with internal, interaction, and potential energy functionals. This idea was already used in practice by~\cite{alvarez2021dataset} for transfer learning problems. Here we propose the following functionals,
\begin{align}
    V_{E}(z) &= - \eta_{V}\sum_{c=1}^{n_{\text{classes}}}y_{c}\log y_{c},\label{eq:entropy}\\
    U_{R}(z,z') &= \begin{cases}
        \eta_{U}h(d(x,x')) & \text{if }y_{\text{hard}}\neq y_{\text{hard}}',\\
        0&\text{otherwise}
    \end{cases},\label{eq:repulsion}
\end{align}
where $h:\mathbb{R}\rightarrow\mathbb{R}$ is lower semi-continuous and bounded from below, and $\eta_{V} > 0$ and $\eta_{U} > 0$ are scaling parameters. For instance, in our experiments we use the hinge loss, $h(u) = \text{max}(0, \text{margin} - d)$, where $\text{margin} \geq 0$ is a margin parameter. Equations~\ref{eq:entropy} and~\ref{eq:repulsion} correspond to entropy and repulsion, respectively. The first functional penalizes barycenters that have fuzzy labels. The second functional encourages classes to be well separated.

\subsection{Convergence}\label{sec:convergence}

One of the difficulties in analyzing the gradient flow of the barycenter functional comes from the fact that $P \mapsto \mathbb{W}_{2}(P,Q)^{2}$ is not geodesically convex in $\mathcal{P}_{2}(\Omega)$~\cite[Section 4.4]{santambrogio2017euclidean}. As a result, we are minimizing a non-convex functional. \textbf{Explicit constants and proofs are available in Appendix~\ref{appx:proof}}. Our analysis relies on a measure-theoretic version of the \gls{pl} inequality~\citep{polyak1964gradient,karimi2016linear},
\begin{align}
     \lVert \gradW \mathbb{B}(P) \rVert_{L_{2}(P)}^{2} \geq C_{\text{PL}} (\mathbb{B}(P) - \mathbb{B}^{\star}).\label{eq:main_pl_ineq}
\end{align}
When this inequality holds, we have the following convergence result.

\begin{theorem}\label{thm:main_result}
   Let $\Omega = \mathcal{B}(0, R)$ be the closed ball in $\mathbb{R}^{d}$ with radius $R > 0$. Let
   \begin{align*}
       P^{\star} = \argmin{P \in \mathcal{P}_{2}(\Omega)} \sum_{k=1}^{K}\lambda_{k}\mathbb{W}_{2}(P, Q_{k})^{2}
   \end{align*}
   be the continuous barycenter of $\mathcal{Q} = \{Q_{k}\}_{k=1}^{K}, Q_{k}\in\mathcal{P}_{2}(\Omega)$ $k=1,\cdots,K$, with barycentric coordinates $\lambda=(\lambda_{1},\cdots,\lambda_{K}) \in \Delta_{K}$. Approximate each $Q_{k}$ with $m$ i.i.d. samples $\{z_{i}^{(Q_{k})}\}_{i=1}^{m}$, $z_{i}^{(Q_{k})} \sim Q_{k}$ and $\{z_{\tau,i}^{(P)}\}_{i=1}^{n}$ be the barycenter support. Let $\hat{P}^{\star}$ be a minimizer of
   \begin{align*}
       \hat{P}^{\star} \in \argmin{\hat{P} \in \text{Emp}_{n}(\Omega)}  \sum_{k=1}^{K}\lambda_{k}\mathbb{W}_{2,\epsilon}(\hat{P}, \hat{Q}_{k})^{2}.
   \end{align*}   
   Under the \gls{pl} inequality in Equation~\ref{eq:main_pl_ineq} the following holds,
   \begin{equation}
       \begin{aligned}
           \mathbb{E}[\hat{\mathbb{B}}(\hat{P}_{\tau}) - \hat{\mathbb{B}}^{\star}] \leq e^{-C_{\text{PL}}\tau}(\mathbb{B}(\hat{P}_{0}) - \mathbb{B}^{\star}) + C_{R}\sqrt{\dfrac{C_{d, n}}{m}}
       \end{aligned},\label{eq:convergence_eq}
   \end{equation}
   where the expectation is taken with respect to samples from $Q_{k}$. The constants $C_{R}$ and $C_{d, n}$ depend on the radius $R$, number of barycenter samples $n$, and dimensions $d$. $m$ denotes the number of samples drawn from each $Q_{k}.$
\end{theorem}

The first term in the r.h.s. of Equation~\ref{eq:convergence_eq} decays exponentially with the iterations $\tau$, and is a direct consequence of the \gls{pl} inequality in Equation~\ref{eq:main_pl_ineq}. The second term captures the empirical approximation error of the involved measures, that is, $Q_{k} \in \mathcal{Q}$ and $P_{\tau}$. In practice, $\tau \rightarrow +\infty$ leads to an error governed by the empirical approximations.

\textbf{On the PL Inequality.} Equation~\ref{eq:main_pl_ineq} is a standard tool used in proving the convergence of non-convex optimization algorithms, however, it is an open question whether it holds in general settings. For instance,~\cite{chewi2020gradient} shows that this inequality holds for Guassian measures under specific spectral conditions. In the Appendix, we show that the same result holds for \emph{location-scatter families of measures} (Appendix~\ref{appx:pl_theory}), based on the isometry between these sets, and the Bures-Wasserstein manifold established by~\cite[Theorem 2.3]{alvarez2016fixed}. On top of this theoretical justification, we provide an experiment validating the bound on the Swiss roll measure (Appendix~\ref{appx:PLEmpiricalValidation}).
\begin{figure*}[!t]
    \centering
    \begin{subfigure}{0.15\linewidth}
        \includegraphics[width=\linewidth]{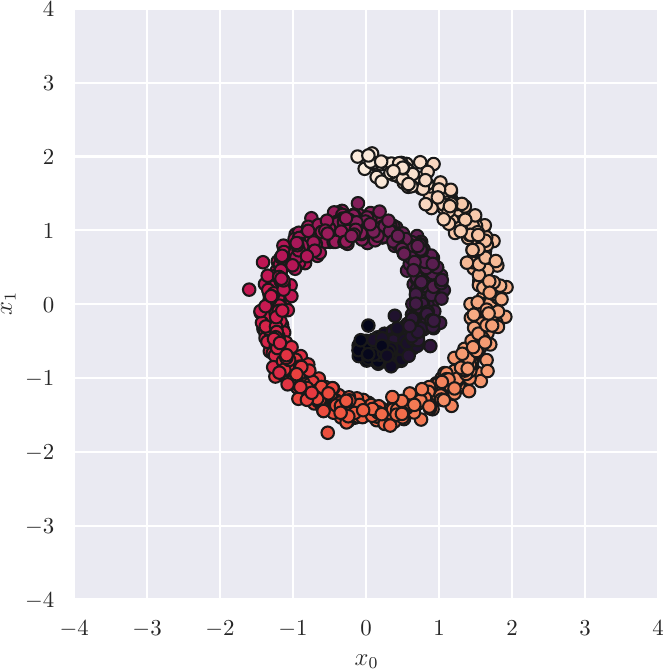}
        \caption{{$Q_{0}$}.}
    \end{subfigure}
    \begin{subfigure}{0.15\linewidth}
        \includegraphics[width=\linewidth]{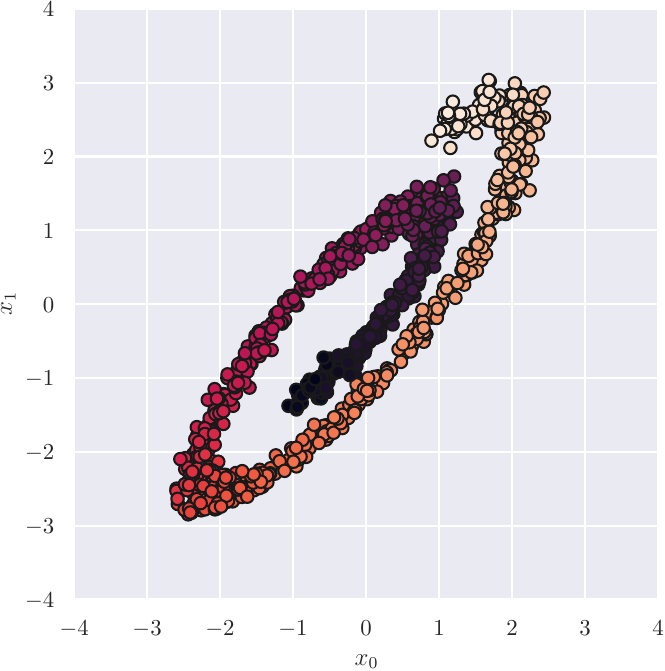}
        \caption{{$Q_{1}$}.}
    \end{subfigure}
    \begin{subfigure}{0.15\linewidth}
        \includegraphics[width=\linewidth]{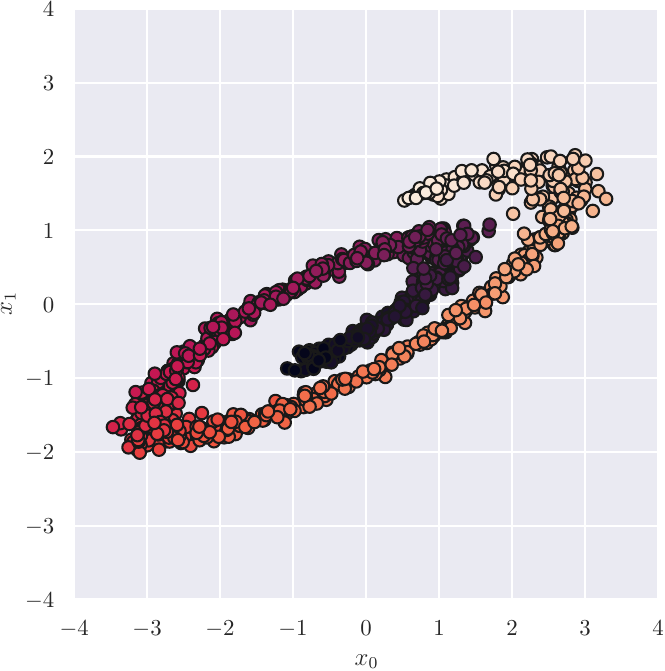}
        \caption{{$Q_{2}$}.}
    \end{subfigure}
    \begin{subfigure}{0.15\linewidth}
        \includegraphics[width=\linewidth]{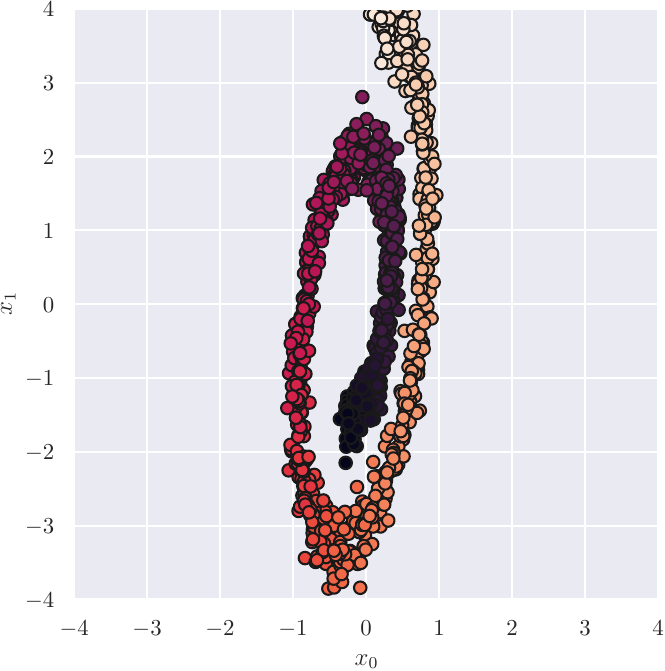}
        \caption{{$Q_{3}$}.}
    \end{subfigure}
    \begin{subfigure}{0.15\linewidth}
        \includegraphics[width=\linewidth]{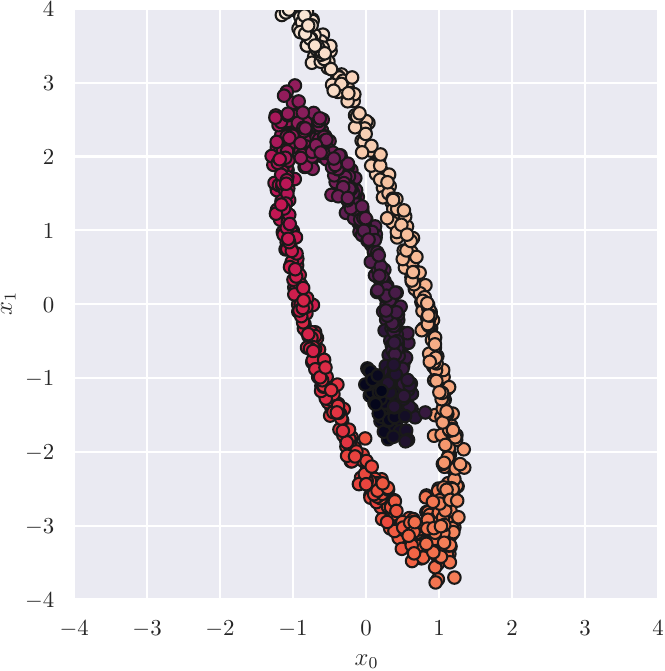}
        \caption{{$Q_{4}$.}}
    \end{subfigure}
    \begin{subfigure}{0.15\linewidth}
        \includegraphics[width=\linewidth]{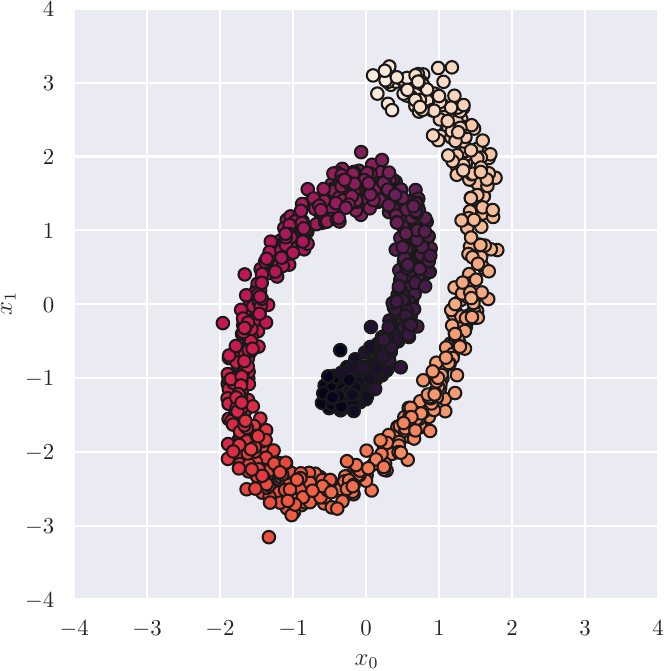}
        \caption{{$P^{\star}$.}}
    \end{subfigure}
    \caption{Swiss roll measures in a location-scatter family, where a source measure is transformed under an affine function $T_{k}(x) = A_{k}x+b_{k}$. Colors reflect the relative position of points in the manifold.}
    \label{fig:swiss-roll-measures}
\end{figure*}

\section{Experiments}\label{sec:experiments}

In the following, all experiments were conducted on a virtual machine with 8 CPUs (AMD EPYC 7413, 48GB of RAM) and an NVIDIA L4 GPU (24GB of VRAM). Appendix~\ref{appx:additional-details-da} includes additional details, fine-grained results, complexity analysis, and ablations.

\subsection{Swiss Roll Measures}\label{sec:swiss_roll}

In Figure~\ref{fig:swiss-roll-measures}, we show the Swiss roll measure~\citep{korotin2021continuous} $Q_{0}$, alongside four variations obtained via $T_{k,\sharp}Q_{0}$, $k=1,\cdots, 4$, $T_{k}(x) = A_{k}x + b_{k}$, for which the ground-truth barycenter is known in closed-form~\citep{alvarez2016fixed}, that is, $P^{\star} = (\sum_{k}\lambda_{k}T_{k})_{\sharp}Q_{0}$. These measures are shown in Figure~\ref{fig:swiss-roll-measures}.

\begin{figure}[!ht]
    \centering
    \begin{subfigure}{0.3\linewidth}
        \includegraphics[width=\linewidth]{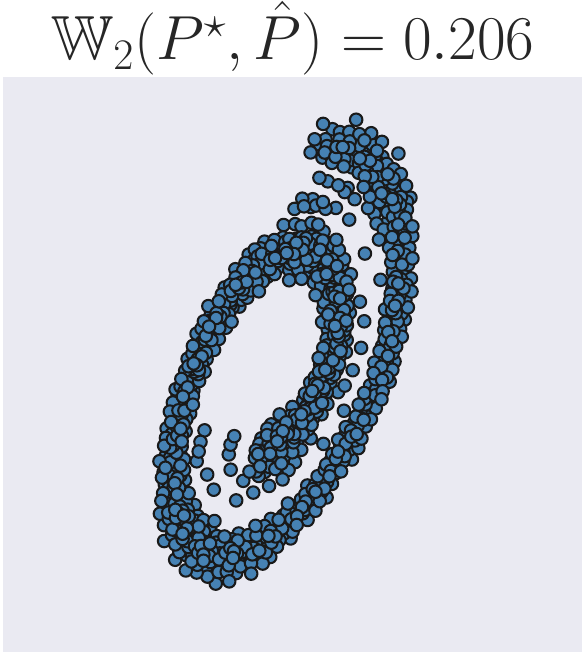}
        \caption{{Discrete}.}
    \end{subfigure}\hfill
    \begin{subfigure}{0.3\linewidth}
        \includegraphics[width=\linewidth]{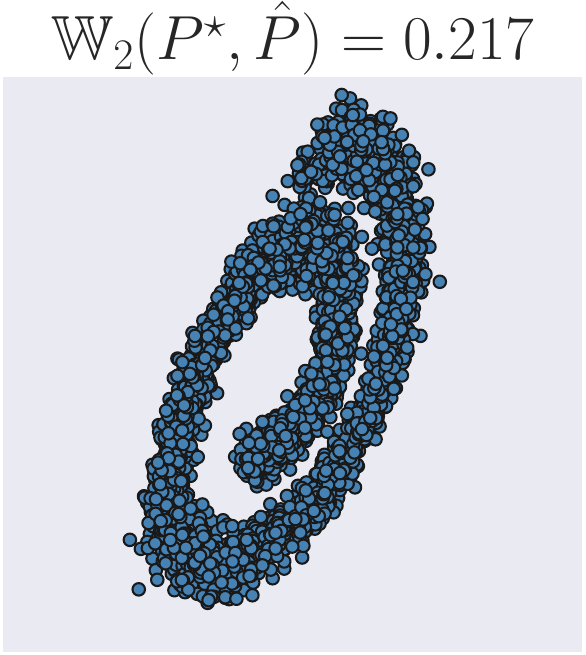}
        \caption{{CW2B}.}
    \end{subfigure}\hfill
    \begin{subfigure}{0.3\linewidth}
        \includegraphics[width=\linewidth]{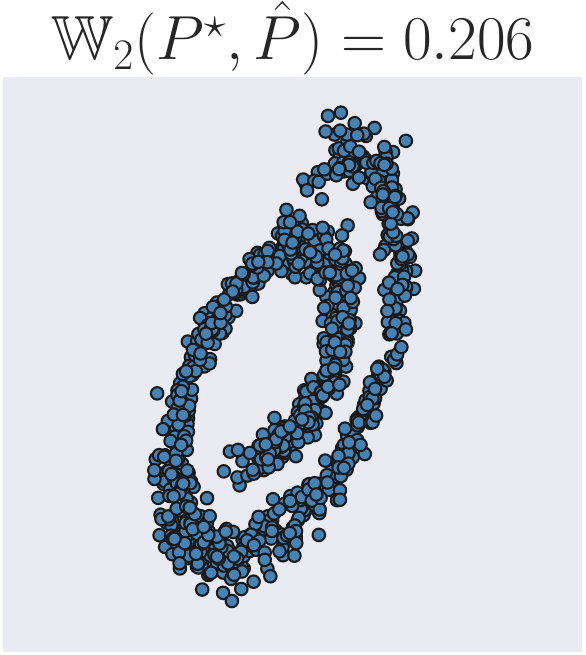}
        \caption{{U-NOT.}}
    \end{subfigure}\hfill

    \begin{subfigure}{0.3\linewidth}
        \includegraphics[width=\linewidth]{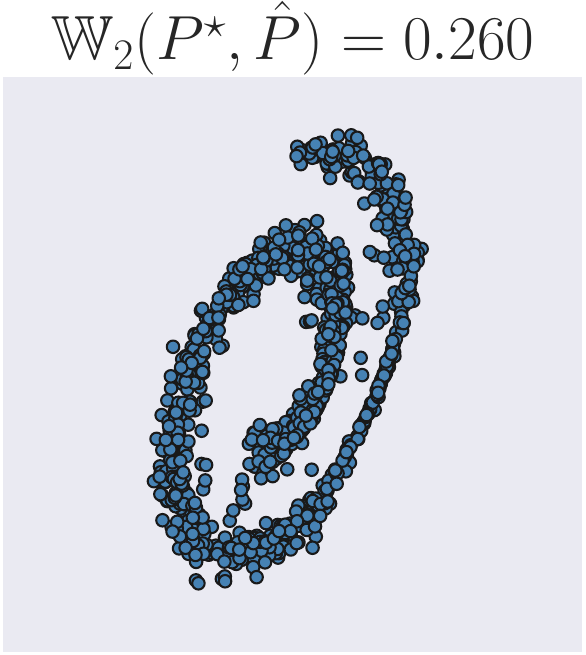}
        \caption{{NormFlow}.}
    \end{subfigure}\hfill
    \begin{subfigure}{0.3\linewidth}
        \includegraphics[width=\linewidth]{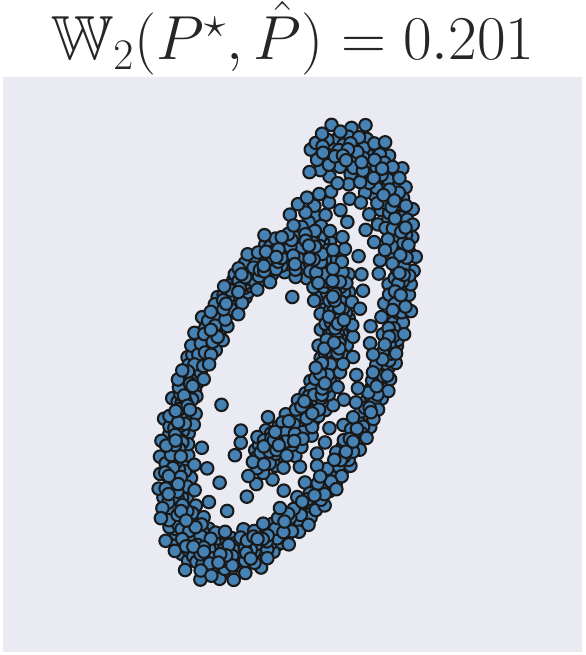}
        \caption{{WGF$^{\star}$.}}
    \end{subfigure}\hfill
    \begin{subfigure}{0.3\linewidth}
        \includegraphics[width=\linewidth]{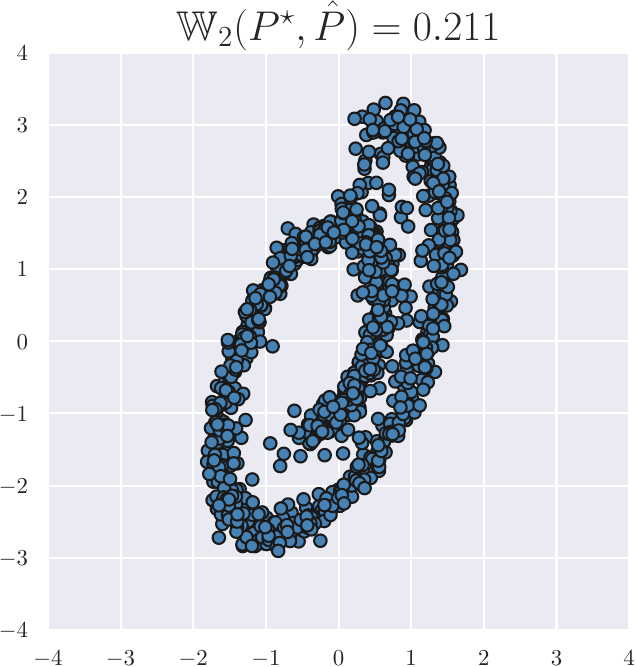}
        \caption{{WGF$_{\epsilon}^{\star}$.}}
    \end{subfigure}\hfill

    \caption{Comparison of unsupervised barycenter methods. Solvers (b, c, d) are neural, and (e, f) are ours.}
    \label{fig:swiss-roll-unlabeled}
\end{figure}

We compare five barycenter solvers, including discrete solvers~\citep{cuturi2014fast,montesuma2023multi}, and neural solvers: CW2B~\citep{korotin2021continuous}, U-NOT~\cite{gazdieva2024robust}, and NormFlow~\cite{visentin2025computing}. First, we benchmark unsupervised solvers in Figures~\ref{fig:swiss-roll-unlabeled} (a) through (f). Quantitatively, empirical methods, notably~\cite[Algorithm 2]{cuturi2014fast} and our \gls{wgf} algorithm (cf. Algorithm~\ref{alg:empirical_flow}) achieve the lowest Wasserstein distance to the ground truth in Figure~\ref{fig:swiss-roll-measures}. Overall, while the neural network solvers are usually more scalable than discrete methods in terms of number of samples, their optimization is more complicated and very sensitive to hyper-parameters.

\begin{figure}[h]
    \centering
    \begin{subfigure}{0.3\linewidth}
        \includegraphics[width=\linewidth]{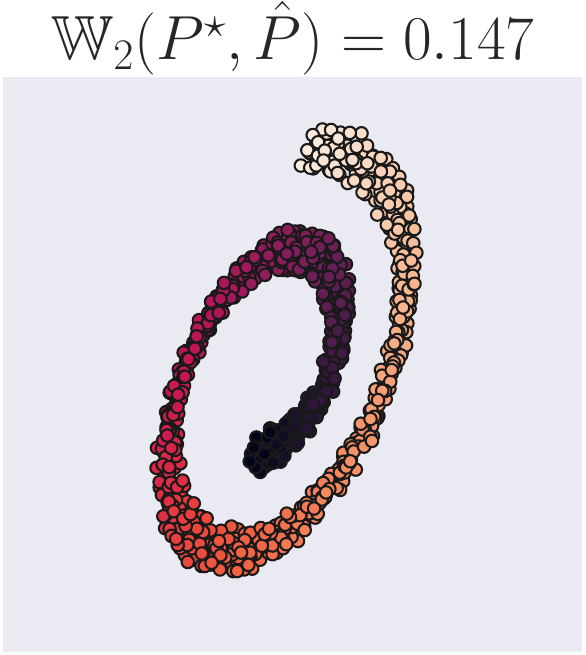}
        \caption{{Discrete.}}
    \end{subfigure}\hfill
    \begin{subfigure}{0.3\linewidth}
        \includegraphics[width=\linewidth]{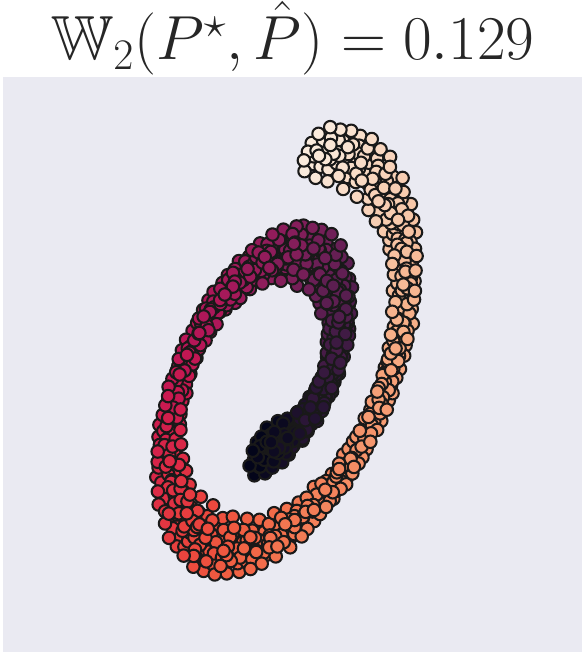}
        \caption{{WGF$^{\star}$.}}
    \end{subfigure}\hfill
    \begin{subfigure}{0.3\linewidth}
        \includegraphics[width=\linewidth]{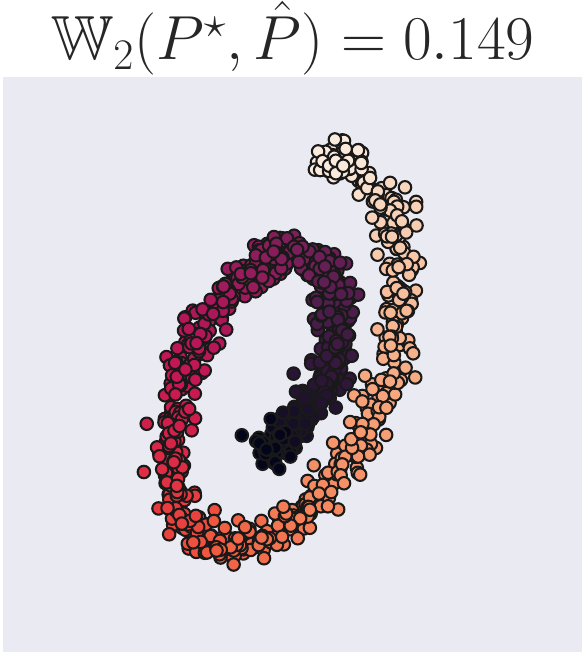}
        \caption{{WGF$_{\epsilon}^{\star}$}.}
    \end{subfigure}\hfill
    
    \caption{Comparison of supervised barycenter methods.}
    \label{fig:swiss-roll-labeled}
\end{figure}

Second, we experiment with integrating labels in the ground cost as described in Section~\ref{sec:joint-measures}. These methods are shown in Figures~\ref{fig:swiss-roll-labeled} (a), (b) and (c). In all cases, integrating labels produces barycenters that are closer to the ground-truth. We conclude that using the labels gives a strong inductive bias in the barycenter computation, which explains the gain in performance of labeled barycenters in the next section.

Next, we compare the running time of our \gls{wgf} solver with the discrete solver of~\cite{cuturi2014fast}, for an increasing barycenter support size $n \in \{2^{10},\cdots,2^{16}\}$ and batch size $m \in \{2^{8}, 2^{9}, 2^{10}\}$ for $200$ iterations. The discrete solver is run for $n \in \{2^{10}, \cdots, 2^{14}\}$. For this algorithm, larger support sizes \textbf{result in memory overflow}. Our \gls{wgf} is able to compute larger barycenters via mini-batching.

\begin{figure}[ht]
    \centering
    \begin{subfigure}{0.48\linewidth}
        \includegraphics[width=\linewidth]{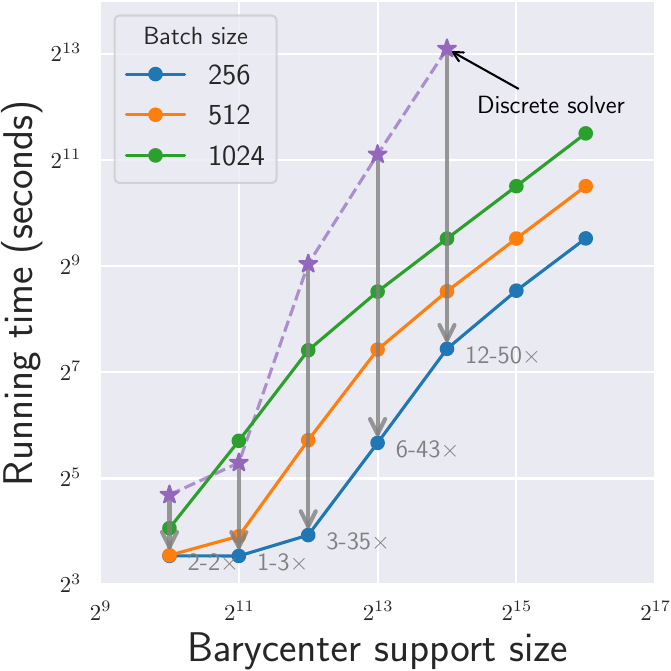}
        \caption{Scaling with $m$ ($\epsilon > 0$)}
    \end{subfigure}\hfill
    \begin{subfigure}{0.48\linewidth}
        \includegraphics[width=\linewidth]{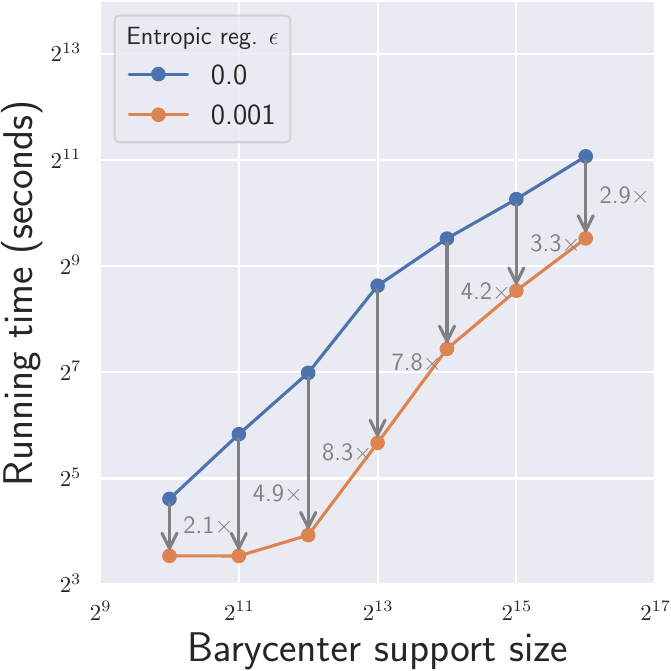}
        \caption{Scaling with $\epsilon$ ($m=256$)}
    \end{subfigure}\hfill
    \caption{Running time analysis for our \gls{wgf} algorithm. In (a), we compare the running time of our technique to that of the discrete solver of~\cite{cuturi2014fast}. In (b), we isolate the effect of GPU acceleration, showing consitent gains with increasing support size.}
    \label{fig:running-time-analysis}
\end{figure}

We summarize our results in Figure~\ref{fig:running-time-analysis}. First, in Figure~\ref{fig:running-time-analysis} (a), our \gls{wgf} leads to speedups ranging from $2\times$ $(n=2^{12},m=2^{10})$ to $50\times$ $(n=2^{14}, m=2^{8})$. Similar gains were obtained using Exact \gls{ot} on CPU (Appendix~\ref{appx:complexity}). Second, in Figure~\ref{fig:running-time-analysis} (b), we isolate the gain of using GPU acceleration and our vectorization strategy (Section~\ref{sec:minibatch}), showing consistent gains ranging from $2.1\times$ to $8.3\times$. Overall, our \gls{wgf} strategy provides compounding speedups from three sources: (i) minibatching, (ii) entropic regularization, and (iii) GPU parallelism.

\subsection{Multi-Source Domain Adaptation}\label{sec:msda}

One of the main applications of Wasserstein barycenters is multi-source domain adaptation (MSDA)~\citep{kouw2019review,sun2015survey,pan2009survey}. In this setting one needs to adapt multiple labeled source measures $Q_{1},\cdots,Q_{K}$ to a single unlabeled target measure $Q_{T}$. The goal is to learn, from samples $\{\{x_{i}^{(Q_{k})}, y_{i}^{(Q_{k})}\}_{i=1}^{n_{k}}\}_{k=1}^{K}$ and $\{x_{i}^{(Q_{T})}\}_{i=1}^{n_{T}}$, a classifier $h$ that achieves low risk or error in the target domain measure,
\begin{align*}
    \mathcal{R}_{Q_{T}}(h) = \mathbb{E}_{(x,y)\sim Q_{T}}[\mathcal{L}(y, h(x))],
\end{align*}
for a loss function $\mathcal{L}$ (e.g., cross-entropy loss). We isolate the quality of barycenters by doing adaptation at the level of embeddings. This approach allows us to perform domain adaptation in a higher semantic space, where distributions are more meaningful and comparable across domains. Thus, we assume that a meaningful feature extractor $\phi$, called the backbone, has been previously learned. We obtain the feature extractor by fine-tuning a neural network on the labeled source domain data (Appendix~\ref{appx:da-src-only}).

\noindent\textbf{Experimental Setup.} The unsupervised \gls{wgf} is fit with \emph{unlabeled} measures, i.e., $Q_{k} = n_{k}^{-1}\sum_{i}\delta_{x_{i}^{(Q_{k})}}$, where $x_{i}^{(Q_{k})}$ are embedding vectors. The supervised \gls{wgf} is fit with \emph{labeled} measures, that is, $Q_{k} = n_{k}^{-1}\sum_{i=1}^{n_{k}}\delta_{(x_{i}^{(Q_{k})}, y_{i}^{(Q_{k})})}$. In both cases, $\hat{P} = n^{-1}\sum_{i=1}^{n}\delta_{(x_{i}^{(P)}, y_{i}^{(P)})}$ which allows us to apply the regularizing functionals defined in Section~\ref{sec:functionals}.

\noindent\textbf{Wasserstein Barycenter Transport.}~\cite{montesuma2021wasserstein} proposed a technique based on Wasserstein barycenters for domain adaptation. The idea is to map the synthesized barycenter support to the target domain through the barycentric mapping~\citep{courty2016optimal}. Then, a classifier can be learned on the target domain based on the transported data. We adapt this idea to unsupervised barycenter algorithms by using the computed barycenter as a pivot domain. See Appendix~\ref{appx:da-unsup-wbt} for more details.

\textbf{Benchmarks.} We run our experiments on five benchmarks: Office 31~\citep{saenko2010adapting}, Office Home~\citep{venkateswara2017deep}, BCI-CIV-2a~\citep{brunner2008bci}, ISRUC~\citep{khalighi2013automatic}, and TEP~\citep{montesuma2024benchmarking}. The first two, second two, and last benchmarks correspond to computer vision, neuroscience, and chemical engineering benchmarks, respectively. We show in Table~\ref{tab:da-datasets} an overview of our experimental setting.

\begin{table}[ht]
    \centering
    \resizebox{\linewidth}{!}{
        \begin{tabular}{llcccc}
            \toprule
             Benchmark & Backbone & \# Samples & \# Domains & \# Dim. & \# Classes \\
             \midrule
             Office31 & ResNet50 & 3287 & 3 & 2048 & 31 \\
             BCI-CIV-2a & CBraMod & 5184 & 10 & 200 & 4 \\
             TEP & CNN & 17289 & 6 & 128 & 29 \\
             Office Home & ResNet101 & 15500 & 4 & 2048 & 65 \\
             ISRUC & CBraMod & 89240 & 100 & 512 & 5\\
             \bottomrule
        \end{tabular}
    }
    \caption{Overview of benchmarks used for domain adaptation, sorted by number of samples.}
    \label{tab:da-datasets}
\end{table}

\begin{figure*}[!b]
    \centering
    \begin{subfigure}[t]{0.3\linewidth}
        \centering
        \includegraphics[width=\linewidth]{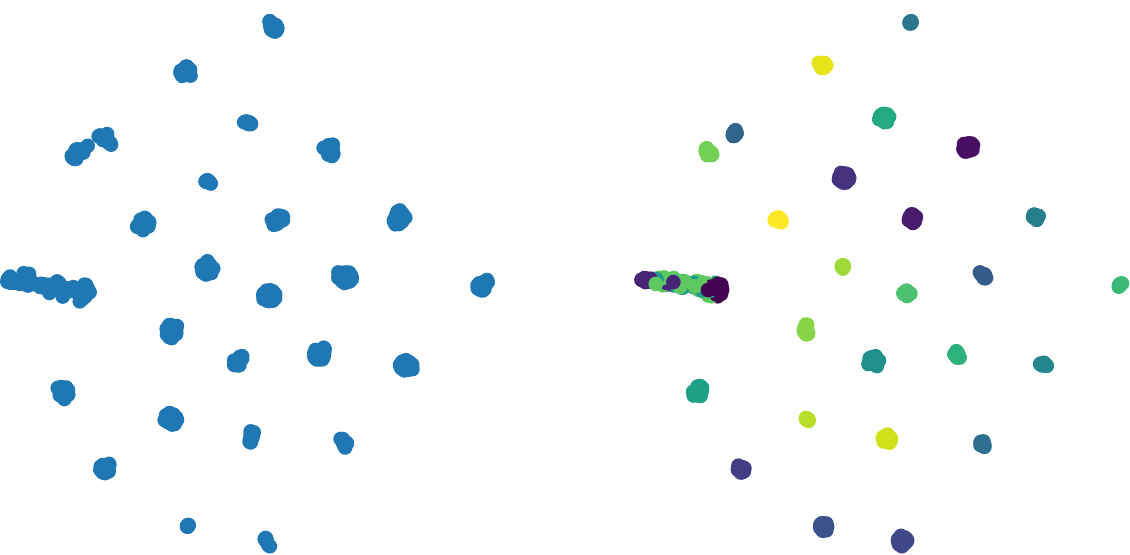}
        \caption{Unsupervised (left) vs. Supervised barycenters (right).}
    \end{subfigure}\hfill
    \begin{subfigure}[t]{0.3\linewidth}
        \centering
        \includegraphics[width=\linewidth]{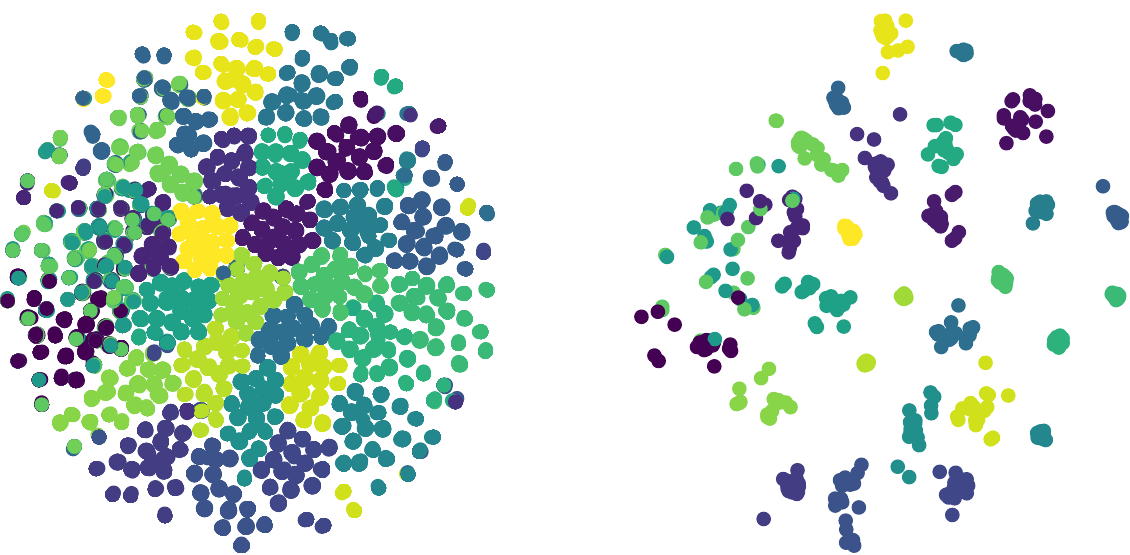}
        \caption{Transported sources (left) vs. supervised barycenter (right).}
    \end{subfigure}\hfill
    \begin{subfigure}[t]{0.3\linewidth}
        \centering
        \includegraphics[width=\linewidth]{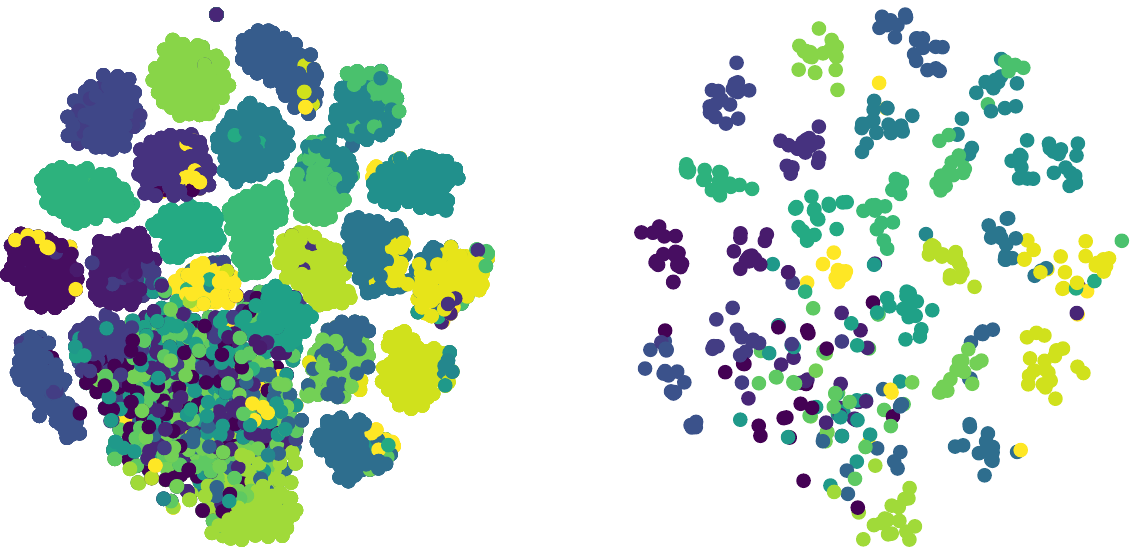}
        \caption{Transported samples in the target domain for the unsupervised (left) and supervised (right) WGF variants.}
    \end{subfigure}
    \caption{Comparison of unsupervised vs. supervised barycenter transports on the TEP benchmark.}
    \label{fig:tep-comparison}
\end{figure*}

\textbf{Compared Methods.} Overall, we compare seven Wasserstein barycenter strategies with ours. Besides those used in the previous section, we include the NOT approach of~\cite{kolesov2024estimating}, and the \gls{gmm} barycenter of~\cite{montesuma2024lighter}. For completeness, we include four other state-of-the-art methods in domain adaptation over embedding vectors. Those methods are: WJDOT~\cite{turrisi2022multi}, DaDiL-R, and E~\cite{montesuma2023multi}, and GMM-DaDiL~\cite{montesuma2024lighter}. For each benchmark, we use one domain as the target domain (e.g., $\text{Amazon}$ vs. $\{\text{dSLR, Webcam}\}$ in the Office 31 benchmark), and we measure the classification accuracy, i.e., the percentage of correct predictions.

\textbf{Main Results.} We present our main results in Table~\ref{tab:results-msda}, which reports the average performance per domain on each benchmark. We provide fine-grained results in Appendix~\ref{appx:additional-details-da}. In general, labeled barycenter methods (the 3 last rows in Table~\ref{tab:results-msda}) have a clear advantage over unsupervised methods. We argue that label information is essential in domain adaptation success, which is consistent with previous research~\citep{courty2016optimal,montesuma2021wasserstein,montesuma2023multi,montesuma2024lighter}. Among barycenter methods, our \gls{wgf} method achieves the best performance in all benchmarks, outperforming previous methods in MSDA, such as WJDOT and DaDiL in the ISRUC, BCI-CIV, and TEP benchmarks.

\begin{table}[ht]
    \centering
    \resizebox{\linewidth}{!}{
    \begin{tabular}{lccccccc}
        \toprule
        Benchmark & $\mathcal{X}\times\mathcal{Y}$ & Office31 & OfficeHome & BCI-CIV-2a & ISRUC & TEP & Avg. Rank\\
        \midrule
        Backbone & - & ResNet-50 & ResNet-101 & CBraMod & CBraMod & CNN & - \\ 
        \midrule
        Source-Only  & - & 86.40 & 75.95 &  50.30 & 76.63 & 78.48 & - \\
        \midrule
        WJDOT & - & 86.80 & 76.59 & N/A & 76.95 & 86.13 & 7.8\\
        DaDiL-R & - & 89.91 & 77.86 & 53.41 & 74.68 & 86.14 & 6.4\\
        DaDiL-E & - & 89.79 & 78.14 & N/A & 75.89 & 85.87 & 5.0\\
        GMM-DaDiL & - & 90.63 & 78.81 & 57.10 & 75.47 & 86.85 & 4.4\\
        \midrule
        Discrete & \xmark & 87.82 & 77.17 &  57.38 & 70.67 & 83.81 & 8.8\\
        NormFlow & \xmark & 85.91 & 76.73 & 55.45 & 78.04 & 82.89 & 10.0\\
        CW2B & \xmark & 86.37 & 76.44 & 57.25 & 75.84 & 85.83 & 9.0\\
        NOT & \xmark & 86.22 & 75.36 & 57.29 & 76.39 & 84.93 & 9.4\\
        U-NOT & \xmark & 86.97 & 76.85 & 57.17 & 77.06 & 85.43 & 7.8\\
        \rowcolor{lightgray} WGF (ours) & \xmark & 88.27 & 77.56 & \textbf{57.68} & 78.16 & 85.51 & 4.9\\
        \midrule
        Discrete & \cmark &  87.93 & 77.09 & 57.43 & 78.20 & 86.09 & 5.0\\
        GMM & \cmark & 88.54 & 77.87 & 57.04 & 74.58 & 84.67 & 8.2\\
        \rowcolor{lightgray}  WGF (ours) & \cmark & \textbf{89.52} & \textbf{78.42} &  \textbf{57.68} & \textbf{80.02} & \textbf{86.87} & \textbf{1.9}\\
         \bottomrule
    \end{tabular}
    }
    \caption{Average classification accuracy on target domains. The top group (WJDOT through GMM-DaDiL) include \textbf{general domain adaptation methods}. The middle and bottom groups show \textbf{barycenter solvers} without and with access to source labels ($\mathcal{X}\times\mathcal{Y} = $\xmark, $\mathcal{X}\times\mathcal{Y} = $\cmark), respectively. Bold indicates best among barycenter methods.}
    \label{tab:results-msda}
\end{table}

\noindent\textbf{Unsupervised vs. Supervised Barycenters.} An intriguing finding of Table~\ref{tab:results-msda} is that, on some benchmarks, unsupervised \gls{wgf} achieves decent performance, even outperforming supervised baselines (e.g., WJDOT of~\cite{turrisi2022multi}). We probe the reason why these methods have unreasonably good performance, by performing a t-SNE~\citep{maaten2008visualizing} visualization of the barycenters against the source domain samples, shown in Figure~\ref{fig:tep-comparison}. We see that even without labels, the barycenter samples cluster around the classes. This effect stems from the geometry of the embeddings, which were optimized (through supervision on source domains) to separate classes. As a result, unsupervised \gls{ot} already captures, to some extent, the cluster structure of the barycentric measure. Figures~\ref{fig:tep-comparison} (b) and (c) investigate the remaining performance gap, as we transport the measures towards the target. Figure~\ref{fig:tep-comparison} (b) shows the transported sources to the barycentric domain (left) and the synthesized barycenter (right), and Figure~\ref{fig:tep-comparison} shows the final transported samples in the target domain for the unsupervised strategy (left) and supervised one (right). We conclude that using a synthesized supervised barycenter is better than using an unsupervised barycenter as pivot domain (Appendix~\ref{appx:da-unsup-wbt}). Therefore, \emph{the inductive bias of using labels in the ground cost is essential for domain adaptation}. This further explains the performance of DaDiL methods, which directly synthesize the target domain through an approximate barycenter.

\begin{table*}[!t]
\centering
\resizebox{\linewidth}{!}{
\begin{tabular}{llccccccc}
\toprule
Dataset & Metric & \textit{Baseline} & \textbf{WGF}~($\mathcal{X}\!\times\!\mathcal{Y}$) & \textbf{WGF}$_{\text{u}}$~($\mathcal{X}$) & NormFlow & NOT & U-NOT & CW2B \\
\midrule
\multirow{2}{*}{Adult}$_{\text{Gender}}$    & Acc & $0.808_{\pm 0.002}$ & $0.793_{\pm 0.034}^{(5)}$          & $0.792_{\pm 0.004}^{(6)}$          & $\mathbf{0.806}_{\pm 0.003}^{(1)}$ & $0.794_{\pm 0.004}^{(3)}$          & $0.804_{\pm 0.002}^{(2)}$          & $0.794_{\pm 0.004}^{(4)}$          \\
                          & DI  & $1.955_{\pm 0.083}$ & $\mathbf{0.990}_{\pm 0.014}^{(1)}$ & $1.011_{\pm 0.013}^{(2)}$          & $1.462_{\pm 0.041}^{(6)}$          & $0.975_{\pm 0.157}^{(3)}$          & $1.253_{\pm 0.090}^{(5)}$          & $0.926_{\pm 0.039}^{(4)}$          \\
\cmidrule(lr){2-9}
\multirow{2}{*}{Adult}$_{\text{Race}}$  & Acc & $0.808_{\pm 0.002}$ & $\mathbf{0.833}_{\pm 0.020}^{(1)}$ & $0.799_{\pm 0.006}^{(6)}$          & $0.807_{\pm 0.002}^{(2)}$          & $0.807_{\pm 0.001}^{(3)}$          & $0.807_{\pm 0.002}^{(4)}$          & $0.806_{\pm 0.001}^{(5)}$          \\
                          & DI  & $0.613_{\pm 0.015}$ & $0.977_{\pm 0.007}^{(3)}$          & $0.985_{\pm 0.015}^{(2)}$          & $0.741_{\pm 0.034}^{(6)}$          & $\mathbf{0.997}_{\pm 0.051}^{(1)}$ & $0.865_{\pm 0.026}^{(5)}$          & $1.041_{\pm 0.019}^{(4)}$          \\
\cmidrule(lr){2-9}
\multirow{2}{*}{Adult}$_{\text{G+R}}$ & Acc & $0.808_{\pm 0.002}$ & $\mathbf{0.848}_{\pm 0.019}^{(1)}$ & $0.792_{\pm 0.002}^{(5)}$          & $0.802_{\pm 0.003}^{(2)}$          & $0.791_{\pm 0.002}^{(6)}$          & $0.801_{\pm 0.002}^{(3)}$          & $0.792_{\pm 0.005}^{(4)}$          \\
                          & DI  & $2.130_{\pm 0.180}$ & $0.905_{\pm 0.061}^{(3)}$          & $1.031_{\pm 0.047}^{(2)}$          & $1.859_{\pm 0.287}^{(6)}$          & $\mathbf{0.990}_{\pm 0.265}^{(1)}$ & $1.201_{\pm 0.254}^{(5)}$          & $0.843_{\pm 0.076}^{(4)}$          \\
\cmidrule(lr){2-9}
\multirow{2}{*}{COMPAS}$_{\text{Race}}$   & Acc & $0.674_{\pm 0.009}$ & $0.666_{\pm 0.019}^{(5)}$          & $0.659_{\pm 0.007}^{(6)}$          & $\mathbf{0.672}_{\pm 0.008}^{(1)}$ & $0.668_{\pm 0.006}^{(4)}$          & $0.670_{\pm 0.003}^{(3)}$          & $0.671_{\pm 0.006}^{(2)}$          \\
                          & DI  & $1.973_{\pm 0.176}$ & $1.050_{\pm 0.078}^{(3)}$          & $\mathbf{1.018}_{\pm 0.020}^{(1)}$ & $1.278_{\pm 0.178}^{(6)}$          & $1.071_{\pm 0.022}^{(4)}$          & $1.244_{\pm 0.060}^{(5)}$          & $1.036_{\pm 0.066}^{(2)}$          \\
\cmidrule(lr){2-9}
\multirow{2}{*}{Credit}$_{\text{Gender}}$   & Acc & $0.812_{\pm 0.002}$ & $\mathbf{0.813}_{\pm 0.012}^{(1)}$ & $0.802_{\pm 0.001}^{(6)}$          & $0.812_{\pm 0.002}^{(4)}$          & $0.811_{\pm 0.001}^{(5)}$          & $\mathbf{0.813}_{\pm 0.002}^{(1)}$ & $0.812_{\pm 0.001}^{(3)}$          \\
                          & DI  & $0.922_{\pm 0.113}$ & $1.087_{\pm 0.068}^{(4)}$          & $1.044_{\pm 0.019}^{(2)}$          & $1.067_{\pm 0.066}^{(3)}$          & $1.094_{\pm 0.135}^{(5)}$          & $\mathbf{1.001}_{\pm 0.084}^{(1)}$ & $1.104_{\pm 0.025}^{(6)}$          \\
\cmidrule(lr){2-9}
\multirow{2}{*}{Law}$_{\text{Gender}}$      & Acc & $0.631_{\pm 0.006}$ & $0.618_{\pm 0.014}^{(6)}$          & $0.627_{\pm 0.007}^{(5)}$          & $0.631_{\pm 0.001}^{(4)}$          & $0.632_{\pm 0.003}^{(3)}$          & $\mathbf{0.635}_{\pm 0.005}^{(1)}$ & $0.635_{\pm 0.001}^{(2)}$          \\
                          & DI  & $0.938_{\pm 0.012}$ & $1.018_{\pm 0.018}^{(3)}$          & $0.996_{\pm 0.004}^{(2)}$          & $\mathbf{1.001}_{\pm 0.009}^{(1)}$ & $1.042_{\pm 0.106}^{(6)}$          & $0.962_{\pm 0.030}^{(5)}$          & $1.022_{\pm 0.026}^{(4)}$          \\
\cmidrule(lr){2-9}
\multirow{2}{*}{Crime}$_{\text{Race}}$    & MSE & $0.0182_{\pm 0.0010}$ & $0.0424_{\pm 0.0045}^{(6)}$        & $0.0370_{\pm 0.0021}^{(5)}$        & $\mathbf{0.0182}_{\pm 0.0008}^{(1)}$ & $0.0306_{\pm 0.0046}^{(4)}$      & $0.0238_{\pm 0.0023}^{(3)}$        & $0.0192_{\pm 0.0014}^{(2)}$        \\
                          & MG  & $0.2122_{\pm 0.0207}$ & $0.0043_{\pm 0.0019}^{(2)}$        & $\mathbf{0.0000}_{\pm 0.0000}^{(1)}$ & $0.2111_{\pm 0.0217}^{(6)}$      & $0.1107_{\pm 0.0143}^{(3)}$        & $0.1287_{\pm 0.0311}^{(4)}$        & $0.2053_{\pm 0.0275}^{(5)}$        \\
\midrule
\multicolumn{3}{l}{\textit{Mean rank} (14 metric cells)} & $\mathbf{3.14}$ & $3.64$ & $3.50$ & $3.64$ & $3.36$ & $3.64$ \\
\bottomrule
\end{tabular}
}
\caption{Fairness-repair results across seven tabular benchmarks
  (mean $\pm$ std over 3 seeds; superscript in each cell is the
  per-dataset rank among the six methods, $1$ = best).
  Classification reports \textbf{Acc} (higher is better) and
  \textbf{DI} (disparate impact, target $1$; distance is
  $|\text{DI}-1|$). Crime (regression) reports \textbf{MSE}
  (lower is better) and \textbf{MG}, the group mean-prediction
  gap (target $0$). Rank-1 means are \textbf{bold}. WGF
  repairs the joint $\mathcal{X}\!\times\!\mathcal{Y}$;
  WGF$_\text{u}$ is the feature-only ablation on $\mathcal{X}$.
}
\label{tab:fairness}
\end{table*}

\subsection{Fairness}

In this section, we examine the performance of our \gls{wgf} against four neural solvers (NF, NOT, U-NOT and CW2B) in the context of algorithmic fairness. Briefly, the problem consists of learning a model $h \in \mathcal{H}$ that is fair against a protected variable $s$ (e.g., gender).~\cite{gordaliza2019obtaining} formalized this task, by considering probability measures $P(X|S=s)$. These authors proposed a method called \emph{repairing}, in which (1) they compute the barycenter of $\{P(X|S=s)\}_{s=1}^{S}$ and (2) they align each $P(X|S=s)$ with the barycenter. Inspired by our domain adaptation results, we also propose a new method based on the joint alignment, meaning, computing the barycenter of $P(X,Y|S=s)$.

We test the two versions of our \gls{wgf} algorithm (marginal, $\mathcal{X}$, and joint, $\mathcal{X}\times\mathcal{Y}$ repairs) against four neural barycenter solvers (NormFlow, NOT, U-NOT and CW2B) over five datasets: Adult income~\citep{kohavi1996scaling}, COMPAS~\citep{angwin2022machine}, Law school bar exam~\citep{wightman1998lsac}, Communities and crime~\citep{redmond2002data}. All algorithms are evaluated against two axes: (i) downstream performance (e.g., accuracy for classification, mean squared error for regression); (ii) fairness, that is, the disparate impact metric,
\begin{align*}
    \text{DI} = \dfrac{\text{Pr}(\hat{Y}=1|S=\text{unprivileged})}{\text{Pr}(\hat{Y}=1|S=\text{privileged})},
\end{align*}
which measures the ratio, in probability, of picking the unprivileged group over the privileged group (the closer to $1$, the better), and for regression we use the maximum gap,
\begin{align*}
    \text{MG} = | \mathbb{E}[\hat{Y}|S=\text{unprivileged}] - \mathbb{E}[\hat{Y}|S=\text{privileged}] |.
\end{align*}
Our results are summarized in Table~\ref{tab:fairness}. Overall, similarly to domain adaptation, repairing the joint measures has the best average performance.

\section{Conclusion}\label{sec:conclusion}

We have presented a new Wasserstein barycenter algorithm (Algorithm~\ref{alg:empirical_flow}), inspired by the gradient flow literature~\citep{santambrogio2015optimal}, that addresses two persistent challenges: scalability and regularization. We have derived a new algorithm that is amenable to mini-batch \gls{ot}~\citep{fatras2021minibatch} (Section~\ref{sec:minibatch}). Under entropic \gls{ot}, we further demonstrate that \gls{ot} computations can be vectorized across the $K$ input measures, leading to further speedups (Figure~\ref{fig:running-time-analysis}).

Our framework is flexible, and accommodates modular plug-in regularization from internal, potential, and interaction energies (Section~\ref{sec:functionals}). Extensive experiments on domain adaptation demonstrate the value of our approach and that incorporating labels in the \gls{ot} ground-cost is necessary for state-of-the-art performance (Table~\ref{tab:results-msda}).

Finally, our work highlights several promising avenues for future research. Specifically, our framework could be extended to more complex differentiable structures, including Riemannian manifolds. Additionally, it prompts further investigation into how to effectively integrate labels and other regularization schemes into neural network solvers.




\bibliography{refs}

\newpage

\onecolumn


\appendix

\tableofcontents

\section{Introduction}\label{appx:intro}

The goal of this supplementary material is to provide additional background on the main paper, as well as to provide missing proofs, additional experiments, and the complete experimental setting. These details were left out of the main paper due to length constraints. We start by summarizing the differences between our algorithm and prior work. Briefly, discrete solvers~\citep{cuturi2014fast,montesuma2023multi} share WGF’s free-support parametrization and parameter count, but operate on full-batches without K-wise Sinkhorn vectorization. Most neural solvers (CW2B~\citep{korotin2021continuous}, NOT~\citep{kolesov2024estimating}, U-NOT~\citep{gazdieva2024robust}) scale via mini-batching, but require $\mathcal{O}(K)$ networks, multi-level optimization, many tuned hyper-parameters, and impose ground-cost restrictions that block extensions to joint measures. A middle ground approach is NormFlow~\citep{visentin2025computing}, based on normalizing flows, which poses a single-level minimization problem and uses conditioning rather than $\mathcal{O}(K)$ neural networks.

\begin{table}[h]
\centering
\scriptsize
\setlength{\tabcolsep}{4pt}
\renewcommand{\arraystretch}{1.25}
\begin{tabularx}{\textwidth}{@{}l*{6}{>{\raggedright\arraybackslash}X}@{}}
\toprule
& \textbf{Discrete}~\citep{cuturi2014fast,montesuma2023multi}
& \textbf{CW2B}~\citep{korotin2021continuous}
& \textbf{NormFlow}~\citep{visentin2025computing}
& \textbf{NOT}~\citep{kolesov2024estimating}
& \textbf{U-NOT}~\citep{gazdieva2024robust}
& \textbf{WGF (ours)} \\
\midrule
Barycenter parametrization
& $n$ free-support samples $\{z_i\}$
& $K$ ICNN potentials $g_k$
& One conditional normalizing flow $f_\theta(\cdot, s)$
& $K$ potentials $f_k$ + $K$ maps $T_k$
& $K$ potentials $f_k$ + $K$ maps $T_k$
& $n$ free-support samples $\{z_i\}$ \\
Parameter complexity
& $n$ (positions)
& $K \cdot |g|$
& $|f|$ (shared across $K$)
& $K \cdot (|f| + |T|)$
& $K \cdot (|f| + |T|)$
& $n$ (positions) \\
Optimization structure
& Single-level (fixed-point)
& Min-max-min over potentials
& Single-level (likelihood + $L^2$)
& Max-min over $(f, T)$
& Max-min over $(f, T)$
& Single-level (gradient flow) \\
Per-iteration cost
& $\mathcal{O}(K \cdot n^3 \log n)$
& $\mathcal{O}(K \cdot \text{backprop on } g_k)$
& $\mathcal{O}(\text{backprop on } f)$
& $\mathcal{O}(K \cdot \text{backprop on } f, T)$
& $\mathcal{O}(K \cdot \text{backprop on } f, T)$
& $\mathcal{O}(K \cdot L \cdot m \cdot n)$ entropic; $\mathcal{O}(K \cdot m \cdot n)$ exact \\
Mini-batch over $Q_k$
& No (full-batch)
& Yes
& Yes
& Yes
& Yes
& Yes \\
$K$-wise OT vectorization
& No
& N/A
& N/A
& N/A
& N/A
& Yes (\S\ref{sec:minibatch}) \\
Ground cost flexibility
& Any differentiable $d$
& Squared Euclidean (ICNN-imposed)
& Any cost satisfying Gangbo--McCann
& Any (weak OT)
& Any (weak OT)
& Any differentiable $d$ \\
Label-aware joint $\mathcal{X} \times \mathcal{Y}$
& Yes~\citep{montesuma2023multi}
& No (ICNN convexity broken by softmax)
& No (softmax breaks invertibility)
& No (no published label extension)
& No (no published label extension)
& Yes (\S\ref{sec:joint-measures}) \\
Number of hyperparameters
& Few ($n$, $\epsilon$)
& Many ($\tau$, $\lambda$, ICNN rank, NN width/depth, LR, $\beta$, pretrain)
& Many ($L$, $K$, NN width, LR schedule)
& Many (LR$_f$, LR$_T$, NN width/depth, $M_T$ inner, $\epsilon$ or $\gamma$)
& Many (LR$_f$, LR$_T$, LR$_m$, $\tau$, $\psi_k$, NT inner)
& Few ($n$, $m$, $\alpha$, $\epsilon$, $\eta_V$, $\eta_U$, $\eta_G$) \\
Reference (paper)
& \S B.2 & \S B.3 & \S B.3 & \S B.3 & \S B.3 & \S\ref{sec:wgfs}, Alg.~\ref{alg:empirical_flow} \\
Reference (timing)
& Tab.~\ref{tab:Running time} & Tab.~\ref{tab:Running time} & Tab.~\ref{tab:Running time} & Tab.~\ref{tab:Running time} & Tab.~\ref{tab:Running time} & Tab.~\ref{tab:Running time} \\
\bottomrule
\end{tabularx}
\caption{Algorithmic comparison between WGF and the compared barycenter solvers.}
\label{tab:algorithmic_comparison}
\end{table}

\noindent\textbf{Organization.} We divide this supplementary material as follows. Section~\ref{appx:survey} provides a brief survey on the methods for computing Wasserstein barycenters. Section~\ref{appx:calculus} provides additional background on \gls{ot}. Section~\ref{appx:gradients} covers additional details on the potential, internal, and interaction energies, as well as the barycenter objectives and their gradients. Section~\ref{appx:proof} provides missing proofs. Finally, section~\ref{appx:additional-details-da} provide additional experiments.

\section{A Brief Survey on Wasserstein Barycenters}\label{appx:survey}

\subsection{Introduction}

Given a finite family of measures $ \mathcal{Q} = \{ Q_{k} \}_{k=1}^{K}$, the Wasserstein barycenter problem was first introduced by~\cite{agueh2011barycenters} as the following minimization problem,
\begin{align}
    P^{\star} = \argmin{P \in \mathcal{P}_{2}(\Omega)} \biggr\{ \mathbb{B}_{\epsilon}(P|\mathcal{Q}) = \sum_{k=1}^{K}\lambda_{k}\mathbb{W}_{2,\epsilon}(P, Q_{k})^{2} \biggr\},\label{eq:WassersteinBarycenter}
\end{align}
where $\epsilon \geq 0$ denotes the entropic regularization weight. This minimization problem is the analogue, in Wasserstein space, to the \emph{Euclidean barycenter} problem $x \mapsto \sum_{k}\lambda_{k}d(x,x_{k})^{2}$, which is made possible since $(\mathcal{P}_{2}(\Omega), \mathbb{W}_{2})$ is a metric space. Solving problem~\ref{eq:WassersteinBarycenter} is, in general, challenging. Except for the Gaussian case, there are no closed-form solutions. In the following, we review a few strategies for computing Wasserstein barycenters.

\subsection{Discrete Barycenter Solvers}

Discrete barycenter solvers are based on the empirical approximation of the underlying measures, that is,
\begin{align}
    \hat{P}(z) = \sum_{i=1}^{n}a_{i}\delta(z - z_{i}^{(P)}) \in \text{Emp}_{n}(\Omega).\label{eq:emp_meas}
\end{align}
Here, $\{z_{i}^{(P)}\}_{i=1}^{n}$ is the support of $\hat{P}$ and $a \in \Delta_{n} = \{a\in\mathbb{R}_{+}^{n}:a_{i} \geq 0 \text{, and }\sum_{i=1}^{n} a_{i} = 1\}$ are the sample weights. There are two prevalent strategies when discretizing $P$. First, one can fix a grid of bins over the underlying space $\Omega$, and compute $a_{i} = P(z_{i}^{(P)})$ as the bin weight. This strategy is equivalent to estimating the density of $P$ over $\Omega$. Second, we can sample $z_{i}^{(P)} \overset{\text{i.i.d.}}{\sim} P$, in which case $a_{i} = \nicefrac{1}{n}$. In this paper, we deal primarily with \emph{free-support} Wasserstein barycenters. Arguably, the first free-support algorithm was proposed by~\cite{cuturi2014fast}. Their strategy comes from plugging Equation~\ref{eq:emp_meas} into~\ref{eq:WassersteinBarycenter},
\begin{equation}
    \begin{aligned}
        Z^{(P),\star} &= \argmin{z_{1}^{(P)}, \cdots, z_{n}^{(P)} \in \Omega}\biggr\{ \mathbb{B}_{\epsilon}(P|\mathcal{Q}) = \sum_{k=1}^{K}\lambda_{k}\sum_{i=1}^{n}\sum_{j=1}^{n_{k}}\gamma_{\epsilon,k,i,j}\lVert z_{i}^{(P)} - z_{j}^{(Q_{k})} \rVert_{2}^{2} \biggr\},\\
        \text{subject to }& \gamma_{\epsilon,k}^{\star} = \argmin{\gamma \in \Gamma(P, Q_{k})}\sum_{i=1}^{n}\sum_{j=1}^{n_{k}}\gamma_{ij}\lVert z_{i}^{(P)} - z_{j}^{(Q)} \rVert_{2}^{2} + \epsilon \sum_{i=1}^{n}\sum_{j=1}^{n_{k}}\gamma_{ij}\log\gamma_{ij}
    \end{aligned}\label{eq:cuturi_objective}
\end{equation}
At this point, one should compare Equation~\ref{eq:cuturi_objective} with Equation~\ref{eq:functionals} in the main paper. \cite{cuturi2014fast} propose solving this equation via a block coordinate descent strategy. Indeed, assuming $\gamma_{k} = \text{OT}(P, Q_{k})$ has been calculated (e.g., through linear programming), then,
\begin{align*}
    \nabla_{z_{i}^{(P)}}\mathbb{B}_{\epsilon}(P|\mathcal{Q}) = 2\sum_{k=1}^{K}\lambda_{k}\sum_{j=1}^{n_{k}}\gamma_{k,i,j}(z_{i}^{(P)}-z_{j}^{(Q_{k})})= \dfrac{2}{n}z_{i}^{(P)} - 2\sum_{k=1}^{K}\lambda_{k}\sum_{j=1}^{n}\gamma_{k,i,j}z_{j}^{(Q_{k})}.
\end{align*}
As a consequence, $\nabla_{z_{i}^{(P)}}\mathbb{B}_{\epsilon}(P|\mathcal{Q}) = 0$ leads to an update rule in terms of the barycentric projection of $P$ onto $Q_{k}$, which is a discrete approximation of the Monge map~\cite{deb2021rates}.

An important question arrises when the samples from $P$ are feature-label joints, first raised by~\cite{montesuma2023multi}, who proposed to use the ground-cost,
\begin{align}
    d(z,z') = \sqrt{\lVert x - x' \rVert_{2}^{2} + \beta\lVert y - y' \rVert_{2}^{2}},\label{eq:joint_metric_appx}
\end{align}
where $y$ and $y'$ are one-hot encoded labels, and $\beta \geq 0$ strikes a balance between the features and labels terms. As we show in the main paper, since this ground-cost is a metric, computing the \gls{ot} cost with $d^{2}$ as the ground-cost is equivalent to computing the $2-$Wasserstein distance on $\mathcal{X}\times\mathcal{Y}$.

The first paper to remark on these computations was~\cite{montesuma2023multi}, who proposed to couple these mappings,
\begin{align}
    x_{\tau+1,i}^{(P)} = \sum_{k=1}^{K}\lambda_{k}\biggr( \underbrace{n\sum_{i=1}^{n}\sum_{j=1}^{n_{k}}\gamma_{k,i,j}x_{j}^{(Q_{k})}}_{T_{P\rightarrow Q_{k}}^{\star}(x_{\tau,i}^{(P)})} \biggr)\text{, and, }y_{\tau+1,i}^{(P)} = \sum_{k=1}^{K}\lambda_{k}\biggr( \underbrace{n\sum_{i=1}^{n}\sum_{j=1}^{n_{k}}\gamma_{k,i,j}y_{j}^{(Q_{k})}}_{T_{P\rightarrow Q_{k}}^{\star}(y_{\tau,i}^{(P)})} \biggr).\label{eq:montesuma_iterations}
\end{align}
These iterations have a few nice properties with respect to the label term. First, the mappings $T_{P\rightarrow Q_{k}}^{\star}(y_{i,\tau}^{(P)})$ correspond to label propagation terms first defined in~\cite{redko2019optimal}. Second, since the summations in Equation~\ref{eq:montesuma_iterations} are convex combinations of the labels, the update rules naturally satisfy $y_{\tau, i}^{(P)} \in \Delta_{n_{\text{classes}}}$. Third, these iterations can be understood as \emph{fixed-point iterations} in the sense of~\cite{alvarez2016fixed}. The authors use this algorithm as the building block of their dictionary learning strategy.


\subsection{Neural Barycenter Solvers}

In this section, we cover a collection of methods that use neural networks as the foundation for computing Wasserstein barycenters. These methods are often referred to as "continuous", thus relying on the idea that the calculated barycenter is a continuous measure. Note that this is different from claiming that these barycenters solve a \emph{continuous problem}. In fact, most methods are formulated in terms of expectations of the involved measures, which are approximated through mini-batching. Indeed, mini-batching is the main tool for scalability in Wasserstein barycenters.

\noindent\textbf{Generative Modeling.} One of the first methods was proposed by~\cite{cohen2020estimating}, who proposed viewing Equation~\ref{eq:WassersteinBarycenter} through the lens of generative modeling. Let $g_{\theta}$ denote a neural network with parameters $\theta$. Then,
\begin{align*}
    \theta^{\star} = \argmin{\theta \in \Theta}\sum_{k=1}^{K}\lambda_{k}\mathbb{W}_{2}(P_{\theta}, Q_{k})^{2},
\end{align*}
where $P_{\theta} = g_{\theta,\sharp}P_{0}$ is the push-forward of a latent measure $P_{0}$ by the generator $g_{\theta}$. Therefore, the Wasserstein barycenter is parametrized via the weights of $g_{\theta}$. In comparison, methods such as~\cite{cuturi2014fast} and~\cite{montesuma2023multi} parametrize the barycenter through its samples. The underlying idea is sampling from each $Q_{k}$ and $P_{0} = \mathcal{N}(0, \sigma^{2} I)$.

\textbf{Input Convex Neural Nets.} More refined settings exploit the properties of the 2-Wasserstein distance with the squared Euclidean ground-cost. Indeed, from~\cite[Equation 4]{makkuva2020optimal}, exploiting the decomposition
\begin{align*}
    \lVert z - z' \rVert_{2}^{2} = \lVert z \rVert_{2}^{2} + \lVert z' \rVert_{2}^{2} - 2\langle z,z' \rangle,
\end{align*}
we can obtain,
\begin{align}
    \dfrac{1}{2}\mathbb{W}_{2}(P, Q)^{2} = \underbrace{\int \dfrac{\lVert z \rVert_{2}}{2}dP(z) + \int \dfrac{\lVert z \rVert_{2}}{2}dQ(z)}_{\mathbb{C}(P,Q)} - \min{f,g \in \Phi}  \int f(z)dP(z) + \int g(z)dQ(z) ,\label{eq:dual_ot}
\end{align}
where $(f, g) \in \Phi = \{f(x)+g(y) \leq \lVert  x - y\rVert_{2}^{2}\}$ are called potentials. The main insight of~\cite{makkuva2020optimal} is rewriting the minimization problem in the r.h.s. of Equation~\ref{eq:dual_ot} as,
\begin{align*}
    \sup{f \in \text{CVX}}\inf{g \in \text{CVX}}  \underbrace{-\int f(z)dP(z) - \int (\langle z,\nabla g(z) \rangle - f(\nabla g(z))) dQ(z)}_{V_{P, Q}(f, g)}
\end{align*}
Based on these equations,~\cite{fan2020scalable} propose solving,
\begin{align}
    (h,\{ f_{k} \}_{k=1}^{K}, \{g_{k}\}_{k=1}^{K}) &= \min{h}\sup{f_{k} \in \text{CVX}}\inf{g_{k} \in \text{CVX}}\dfrac{1}{2} \int \lVert h(z) \rVert_{2}dP_{0}(z) + \sum_{k=1}^{K}\lambda_{k}V_{P,Q_{k}}(f_{k}, g_{k}).\label{eq:crwb}
\end{align}
Here, samples in the Wasserstein barycenter are generated through the forward pass $h(z)$, where $z \sim P_{0} = \mathcal{N}(0, \sigma^{2} I)$. This is a min-max-min problem, which, as with many problems involving neural nets (e.g., generative adversarial nets~\citep{goodfellow2014generative}), is subject to saddle points.

An important aspect of Equation~\ref{eq:crwb} is the minimization with respect to convex functions $f_{k}$ and $g_{k}$. Indeed, these are the potentials of the transportation problem between the barycenter $P$ and each $Q_{k}$. More specifically, in light of Brenier's theorem~\citep{brenier1991polar}, $\nabla g_{k}$ is the Monge map from $Q_{k}$ to the barycenter $P$. These elements motivate~\cite{fan2020scalable} to use \glspl{icnn}, as in~\cite{makkuva2020optimal}, for the optimization problem in Equation~\ref{eq:crwb}.

\textbf{Continuous 2-Wasserstein Barycenter.} We now cover a family of methods relying on the fixed-point view of Wasserstein barycenters introduced by~\cite{alvarez2016fixed}. Indeed, these authors show that the Wasserstein barycenter is a fixed point of the operator $\Psi:P \mapsto (\sum_{k}\lambda_{k}T_{P\rightarrow Q_{k}})_{\sharp}P$. This is actually the principle behind most of the discrete methods described in the previous section. Now, with respect to \gls{ot} potentials, if $\Psi(P) = P$, then one has,
\begin{align*}
    \sum_{k=1}^{K}\lambda_{k}T_{Q_{k}\rightarrow P}^{\star}(z) = z \overset{(a)}{\implies} \sum_{k=1}^{K}\lambda_{k}\nabla g_{k}(z) = z \overset{(b)}{\implies} \sum_{k=1}^{K}\lambda_{k}g_{k}(z) = \dfrac{\lVert z \rVert_{2}^{2}}{2} + c
\end{align*}
where implication (a) follows from Brenier's theorem, and $g_{k}$ is the potential of the transportation problem from $Q_{k}$ to $P$, and implication (b) follows by integration. Here, since $c$ is an integration constant, one can conveniently take $c = 0$. Now, denote $\text{Congruent}(\lambda) = \{ g_{1:K} = (g_{1},\cdots,g_{K}): \sum_{k}\lambda_{k}g_{k}(z) = \nicefrac{\lVert z \rVert_{2}^{2}}{2} \}$ as the set of congruent potentials $g_{1:K} = (g_{1}, \cdots, g_{K})$ with respect to coordinates $\lambda$. Using the dual formulation in Equation~\ref{eq:dual_ot} and restricting the set of potentials to congruent potentials,
\begin{align*}
    \mathbb{B}(P|\mathcal{Q}) =\sum_{k=1}^{K}\lambda_{k} \mathbb{E}_{z\sim Q_{k}}\biggr[ \dfrac{\lVert z \rVert_{2}^{2}}{2} \biggr] + \mathbb{E}_{z \sim P}\biggr[ \dfrac{\lVert z \rVert_{2}^{2}}{2} \biggr] - \min{g_{1:K} \in \text{Congruent}(\lambda)}\biggr( \sum_{k=1}^{K}\lambda_{k}(\mathbb{E}_{z\sim Q_{k}}[f_{k}(z)] + \mathbb{E}_{z \sim P}[g_{k}])(z) \biggr),
\end{align*}
where $g_{1:K} = (g_{1},\cdots,g_{K})$. Now, taking the expectation on the congruence constraint, we have,
\begin{align*}
    \sum_{k=1}^{K}\lambda_{k}\mathbb{E}_{z\sim P}[g_{k}(z)] = \mathbb{E}_{z \sim P}\biggr[ \dfrac{\lVert z \rVert_{2}^{2}}{2} \biggr],
\end{align*}
which allowed~\cite{korotin2021continuous} to simplify $\mathbb{B}(P|\mathcal{Q})$ into,
\begin{align}
    \mathbb{B}(P|\mathcal{Q}) = \sum_{k=1}^{K}\lambda_{k} \mathbb{E}_{z\sim Q_{k}}\biggr[ \dfrac{\lVert z \rVert_{2}^{2}}{2} \biggr] - \min{g_{1:K} \in \text{Congruent}(\lambda)} \sum_{k=1}^{K}\lambda_{k}\mathbb{E}_{z\sim Q_{k}}[g_{k}(z)],\label{eq:cw2b_obj}
\end{align}
which does not involve terms with respect to the barycentric measure $P$. While the approach of~\cite{korotin2021continuous} simplifies the barycenter objective to expectations with respect to the known measures $Q_{k}$, it introduces a challenging constraint -- the congruence of potentials. This constraint is challenging, as it is a functional inequality over the whole space $\Omega$. Furthermore, it adds to the already constrained setting of fitting convex functions.~\cite{korotin2021continuous} proposes solving it \emph{approximately}, by penalizing potentials that do not meet the congruence condition. There is a circular dependence here, as one needs an estimate of the Wasserstein barycenter to estimate the constraint violation.

\noindent\textbf{Neural (Weak) Optimal Transport.} Weak \gls{ot}~\cite{gozlan2017kantorovich, backhoff2019existence} comes from the interpretation of the Kantorovich formulation as,
\begin{align}
    \gamma^{\star} = \arginf{\gamma \in \Gamma(P, Q)}\int_{\Omega}C(z,\gamma(\cdot | z))dP(z)\text{, where }C(z,Q) = \int_{\Omega} c(z,z')dQ(z').
\end{align}
This formulation allows for more general costs $C:\Omega\times\mathcal{P}(\Omega)\rightarrow\mathbb{R}$, i.e., costs that take a sample and a probability measure as inputs. In this sense,~\cite{kolesov2024estimating} extends Equation~\ref{eq:cw2b_obj} to
\begin{align}
    \mathbb{B}(P|\mathcal{Q}) = \sum_{k=1}^{K}\lambda_{k}\biggr( \mathbb{E}_{z \sim Q_{k}}[C_{k}(z, \gamma(\cdot|z))] - \mathbb{E}_{z \sim Q_{k}}[\mathbb{E}_{z' \sim \gamma(\cdot|z)}(g_{k}(z')) ] \biggr),\label{eq:not}
\end{align}
which holds for congruent potentials $\sum_{k}\lambda_{k}g_{k} = 0$. As with Equation~\ref{eq:cw2b_obj}, $\mathbb{B}$ does not depend explicitly on the unknown barycentric measure $P$. As such, the authors compute the barycenter through a max-min optimization problem,
\begin{align*}
    \max{\{g_{1:K}:\sum_{k}\lambda_{k}g_{k} = 0\}}\min{\{\gamma_{1},\cdots,\gamma_{K} : \gamma_{k}(\cdot|z) \in \Gamma(P_{k}) \}}\biggr\{ \sum_{k=1}^{K}\lambda_{k}\biggr( \mathbb{E}_{z \sim Q_{k}}[C_{k}(z, \gamma_{k}(\cdot|z))] - \mathbb{E}_{z \sim Q_{k}}[\mathbb{E}_{z' \sim \gamma_{k}(\cdot|z)}(g_{k}(z')) ] \biggr) \biggr\}.
\end{align*}
This extension opens up new possibilities for weak \gls{ot} barycenters. For instance,
\begin{align*}
    C_{\text{classical}}(x, \gamma) = \dfrac{1}{n}\sum_{j=1}^{n}c(z_{i}^{(P)},z_{j}^{(Q)})\quad C_{\text{KL}}(x,\gamma) = \dfrac{1}{n}\sum_{j=1}^{n}c(z_{i}^{(P)},z_{j}^{(Q)}) + \epsilon \text{KL}(\mathcal{N}(\mu(z), \sigma(z))|P_{0}),
\end{align*}
where $C_{\text{classical}}$ allows us to retrieve the original \gls{ot} problem. $C_{\text{KL}}$ is an alternative, when modeling $\gamma(\cdot|z) = \mathcal{N}(\mu(z), \sigma(z))$. On top of this reformulation, the authors in~\cite{kolesov2024estimating} propose approximating $\gamma(\cdot|z)$ through a mapping, as in~\cite{korotin2022neural}, that is,
\begin{align*}
    \gamma(\cdot|z) = T(z,s),\text{ subject to }s \sim \mathbb{S},
\end{align*}
where $\mathbb{S}$ is some prior measure, such as $\mathbb{S} = \mathcal{N}(0, \text{I})$. In this context,~\cite{gazdieva2024robust} extends this idea in the semi-unbalanced \gls{ot} setting~\cite{liero2018optimal}, where one of the mass preservation constraints on one of the marginals of $\gamma$ is relaxed. More specifically, given $\xi:\mathbb{R}_{+}\rightarrow\mathbb{R}_{+}$,
\begin{align*}
    D_{\xi}(P||Q) = \int_{\Omega}\xi\biggr( \dfrac{P(z)}{Q(z)} \biggr)dQ(z)\text{ if }P \ll Q\text{ and }+\infty\text{ otherwise},
\end{align*}
where $P\ll Q$ means that $P$ is absolutely continuous with respect to $Q$. We denote $\bar{\xi}(z') = \text{sup}_{z \in \mathbb{R}}\{ zz' - \xi(z) \}$ as its convex conjugate. With that in mind,~\cite{gazdieva2024robust} shows, in Theorem 1 and Corollaries 1 and 2, that minimizing
\begin{align*}
    \argmin{P\in\mathcal{P}_{2}}\sum_{k=1}^{K}\lambda_{k}\biggr( \int_{\Omega}\int_{\Omega} c(z,z')d\gamma(z,z') + D_{\xi}(\gamma(\cdot|z)||Q_{k}) \biggr),
\end{align*}
is equivalent to the following max-min problem,
\begin{align*}
    \max{\{g_{1:K}:\sum_{k}\lambda_{k}g_{k} = 0\}}\min{\{\gamma_{1},\cdots,\gamma_{K} : \gamma_{k}(\cdot|z) \in \Gamma(P_{k}) \}}\biggr\{ \sum_{k=1}^{K}\lambda_{k}\biggr( 
        \mathbb{E}_{z\sim Q_{k}}\biggr[ -\bar{\xi}_{k}\biggr( -\mathbb{E}_{z' \sim \gamma_{k}(\cdot|z)}[c_{k}(z, z') - g_{k}(z')] \biggr) \biggr]
    \biggr) \biggr\}.
\end{align*}

\noindent\textbf{Normalizing Flows.} This approach was proposed by~\cite{visentin2025computing}, and relies on the analysis of~\cite{gangbo1996geometry}. Their main idea is rewriting the Wasserstein distance in terms of functions $f, g$,
\begin{align}
    \mathbb{W}_{2}(P, Q)^{2} = \inf{f \in \text{Borel}(P_{0}, P),g\in\text{Borel}(P_{0},Q)}\lVert f - g  \rVert_{L_{2}(P_{0})}^{2},  \label{eq:gangbo}
\end{align}
where $\text{Borel}(P, Q)$ is the set of Borel functions such that $f_{\sharp}P = Q$. Naturally, a solution $(f^{\star}, g^{\star})$ to the previous problem gives the transport map $T = g^{\star}\circ f^{-1}$, where $f^{-1}$ is the inverse of $f^{\star}$. Furthermore, $P_{0}$ is an arbitrary latent measure, which further links the approach with generative modeling (i.e., take $P_{0} = \mathcal{N}(0, \text{I})$).

Based on Equation~\ref{eq:gangbo}, the insight of~\cite{visentin2025computing} is modeling $f$ and $g$ through a conditional function $f(z, s)$, $s \in \{0, 1\}$. The Wasserstein barycenter problem is a straightforward generalization of the two-measure case. For instance, the authors consider $S = \{1,\cdots,K\}$, which means that,
\begin{align*}
    \mathbb{B}(P|\mathcal{Q}) = \sum_{k=1}^{K}\lambda_{k}\mathbb{W}_{2}(P, Q_{k})^{2} = \sum_{k=1}^{K}\lambda_{k}\mathbb{W}_{2}(g_{\sharp}P_{0},Q_{k})^{2} = \sum_{k=1}^{K}\lambda_{k}\lVert g - f(z,k) \rVert_{L_{p}(P_{0})}^{2},
\end{align*}
where $g$ is the push-forward of $P_{0}$ to the barycenter $P^{\star}$, and $f(z,k)$ is the push-forward of $Q_{k}$ to $P^{\star}$.  As the authors show in~\cite[Theorem 3.3]{visentin2025computing},
\begin{align*}
    g(z) = \sum_{k=1}^{K}\lambda_{k}f(z,k).
\end{align*}
The main challenge in~\cite{visentin2025computing} is formulating a conditional model $f:\Omega\times S$. As the authors discuss, one possibility would be using $K = |S|$ different neural nets for modeling $f(z,\cdot)$, which is similar to the approach used in the neural-network methods. The authors take a different route and use conditional normalizing flows~\cite{kobyzev2020normalizing} to model these functions.

\subsection{Future Directions}

From the previous discussion on neural net-based methods for computing Wasserstein barycenters, we can identify a few common design choices. For instance,~\cite{korotin2021continuous,kolesov2024estimating} and~\cite{gazdieva2024robust} use $\mathcal{O}(K)$ neural nets for computing Wasserstein barycenters. This design choice results in a parametrization of the underlying barycenter that grows with the number of input measures. The only neural net-based method that differs from this idea is~\cite{visentin2025computing}, which uses a conditional model $f:\Omega\times S$.

Meanwhile, as we cover in our experiments (c.f. Section 4 in the main paper), there is a clear advantage tp incorporating labels in the ground-cost when these are available. Unfortunately, it is not clear how to integrate label information into the existing methods. For instance, one could consider using the continuous parametrization in terms of logits $\ell$, as we described in Section 3.3. However, there are a few caveats,
\begin{enumerate}
    \item Normalizing flows require invertible transformations. However, the softmax operation used to obtain the labels, i.e., $y = \text{softmax}(\ell)$, is not invertible.
    \item Using \glspl{icnn}, one can enforce the convexity of $z = (x, \ell)$. However, applying the softmax breaks the convexity requirement.
\end{enumerate}
As a result, adapting neural-net-based methods for the joint $\mathcal{X}\times\mathcal{Y}$ space is challenging and not straightforward. We leave this question for future work.

\section{Additional Background}
\subsection{Calculus in Wasserstein Spaces}\label{appx:calculus}

In this section, we consider $\Omega = \mathbb{R}^{d}$, $d(z,z') = \lVert z- z' \rVert_{2}$ and $p = 2$ to befixed. While our algorithm is more general than this setting, we use these restrictions as they are the most widely analyzed in the current literature. We leave the theoretical results for general metrics and $p$ to future work.

\textbf{Notation.} For the following theoretical analysis, we use the following notation. Given a measurable map $f,g:\Omega\rightarrow\Omega$ and a measure $P \in \mathcal{P}_{2}(\Omega)$, we denote,
\begin{align*}
    \lVert f \rVert_{L_{2}(P)}^{2} = \int_{\Omega} \lVert f(x) \rVert_{2}^{2}dP(x).
\end{align*}
Likewise, we have a notion of inner product,
\begin{align*}
    \langle f, g \rangle_{P} = \int_{\Omega} \langle f(x), g(x) \rangle_{\Omega}dP(x).
\end{align*}

\begin{definition}
    Given a functional $\mathbb{F}:\mathcal{P}_{2}(\Omega)(\mathbb{R}^{d})\rightarrow\mathbb{R}$, we call $\delta \mathbb{F}(P):\mathbb{R}^{d}\rightarrow\mathbb{R}$, if it exists, the unique (up to additive constants) function such that
    \begin{equation}
        \dfrac{d}{d\varepsilon}\mathbb{F}(P + \varepsilon \chi) \biggr{|}_{\epsilon = 0} = \int \delta \mathbb{F}(P)d\chi,\label{eq:first-variation}
    \end{equation}
    for every perturbation $\chi$ such that, at least for $\varepsilon \in [0, \varepsilon_{0}]$, the measure $P + \varepsilon\chi \in \mathcal{P}_{2}(\mathbb{R}^{d})$. This functional is called \emph{the first variation} of $\mathbb{F}$ at $P$.
\end{definition}

We refer readers to~\cite[Chapter 5]{chewi2024statistical} and~\cite[Chapter 8]{ambrosio2008gradient} for further details on gradient flows in Wasserstein spaces. The main tool we will use here is the continuity equation,
\begin{align*}
    \partial_{t}P_{t} = -\text{div}(P_{t}v_{t}),
\end{align*}
where the divergence operator should be understood in the measure-theoretic sense. Here, $v_{t}:\Omega\rightarrow\Omega$ is a vector field. Now, given $\mathbb{F}:\mathcal{P}_{2}(\Omega)\to\mathbb{R}$ with first variation $\delta \mathbb{F}(P):\Omega\to\mathbb{R}$, the time derivative of $\mathbb{F}$ along the curve $\{ P_{t} \}_{t\geq 0}$ can be computed as,
\begin{align*}
    \partial_{t}\mathbb{F}(P_{t}) \overset{(1)}{=} \int \delta\mathbb{F}(P_{t})d(\partial_{t}P_{t}) \overset{(2)}{=} -\int \delta\mathbb{F}(P_{t})\text{div}(P_{t}v_{t})dz \overset{(3)}{=} \int\langle \nabla\delta\mathbb{F}(P_{t}),v_{t} \rangle dP_{t} \overset{(4)}{=} \langle \nabla\delta\mathbb{F}(P_{t}), v_{t} \rangle_{P_{t}},
\end{align*}
where (1) is obtained by taking $\chi = \partial_{t}P_{t}$ in Equation~\ref{eq:first-variation}; (2) is obtained by applying the continuity equation; (3) is obtained by using integration by parts; finally, (4) is obtained via the definition of inner product. This motivates the equality,
\begin{align}
    \gradW \mathbb{F}(P) = \nabla\delta\mathbb{F}(P).\label{eq:grad_W}
\end{align}
We refer readers to~\cite[Proposition 5.10]{chewi2024statistical}, for a more formal derivation. This discussion characterizes the Wasserstein gradient $\gradW \mathbb{F}(P):\Omega\to\Omega$ as a vector field. Therefore, taking the direction of steepest descent of the functional $\mathbb{F}$, $v_{t} = -\gradW \mathbb{F}(P_{t})$, leads to the following connection between $\partial_{t}\mathbb{F}(P_{t})$ and $\gradW \mathbb{F}(P_{t})$,
\begin{align}
    \partial_{t}\mathbb{F}(P_{t}) = -\lVert \gradW \mathbb{F}(P_{t}) \rVert_{L_{2}(P_{t})}^{2},\label{eq:wgf_steepest_descent}
\end{align}

\subsection{Wasserstein Gradient of Functionals}\label{appx:gradients}

In this section, we give an overview of the Wasserstein gradient of the terms used in Equation 16. We refer readers to~\cite[Chapter 5]{chewi2024statistical} and~\cite[Section 4]{santambrogio2017euclidean} for further details.

\textbf{Potential Energy, $\mathbb{V}$.} For reference, this functional corresponds to
\begin{align*}
    \mathbb{V}(P) = \mathbb{E}_{z \sim P}[V(z)] = \int V(z) dP(z),
\end{align*}
for some potential function $V:\Omega\rightarrow \mathbb{R}$. From the definition of the first variation,
\begin{align*}
    \mathbb{V}(P + \epsilon\chi) = \int V(z)dP(z) + \epsilon\int V(z) d\chi(z),
\end{align*}
then, taking the derivative with respect to $\epsilon$,
\begin{align*}
    \dfrac{d}{d\epsilon}\mathbb{V}(P + \epsilon \chi) = \int V(z)d\chi,
\end{align*}
which means that $V = \delta\mathbb{V}(P)$. As a consequence, the Wasserstein gradient of $\mathbb{V}$ is,
\begin{align*}
    \gradW \mathbb{V} = \nabla V.
\end{align*}
which is a vector field over $\Omega$. Since $\mathbb{V}$ is linear in $P$, it is convex. Furthermore, its continuity with respect to the weak convergence depends on the regularity of $V$. For instance, if $V$ is bounded, $\mathbb{V}$ is continuous. If $V$ is lower semi-continuous and bounded from below, then $\mathbb{V}$ is semi-continuous.

\textbf{Internal Energy, $\mathbb{G}$.} The internal energy takes the form,
\begin{align*}
    \mathbb{G}(P) = \int G(P(z))dz,
\end{align*}
where $P(z)$ is understood as the density of the measure $P$, with respect to the Lebesgue measure. Let us assume $\chi$ admits a density with respect the Lebesgue measure as well. In this case,
\begin{align*}
    \mathbb{G}(P + \epsilon \chi) = \int G(P(z) + \epsilon \chi(z))dz.
\end{align*}
Taking the derivative under the integral sign,
\begin{align*}
    \dfrac{d}{d\epsilon}\mathbb{G}(P + \epsilon\chi)\biggr{|}_{\epsilon = 0} = \int G'(P(z))\chi(z)dz = \int G'(P(z))d\chi(z),
\end{align*}
which means that $\delta \mathbb{G} = G'(P)$ and $\gradW \mathbb{G} = \nabla G'(P(z))$.

\textbf{Interaction Energy, $\mathbb{U}$.} Here, let us assume $U(z, z')$ is symmetric for simplicity,
\begin{align*}
    \mathbb{U}(P) = \int \int U(z,z')dP(z)dP(z'),
\end{align*}
hence,
\begin{align*}
    \mathbb{U}(P+\epsilon\chi) = \int\int U(z,z')dP(z)dP(z') + 2\epsilon\int\int U(z,z')dP(z)d\chi(z') +\mathcal{O}(\epsilon^{2}).
\end{align*}
Again, taking derivatives with respect to $\epsilon$,
\begin{align*}
    \dfrac{d}{d\epsilon}\mathbb{U}(P + \epsilon\chi)\biggr{|}_{\epsilon = 0} = 2\int\biggr(\int U(z,z')dP(z)\biggr)d\chi(z'),
\end{align*}
which means that,
\begin{align*}
    \delta\mathbb{U}(P) = 2\int U(z,z')dP(z).
\end{align*}
When $U(z,z') = h(z-z')$ for an even $h$, $\int U(z,z')dP(z) = \int h(z-z')P(z)dz = h\star P$, i.e., the convolution of $h$ and $P$. Furthermore, $\gradW \mathbb{U} = (\nabla h) \star P$.

\textbf{Barycenter Functional, $\mathbb{B}$.} For the barycenter functional, we can derive its first variation from the $2-$Wasserstein distance. Assume there exists an \gls{ot} map $T_{P\rightarrow Q}^{\star}$. In this case,
\begin{align}
    \gradW \mathbb{W}_{2}(P, Q)^{2} = 2(\text{Id} - T_{P\rightarrow Q}^{\star}).\label{eq:wasserstein_grad_w2}
\end{align}
Therefore, given $\mathcal{Q} = \{Q_{1},\cdots,Q_{K}\}$, it is straightforward to obtain:
\begin{align}
    \gradW \mathbb{B}(P|\mathcal{Q}) &= \gradW \biggr(\sum_{k=1}^{K}\lambda_{k}\mathbb{W}_{2}(P, Q_{k})^{2}\biggr) = 2\sum_{k=1}^{K}\lambda_{k}(\text{Id} - T_{P\rightarrow Q_{k}}^{\star}).\label{eq:wasserstein_grad_B}
\end{align}

\noindent\textbf{A Connection with~\cite[Algorithm 2]{cuturi2014fast}.} Let us now focus on the empirical case with an Euclidean ground-cost. Assume $\gamma^{\star}$ computed between $\hat{P}$ and $\hat{Q}$, which means that,
\begin{align*}
    \mathbb{W}_{2}(\hat{P}, \hat{Q})^{2} = \sum_{i=1}^{n}\sum_{j=1}^{m}\gamma_{ij}^{\star}\lVert z_{i}^{(P)} - z_{j}^{(Q)} \rVert_{2}^{2}\text{, and }\mathbb{B}(\hat{P}_{\tau}|\mathcal{Q}) = \sum_{k=1}^{K}\lambda_{k}\sum_{i=1}^{n}\sum_{j=1}^{n_{k}}\gamma_{k,i,j}^{\star}\lVert z_{\tau,i}^{(P)} - z_{j}^{(Q_{k})} \rVert_{2}^{2}.
\end{align*}
We can compute the Wasserstein gradient $\gradW \mathbb{B}$ by computing $\nabla_{z_{\tau,i}^{(P)}}\mathbb{B}$, that is,
\begin{align*}
    \nabla_{z_{\tau,i}^{(P)}}\mathbb{B}(\hat{P}_{\tau}|\mathcal{Q}) = 2\sum_{k=1}^{K}\lambda_{k}\sum_{j=1}^{n_{k}}\gamma_{k,i,j}^{\star}(z_{\tau,i}^{(P)} - z_{j}^{(Q_{k})}) = \dfrac{2}{n}z_{\tau,i}^{(P)}-2\sum_{k=1}^{K}\lambda_{k}\sum_{j=1}^{K}\gamma_{k,i,j}^{\star}z_{j}^{(Q_{k})}.
\end{align*}
From this last equation, we can use the barycentric mapping definition,
\begin{align}
    T_{P\rightarrow Q_{k}}^{\star}(z_{i,\tau}^{P}) = n\sum_{j=1}^{n_{k}}\gamma_{k,i,j}^{\star}z_{j}^{(Q_{k})},\label{eq:barycentric_mapping}
\end{align}
to write,
\begin{align*}
    \nabla_{z_{\tau,i}^{(P)}}\mathbb{B}(\hat{P}_{\tau}|\mathcal{Q}) = \dfrac{2}{n}\biggr( z_{\tau,i}^{(P)} - \sum_{k=1}^{K}\lambda_{k}T_{P\rightarrow Q_{k}}^{\star}(z_{\tau,i}^{(P)}) \biggr),
\end{align*}
which means we can write the gradient flow update rule as,
\begin{align*}
    z_{\tau+1,i}^{(P)} = \biggr(1 - \dfrac{2\alpha}{n}\biggr)z_{i,\tau}^{(P)} + \dfrac{2\alpha}{n}\sum_{k=1}^{K}\lambda_{k}T_{P\rightarrow Q_{k}}^{\star}(z_{i,\tau}^{(P)}),
\end{align*}
which is the update rule in~\cite[Algorithm 2]{cuturi2014fast} with $\theta = \nicefrac{2\alpha}{n}$. Furthermore, we can also instantiate the fixed-point algorithm of~\cite{montesuma2023multi}, for $\alpha = n/2$. We now highlight two differences of our work with respect to these two strategies. First, our setting is more general. In fact, our algorithm is capable of handling ground metrics other than the squared Euclidean metric. Furthermore, our framework couples the barycenter objective with other functionals. Second, while~\cite{cuturi2014fast} assume that \emph{all samples} of $\mathcal{Q} = \{Q_{1},\cdots,Q_{K}\}$ are available at once, our algorithm assumes that these measures are available through sampling. This treatment is often called \emph{continuous} in the literature~\cite{korotin2021continuous,korotin2022wasserstein}; however, one still relies on samples from these measures rather than some truly continuous form.

\newpage

\section{Proof of Theorem 3.1}\label{appx:proof}

In this section, we present the proof for Theorem 3.1. We mainly use two tools to obtain this result,
\begin{enumerate}
    \item The exponential convergence given by the \gls{pl} inequality,
    \begin{align}
        \lVert \gradW \mathbb{B}(P) \rVert_{L_{2}(P)}^{2} \geq C_{\text{PL}} (\mathbb{B}(P) - \mathbb{B}^{\star}),\label{eq:pl_ineq}
    \end{align}
    \item A uniform bound on $P \mapsto \mathbb{B}(P|\mathcal{Q})$ controlling the error of approximating $Q_{k}$ with $\hat{Q}_{k}$.
\end{enumerate}
These results are available, respectively, in~\cite{chewi2024statistical} and~\cite{portales2025sample}. For completeness, we re-state and demonstrate them. We start with a modified version of~\cite[Corollary 5.17]{chewi2024statistical},
\begin{proposition}{(Measure theoretic Gronwall's Lemma~\cite{chewi2024statistical})}
    Let $\mathbb{F}:\mathcal{P}_{2}(\Omega)\rightarrow\mathbb{R}$ be a functional that satisfies the \gls{pl} inequality (c.f.~\ref{eq:pl_ineq}) with constant $C_{\text{PL}} > 0$, then,
    \begin{align}
        \mathbb{F}(P_{\tau}) - \mathbb{F}^{\star} \leq e^{-2C_{\text{PL}}\tau}(\mathbb{F}(P_{0}) - \mathbb{F}^{\star}).
    \end{align}
\end{proposition}
\begin{proof}
    Define $\phi(t) = \mathbb{F}(P_{t}) - \mathbb{F}^{\star}$. Then,
    \begin{align*}
        \partial_{t}\phi(t) &= \partial_{t}\mathbb{F}(P_{t}) = -\lVert \gradW \mathbb{F}(P_{t}) \rVert_{L_{2}(P_{t})}^{2},\\
        &\geq -C_{\text{PL}}(\mathbb{F}(P_{0}) - \mathbb{F}^{\star}) = -C_{\text{PL}}\phi(t).
    \end{align*}
    From Gr\"onwall's Lemma~\cite[Lemma 5.16]{chewi2024statistical},
    \begin{align*}
         \underbrace{\mathbb{F}(P_{t}) - \mathbb{F}^{\star}}_{\partial_{t}\phi(t)} \leq e^{-C_{\text{PL}}}\underbrace{(\mathbb{F}(P_{0}) - \mathbb{F}(P)^{\star})}_{\phi(0)}
    \end{align*}
\end{proof}

We now present the uniform approximation bound of~\cite{portales2025sample} for empirical barycenters,
\begin{theorem}{(Barycenter Empirical Approximation error~\cite{portales2025sample})}\label{thm:portales}
    Let $\mathcal{Q} = \{Q_{1},\cdots,Q_{K}\}$ be a finite family of measures on $\mathcal{P}_{2}(\mathcal{B}(0, R))$ for $R > 0$. Let each $Q_{k}$ be approximated with $m$ i.i.d. samples $\{z_{i}^{(Q_{k})}\}_{i=1}^{m}$. Let $\hat{P} = n^{-1}\sum_{i=1}^{n}\delta_{z_{i}^{(P)}}$. Let $p \in [1, +\infty)$ and $\epsilon \geq 0$. Then,
    \begin{align}
        \mathbb{E}\biggr[ \sup{z_{1}^{(P)},\cdots,z_{n}^{(P)} \in \Omega}|\mathbb{B}_{\epsilon}(\hat{P}|\mathcal{Q}) - \widehat{\mathbb{B}}_{\epsilon}(\hat{P}|\mathcal{Q})| \biggr] \leq C_{p, R}\sqrt{\dfrac{C_{d,n}}{m}},\label{eq:approx_portales}
    \end{align}
    where,
    \begin{align*}
        C_{p,R} &= \begin{cases}
            8\sqrt{2}\int_{0}^{3(2R)^{p}}\sqrt{\log\biggr( 2 + \dfrac{32p(2R)^{p}}{s} \biggr)}ds & \epsilon = 0,\\
            8\sqrt{2}\int_{0}^{(4p+1)(2R)^{p}}\sqrt{\log(2+\frac{64p(2R)^{p}}{s})}ds & \epsilon > 0,
        \end{cases},\\
        C_{d,n} &= n(d+1).
    \end{align*}
\end{theorem}

With these bounds established, we can then prove Theorem 3.1.
\begin{proof}
    Let $\tau > 0$ be the current optimization step. We have the following,
    \begin{align}
        \underbrace{\hat{\mathbb{B}}(\hat{P}_{\tau}) - \hat{\mathbb{B}}^{\star}}_{\text{emp. optimization gap}} = \underbrace{(\hat{\mathbb{B}}(\hat{P}_{\tau}) - \mathbb{B}(\hat{P}_{\tau}))}_{\text{emp. approx. gap}} +  \underbrace{(\mathbb{B}(\hat{P}_{\tau}) - \mathbb{B}^{\star})}_{\text{true optimization gap}} + \underbrace{(\mathbb{B}^{\star}-\hat{\mathbb{B}}^{\star})}_{\text{emp. approx. gap}}.\label{eq:pre-bound}
    \end{align}
    The optimization gap reflects how far we are from $\text{inf }\mathbb{B}(\hat{P})$ (resp. $\hat{\mathbb{B}}$), whereas the empirical approximation gap reflects the inherent errors of approximating each $Q$ by $\hat{Q}_{k}$. \emph{In the following, we take expectations with respect to samples from $Q_{k}$}.

    From Equation~\ref{eq:pre-bound}, we can bound the expected empirical optimization gap,
    \begin{align}
        \mathbb{E}[|\hat{\mathbb{B}}(\hat{P}_{\tau}) - \hat{\mathbb{B}}^{\star}|] \leq& \mathbb{E}[|\hat{\mathbb{B}}(\hat{P}_{\tau}) - \mathbb{B}(\hat{P}_{\tau})|] +  \mathbb{E}[|\mathbb{B}(\hat{P}_{\tau}) - \mathbb{B}^{\star}|] +  \mathbb{E}[| \mathbb{B}^{\star} - \hat{\mathbb{B}}^{\star} |].\label{eq:bound1}
    \end{align}
    Now, we can bound the first and third terms using the approximation error of \cite{portales2025sample} (i.e., Equation~\ref{eq:approx_portales}),
\begin{equation}
    \begin{aligned}
        \mathbb{E}[|\hat{\mathbb{B}}(\hat{P}_{\tau}) - \mathbb{B}(\hat{P}_{\tau})|] &\leq C_{p, R}\sqrt{\dfrac{C_{d,n}}{m}},\\
        \mathbb{E}[|\hat{\mathbb{B}}^{\star} - \mathbb{B}^{\star}|] &\leq C_{p, R}\sqrt{\dfrac{C_{d,n}}{m}}.
    \end{aligned}\label{eq:bound_endpoints}
\end{equation}
    Here, we recall that $\mathbb{B}^{\star} = \mathbb{B}(P^{\star}) = \text{inf}_{P}\mathbb{B}(P)$. Returning to the second term, note that the true optimization gap $\mathbb{B}(\hat{P}_{\tau}|\mathcal{Q}) - \mathbb{B}^{\star}$ is deterministic, as it does not involve sampling from the measures in $\mathcal{Q}$. This means that,
    \begin{equation}
        \begin{aligned}
            \mathbb{E}[|\mathbb{B}(\hat{P}_{\tau}) - \mathbb{B}^{\star}|] = |\mathbb{B}(\hat{P}_{\tau}) - \mathbb{B}^{\star}| \leq e^{-C_{\text{PL}}\tau}(\hat{\mathbb{B}}(\hat{P}_{0}) - \hat{\mathbb{B}}^{\star}).
        \end{aligned}\label{eq:empirical_pl}
    \end{equation}
    Combining Equations~\ref{eq:bound1},~\ref{eq:bound_endpoints}, and~\ref{eq:empirical_pl}, we get,
    \begin{align*}
        \mathbb{E}[|\hat{\mathbb{B}}(\hat{P}_{\tau}) - \hat{\mathbb{B}}^{\star}|] \leq e^{-C_{\text{PL}}\tau}(\mathbb{B}(\hat{P}_{0}) - \mathbb{B}^{\star}) + C_{p,R}\sqrt{\dfrac{C_{d,n}}{m}},
    \end{align*}
    which is the desired inequality.
\end{proof}

\subsection{Empirical Verification of the PL Inequality}\label{appx:PLEmpiricalValidation}

\begin{figure}[ht]
    \centering
    \includegraphics[width=\linewidth]{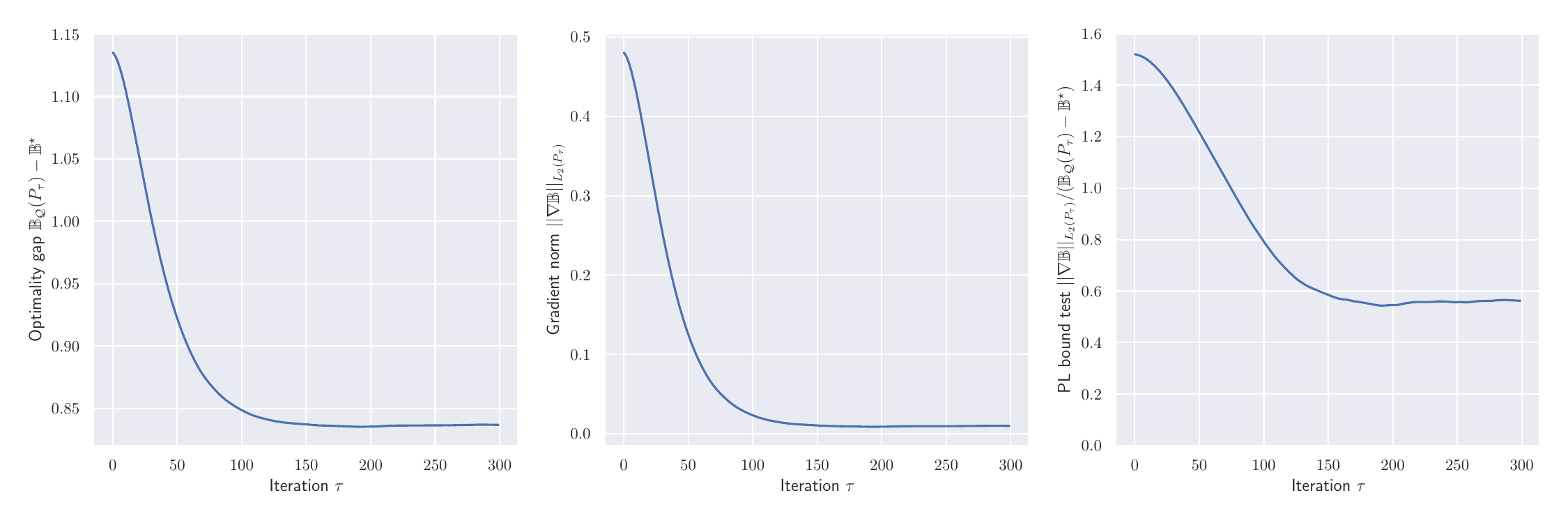}
    \caption{Empirical test of the \gls{pl} inequality. From left to right, we show: (i) the optimality gap, (ii) the gradient norm, and (iii) the \gls{pl} ratio as specified in Equation~\ref{eq:pl-ratio}.}
    \label{fig:pl-ratio-test}
\end{figure}

An open question in the literature is the verification of the \gls{pl} inequality (Equation~\ref{eq:pl_ineq}) for general measures, such as empirical ones. To give further context, current results prove that it indeed holds for Gaussian measures, under specific spectral conditions~\cite{chewi2020gradient}. Therefore, we do an empirical validation of whether this bound holds in the empirical case. In a nutshell, to test Equation~\ref{eq:pl_ineq}, we need to compute the \gls{pl} ratio,
\begin{align}
    \text{PL}_{\text{ratio}}(\hat{P}_{\tau}) = \dfrac{\lVert \gradW \mathbb{B}(\hat{P}_{\tau}) \rVert_{L_{2}(\hat{P}_{\tau})}}{\mathbb{B}(\hat{P}_{\tau}) - \mathbb{B}^{\star}},\label{eq:pl-ratio}
\end{align}
and verify that it indeed remains strictly positive over the course of our optimization. Computing the \gls{pl} ratio in Equation~\ref{eq:pl-ratio} requires knowledge of $\mathbb{B}^{\star}$, which is, in general, intractable. However, for location-scatter families, it is known in closed form, since $P^{\star}$ itself is known. Therefore, we verify it in the context of the Swiss-roll measure, where all quantities are known. We show our results in Figure~\ref{fig:pl-ratio-test}, which indicates that the ratio converges with iterations to approximately $0.56$.

\subsection{PL Inequality on Location-Scatter Family}
\label{appx:pl_theory}

An open question in the literature is the verification of the PL inequality (Equation~\ref{eq:pl_ineq}) for general measures. Current results establish that it holds for Gaussian measures under specific spectral conditions~\citep{chewi2020gradient}. In this section, we show that the PL inequality \emph{transfers} from Gaussians to the broader class of \emph{location-scatter families}, which includes our Swiss roll experiments (Section~\ref{sec:experiments}).

\noindent\textbf{Location-Scatter Families.} Following \citet{alvarez2016fixed}, let $Q_0 \in \mathcal{P}_{2}(\mathbb{R}^d)$ be a measure with mean $\mu_{0}$ and covariance $\Sigma_{0}$. The \emph{location-scatter family} generated by $Q_0$ is,
\begin{equation}
    \mathcal{L}(Q_{0}) := \left\{ T_{\sharp}Q_{0} : T(x) = Ax + m, b \in \mathbb{R}^d, A \in \mathcal{S}^{++}_d \right\},
    \label{eq:loc_scatter_family}
\end{equation}
where $\mathcal{S}^{++}_d$ denotes the cone of $d \times d$ symmetric positive definite matrices. Each $Q \in \mathcal{L}(Q_0)$ has mean $b$ and covariance $A$, but inherits the higher-order structure of~$Q_0$.

\noindent\textbf{Isometry to the Bures-Wasserstein Manifold.} The key insight comes from \citet[Theorem~2.3]{alvarez2016fixed}: for any $Q_{1}, Q_{2} \in \mathcal{L}(Q_0)$, the 2-Wasserstein distance equals
\begin{equation}
    \mathbb{W}_2^2(Q_{1}, Q_{2}) = \|\mu_{1} - \mu_{2} \|^2 + \text{Tr}(\Sigma_{1}+\Sigma_{2}-2(\Sigma_{2}^{1/2}\Sigma_{1}\Sigma_{2}^{1/2})^{1/2}),
    \label{eq:w2_loc_scatter}
\end{equation}
This is precisely the Wasserstein distance on the Bures-Wasserstein manifold: $\mathcal{BW} = \{\mathcal{N}(\mu,\Sigma): \mu \in \mathbb{R}^{d}, \Sigma \in \mathcal{S}_{d}^{++}\}$~\citep{takatsu2011wasserstein}. Henceforth, we denote the elements of $\mathcal{BW}$ by $G_{\mu,\Sigma}$. In that sense, $G_{\mu_{1},\Sigma_{1}}$ and $G_{\mu_{2},\Sigma_{2}}$ denote the Gaussian measures matching the moments of $Q_{1}, Q_{2} \in \mathcal{L}(Q_{0})$. As a consequence, the restriction of $\mathbb{W}_2$ to $\mathcal{L}(Q_0)$ is \emph{isometric} to the Bures-Wasserstein manifold. Moreover, the optimal transport map between elements of $\mathcal{L}(Q_0)$ takes the same form as for Gaussians:
\begin{equation}
    T^\star_{Q_{1} \to Q_{2} }(x) = \mu_2 + A(x - \mu_1), \quad A = \Sigma_1^{-1/2}\left(\Sigma_1^{1/2}\Sigma_2\Sigma_1^{1/2}\right)^{1/2}\Sigma_1^{-1/2}.
    \label{eq:ot_map_loc_scatter}
\end{equation}

\noindent\textbf{Barycenter Functional Equivalence.} Let $\mathcal{Q} = \{Q_k\}_{k=1}^K \subset \mathcal{L}(Q_0)$ with $Q_k = T_{k}Q_{0}$, such that $\mu_{k}$ and $\Sigma_{k}$ are the mean and covariance of $Q_{k}$. Let $\mathcal{G} = \{ G_{\mu_{k},\Sigma_{k}} \}_{k=1}^{K}$ be the corresponding Gaussian measures. Equation~\ref{eq:w2_loc_scatter} establishes:
\begin{align}
    \mathbb{B}(P|\mathcal{Q}) = \mathbb{B}(G_{\mu,\Sigma}|\mathcal{G})\text{, where }P\in\mathcal{L}(Q_{0})\text{, and }G_{\mu,\Sigma} \in \mathcal{BW}, \label{eq:functional_equivalence}
\end{align}
where $\mu, \Sigma$ are the mean and covariance of $P \in \mathcal{L}(Q_{0})$. Furthermore, by \citet[Corollary~4.5]{alvarez2016fixed}, the barycenter $P^\star \in \mathcal{L}(Q_0)$ corresponds to the Bures-Wasserstein barycenter: if $P^{\star}$ is the minimizer of $P \mapsto \mathbb{B}(P|\mathcal{Q})$ over $\mathcal{L}(Q_{0})$ with mean $\mu^{\star}$ and covariance $\Sigma^{\star}$, then $G_{\mu^{\star}, \Sigma^{\star}}$ is the minimizer of $G_{\mu,\Sigma} \mapsto \mathbb{B}(G_{\mu,\Sigma}|\mathcal{G})$.

\noindent\textbf{Wasserstein Gradient Structure.} The Wasserstein gradient of $\mathbb{B}$ at $P$ with mean $\mu$ and covariance $\Sigma$ is (cf.\ Equation~\ref{eq:wasserstein_grad_B})
\begin{equation}
    \gradW \mathbb{B}(P|\mathcal{Q}) = 2\sum_{k=1}^K \lambda_k \left(\mathrm{Id} - T^\star_{P \to Q_k}\right).
    \label{eq:wass_grad_bary}
\end{equation}
Since each $T^\star_{P \to Q_k}$ is affine (Equation~\ref{eq:ot_map_loc_scatter}), the gradient $\gradW \mathbb{B}(P|\mathcal{Q})$ is itself an affine map:
\begin{align}
    \gradW \mathbb{B}(P|\mathcal{Q})(x) &= 2\sum_{k=1}^{K}\lambda_{k}(x - (A_{k}x + b_{k})),\nonumber \\
    &= 2\biggr(x - (\sum_{k=1}^{K} \lambda_{k}A_{k})(x) + \sum_{k=1}^{K}\lambda_{k}b_{k}\biggr),\nonumber\\
    &= \underbrace{\biggr(2(\text{Id} - \sum_{k=1}^{K}\lambda_{k}A_{k})\biggr)}_{M}(x) + \underbrace{2\sum_{k=1}^{K}\lambda_{k}b_{k}}_{c},\nonumber\\
    &= Mx + c,\label{eq:grad_affine}
\end{align}
where $M = 2(\text{Id} - \bar{A})$, $c = 2(\bar{A}\mu - \bar{\mu})$, where $\bar{A} = \sum_{k=1}^{K}\lambda_{k}A_{k}$ and $\bar{\mu} = \sum_{k=1}^{K}\lambda_{k}\mu_{k}$. More importantly, by Equation~\eqref{eq:ot_map_loc_scatter}, this is \emph{the same affine map} as for the Gaussian case, resulting in,
\begin{equation}
    \gradW \mathbb{B}(P|\mathcal{Q}) = \gradW \mathbb{B}(G_{\mu,\Sigma}|\mathcal{G}) \quad \text{as vector fields } \mathbb{R}^d \to \mathbb{R}^d.
    \label{eq:grad_function_equality}
\end{equation}

We now establish that the $L^2$-norms of these gradients are also equal, which is the key step for transferring the PL inequality.

\begin{lemma}[$L^2$-norm of Affine Maps]
\label{lem:l2_affine}
Let $T(x) = Mx + c$ be an affine map with $M \in \mathbb{R}^{d \times d}$ and $c \in \mathbb{R}^d$. For any probability measure $P \in \mathcal{P}_2(\mathbb{R}^d)$ with mean $\mu_P$ and covariance $\Sigma_P$,
\begin{equation}
    \| T \|_{L^2(P)}^2 := \int_{\mathbb{R}^d} \| T(x) \|^2 \, dP(x) = \text{Tr}(M^{T} M \Sigma_P) + \|M\mu_P + c\|^2.
    \label{eq:l2_affine}
\end{equation}
In particular, $\| T \|_{L^2(P)}^2$ depends on $P$ only through its first two moments.
\end{lemma}

\begin{proof}
Expanding the squared norm:
\begin{align*}
    \| T \|_{L^2(P)}^2 &= \int \|Mx + c\|^2 \, dP(x) \\
    &= \int \|M(x - \mu_P) + (M\mu_P + c)\|^2 \, dP(x) \\
    &= \int \|M(x - \mu_P)\|^2 \, dP(x) + 2\langle M\mu_P + c, M\underbrace{\int (x - \mu_P) \, dP(x)}_{= 0} \rangle + \|M\mu_P + c\|^2 \\
    &= \int (x - \mu_P)^{T} M^{T} M (x - \mu_P) \, dP(x) + \|M\mu_P + c\|^2 \\
    &= \text{Tr}(M^{T} M \Sigma_P) + \|M\mu_P + c\|^2,
\end{align*}
where the last step uses $\mathbb{E}_{x \sim P}[(x - \mu_P)(x - \mu_P)^\top] = \Sigma_P$.
\end{proof}

\begin{lemma}[Gradient Norm Equality]
\label{lem:grad_norm_equality}
For any $P \in \mathcal{L}(Q_0)$ and its corresponding Gaussian $G_{\mu,\Sigma}$:
\begin{equation}
    \| \gradW \mathbb{B}(P|\mathcal{Q})\|_{L^2(P)}^2 = \| \gradW \mathbb{B}(G|\mathcal{G}) \|_{L^2(G_{\mu,\Sigma})}^2.
    \label{eq:grad_norm_equality}
\end{equation}
\end{lemma}

\begin{proof}
First, define $T_{\text{LS}} = \gradW \mathbb{B}(P|\mathcal{Q})$ and $T_{\text{Gaussian}} = \gradW \mathbb{B}(G_{m,\Sigma}|\mathcal{G})$. By construction, both of these mappings are affine, and by Equation~\ref{eq:grad_function_equality} $T_{\text{LS}} = T_{\text{Gaussian}}$. From Lemma~\ref{lem:l2_affine}, since $P$ and $G_{\mu,\Sigma}$ share the first and second order moments by construction, they are equal, which means,
\begin{align*}
    \lVert \gradW \mathbb{B}(P|\mathcal{Q}) \rVert_{L_{2}(P)}^{2} = T_{\text{LS}} = T_{\text{Gaussian}} = \lVert \gradW \mathbb{B}(G|\mathcal{G}) \rVert_{L_{2}(G_{\mu,\Sigma})}^{2},
\end{align*}
which is the desired result.
\end{proof}

We can now state and prove the following result.

\begin{proposition}[PL Inequality Transfer]
\label{prop:pl_transfer}
Let $\mathcal{L}(Q_0)$ be a location-scatter family and $\mathcal{Q} = \{Q_k\}_{k=1}^K \subset \mathcal{L}(Q_0)$. Let $\mathcal{G} = \{G_{\mu_{k},\Sigma_{k}}\}_{k=1}^K$ be the corresponding Gaussians with matching moments. If the Bures-Wasserstein barycenter functional $\mathbb{B}(\cdot|\mathcal{G})$ satisfies the PL inequality with constant $C_{\text{PL}} > 0$, i.e.,
\begin{equation}
    \| \gradW \mathbb{B}(G_{\mu,\Sigma}|\mathcal{G})\|_{L^2(G_{\mu,\Sigma})}^2 \geq C_{\text{PL}} \left( \mathbb{B}(G_{\mu,\Sigma}|\mathcal{G}) - \mathbb{B}(G_{\mu^\star,\Sigma^\star}|\mathcal{G}) \right),
    \label{eq:pl_gaussian}
\end{equation}
then $\mathbb{B}(\cdot|\mathcal{Q})$ satisfies the PL inequality on $\mathcal{L}(Q_0)$ with the same constant,
\begin{equation}
    \| \gradW \mathbb{B}(P|\mathcal{Q})\|_{L^2(P_{m,\Sigma})}^2 \geq C_{\mathrm{PL}} \left( \mathbb{B}(P|\mathcal{Q}) - \mathbb{B}(P^{\star}|\mathcal{Q}) \right).
    \label{eq:pl_loc_scatter}
\end{equation}
\end{proposition}

\begin{proof}
Starting from the PL inequality for Gaussians (Equation~\ref{eq:pl_gaussian}):
\begin{equation*}
    \| \gradW \mathbb{B}(G_{\mu,\Sigma}|\mathcal{G})\|_{L^2(G_{\mu,\Sigma})}^2 \geq C_{\text{PL}} \left( \mathbb{B}(G_{\mu,\Sigma}|\mathcal{G}) - \mathbb{B}(G_{\mu^\star,\Sigma^\star}|\mathcal{G}) \right).
\end{equation*}
Applying Lemma~\ref{lem:grad_norm_equality} to the left-hand side:
\begin{equation*}
    \| \gradW \mathbb{B}(P | \mathcal{Q})\|_{L^2(P)}^2 \geq C_{\text{PL}} \left( \mathbb{B}(G_{\mu,\Sigma}|\mathcal{G}) - \mathbb{B}(G_{\mu^\star,\Sigma^\star}|\mathcal{G}) \right).
\end{equation*}
Applying Equation~\eqref{eq:functional_equivalence} and barycenter correspondence to the right-hand side:
\begin{equation*}
    \| \gradW \mathbb{B}(P|\mathcal{Q})\|_{L^2(P)}^2 \geq C_{\mathrm{PL}} \left( \mathbb{B}(P|\mathcal{Q}) - \mathbb{B}(P^\star|\mathcal{Q}) \right). \qedhere
\end{equation*}
\end{proof}

\noindent\textbf{Explicit PL Constant.} By \citet[Theorem~19]{chewi2020gradient}, the PL inequality holds for Gaussians under $\zeta$-regularity: if $\|\Sigma_k\|_{\mathrm{op}} \leq 1$ and $\det \Sigma_k \geq \zeta$ for all $k$, then $C_{\mathrm{PL}} = \zeta^2/4$. Then, Proposition~\ref{prop:pl_transfer} extends this \gls{pl} constant to location-scatter families satisfying the same spectral bounds.

\noindent\textbf{Scope and Limitations.} Proposition~\ref{prop:pl_transfer} applies to location-scatter families, which are generated by affine transformations of a base measure. This includes elliptical distributions (e.g., multivariate $t$-distributions, uniform distributions on ellipsoids) and, importantly, the Swiss roll measures used in our experiments. This result does not directly apply to arbitrary measures.


\section{Domain Adaptation}\label{appx:da}


\subsection{Introduction}\label{appx:da-intro}

Statistical learning theory~\cite{vapnik1999overview} deals with the problem of learning a function $h:\mathcal{X}\rightarrow\mathcal{Y}$, such that it minimizes the \emph{risk} of making a mistake, that is,
\begin{align}
    h^{\star} = \arginf{h \in \mathcal{H}}\biggr\{ \mathcal{R}_{Q}(h) = \int_{\mathcal{X}\times\mathcal{Y}} \mathcal{L}(y,h(x))dQ(x,y) \biggr\},\label{eq:risk}
\end{align}
where $Q \in \mathcal{P}(\mathcal{X}\times\mathcal{Y})$ is a probability measure, and $\mathcal{H}$ is a family of functions from $\mathcal{X}$ to $\mathcal{Y}$. $\mathcal{R}_{Q}(h)$ denotes the risk of $h$ under $Q$. Equation~\ref{eq:risk} interprets the learning problem as a minimization problem.

In practical scenarios, one has a dataset $\{x_{i}^{(Q)}, y_{i}^{(Q)}\}_{i=1}^{n}$, $(x_{i}^{(Q)}, y_{i}^{(Q)}) \overset{i.i.d.}{\sim} Q$. As a result, one can estimate the \emph{empirical} risk, and consequently minimize it,
\begin{align}
    \hat{h} = \arginf{h \in \mathcal{H}}\biggr\{ \hat{\mathcal{R}}_{Q}(h) = \dfrac{1}{n}\sum_{i=1}^{n}\mathcal{L}(y_{i}^{(Q)}, h(x_{i}^{(Q)})) \biggr\},\label{eq:erm}
\end{align}
which is the foundation of the \emph{\gls{erm}} framework. In this sense, a solution $\hat{h}$ to Equation~\ref{eq:erm} is said to \emph{generalize} if it achieves low risk $\mathcal{R}_{Q}(h)$.

There is a fundamental assumption at play in the theory of generalization under the \gls{erm} framework. Indeed, both $\mathcal{R}_{P}$ and $\hat{\mathcal{R}}_{P}$ evaluate the risk under a fixed probability measure $P$. However, machine learning models are often confronted with data following \emph{different but related} probability measures~\cite{quinonero2022dataset}, a process known as \emph{distribution shift}.

The main tool for tackling distribution shift is \emph{domain adaptation}. This problem now considers two datasets, namely, a labeled source dataset $\{x_{i}^{(Q_{S})}, y_{i}^{(Q_{S})}\}_{i=1}^{n_{S}}$, and an \emph{unlabeled} target dataset $\{x_{j}^{(Q_{T})}\}_{j=1}^{n_{T}}$. The goal is to obtain a function $h$ that achieves low risk under $Q$, with the available data. A natural extension to this problem is \emph{multi-source} domain adaptation, where \emph{multiple source domains} are available, i.e.,  $\mathcal{Q} = \{Q_{k}\}_{k=1}^{K}$, each with its own associated dataset $\{ (x_{i}^{(Q_{k})}, y_{i}^{(Q_{k})}) \}_{i=1}^{n_{k}}$.

\subsection{Source-Only Training}\label{appx:da-src-only}

A straightforward baseline in domain adaptation is the \emph{source-only} training. Let $h = g\circ f$ be the composition of a feature extractor $g:\mathcal{X}\rightarrow \mathcal{U}$ and a feature classifier $f:\mathcal{U}\rightarrow \mathcal{Y}$. We can train $h$ \emph{end-to-end} by minimizing,
\begin{align*}
    \mathcal{L}(\theta) = \dfrac{1}{K}\sum_{k=1}^{K}\dfrac{1}{n_{k}}\sum_{i=1}^{n_{k}}\mathcal{L}(y_{i}^{(Q_{k})},f(g(x_{i}^{(Q_{k})}))).
\end{align*}
This training produces an embedding space $\mathcal{U}$ that is discriminative for the classes in the domain adaptation problem. Following previous work~\cite{montesuma2021wasserstein,montesuma2023multi,montesuma2024distillation}, we perform adaptation in this space.

\subsection{Wasserstein Barycenter Transport}\label{appx:wbt}

The motivation for using Wasserstein barycenters in multi-source domain adaptation comes from a theoretical result of~\cite{redko2019optimal}. Indeed, by~\cite[Theorem 2]{redko2019optimal},
\begin{align}
    \mathcal{R}_{Q_{T}}(h) \leq \mathcal{R}_{Q_{S}}(h) + \mathbb{W}_{1}(\hat{Q}_{S}, \hat{Q}_{T}) + \mathcal{O}(n_{S}^{-1/2} + n_{T}^{-1/2}) + \mathbb{R}(Q_{S},Q_{T}),\label{eq:redko-bound}
\end{align}
where $\mathbb{R}(Q_{S}, Q_{T}) = \text{inf}_{h \in \mathcal{H}} \mathcal{R}_{Q_{S}}(h) + \mathcal{R}_{Q_{T}}(h)$ is the joint risk. This result establishes that when source and target measures are close, their risks are close as well. Now, assuming Equation~\ref{eq:redko-bound} holds for each pair $(Q_{k}, Q_{T})$, summing over $k$ and weighting by $\lambda_{k}$ yields
\begin{align*}
    \mathcal{R}_{Q_{T}}(h) \leq \sum_{k=1}^{K}\lambda_{k}\mathcal{R}_{Q_{k}}(h) + \sum_{k=1}^{K}\lambda_{k}\mathbb{W}_{1}(\hat{Q}_{k}, \hat{Q}_{T}) + \mathcal{O}(n^{-1/2}) + \sum_{k=1}^{K}\lambda_{k}\mathbb{R}(Q_{k}, Q_{T}).
\end{align*}
Using the triangle inequality on $\mathbb{W}_{1}(\hat{Q}_{k}, \hat{Q}_{T})$,
\begin{align*}
    \mathcal{R}_{Q_{T}}(h) \leq \sum_{k=1}^{K}\lambda_{k}\mathcal{R}_{Q_{k}}(h) + \sum_{k=1}^{K}\lambda_{k}\mathbb{W}_{1}(\hat{Q}_{k}, P) + \mathbb{W}_{1}(P, \hat{Q}_{T}) + \mathcal{O}(n^{-1/2}) + \sum_{k=1}^{K}\lambda_{k}\mathbb{R}(Q_{k}, Q_{T}).
\end{align*}
A reasonable choice to make this bound tight is to choose a $P$ that minimizes $P \mapsto \sum_{k=1}^{K}\lambda_{k}\mathbb{W}_{1}(\hat{Q}_{k}, P) + \mathbb{W}_{1}(P, \hat{Q}_{T})$. 

\noindent\textbf{From $p=1$ to $p > 1$.} The argument for $p > 1$ is a bit more intricate. Since $\mathbb{W}_{1}$ is the weakest of the Wasserstein distances~\cite[Remark 6.6]{villani2008optimal}, we can directly move from $p=1$ to $p > 1$,
\begin{align*}
    \mathcal{R}_{Q_{T}}(h) \leq \sum_{k=1}^{K}\lambda_{k}\mathcal{R}_{Q_{k}}(h) + \sum_{k=1}^{K}\lambda_{k}\mathbb{W}_{p}(\hat{Q}_{k}, \hat{Q}_{T}) + \mathcal{O}(n^{-1/2}) + \sum_{k=1}^{K}\lambda_{k}\mathbb{R}(Q_{k}, Q_{T}),
\end{align*}
which, from the triangle inequality, results in,
\begin{align}
    \mathcal{R}_{Q_{T}}(h) \leq \sum_{k=1}^{K}\lambda_{k}\mathcal{R}_{Q_{k}}(h) + \sum_{k=1}^{K}\lambda_{k}\mathbb{W}_{p}(\hat{Q}_{k}, P) + \mathbb{W}_{p}(P, \hat{Q}_{T}) + \mathcal{O}(n^{-1/2}) + \sum_{k=1}^{K}\lambda_{k}\mathbb{R}(Q_{k}, Q_{T}).\label{eq:pre-bound-wbt}
\end{align}
The term $\sum_{k=1}^{K}\lambda_{k}\mathbb{W}_{p}(\hat{Q}_{k}, P) + \mathbb{W}_{p}(P, \hat{Q}_{T})$ does not correspond to the $p-$Wasserstein barycenter due to the missing power of $p$. We can retrieve it through Young's inequality for products. Let $a, b > 0$ be real numbers, and $p, q > 1$. Then $ab \le \frac{a^{p}}{p} + \frac{b^{q}}{q}.$ We can get the inequality,
\begin{align*}
    \mathbb{W}_{p}(\hat{P}, \hat{Q}) \leq \dfrac{\mathbb{W}_{p}(\hat{P}, \hat{Q})^{p}}{p} + \dfrac{p - 1}{p},
\end{align*}
by letting $a = \mathbb{W}_{p}(\hat{P}, \hat{Q})$, $b = 1$ and $q = \frac{p}{p-1}$. We apply this inequality to the $K+1$ terms in the previous bound. More precisely,
\begin{align*}
    \mathbb{W}_{p}(\hat{Q}_{k}, P) \leq \frac{1}{p}\mathbb{W}_{p}(\hat{Q}_{k}, P)^{p} + \frac{p-1}{p}\text{, and }\mathbb{W}_{p}(P, \hat{Q}_{T}) \leq \frac{1}{p}\mathbb{W}_{p}(P, \hat{Q}_{T})^{p} + \frac{p-1}{p}
\end{align*}

Plugging this back into inequality~\ref{eq:pre-bound-wbt},
\begin{align*}
    \mathcal{R}_{Q_{T}}(h) \leq \sum_{k=1}^{K}\lambda_{k}\mathcal{R}_{Q_{k}}(h) + &\dfrac{1}{p}\biggr(\sum_{k=1}^{K}\lambda_{k}\mathbb{W}_{p}(\hat{Q}_{k}, P)^{p} + \mathbb{W}_{p}(P, \hat{Q}_{T})^{p}\biggr) + \dfrac{2(p-1)}{p} + \mathcal{O}(n^{-1/2}) + \sum_{k=1}^{K}\lambda_{k}\mathbb{R}(Q_{k}, Q_{T}).
\end{align*}
Modulo some additional constants depending on $p > 1$, we have the same reasoning as for $p =1$.

\begin{wrapfigure}[12]{r}{0.45\textwidth} 
\resizebox{0.45\textwidth}{!}{%
\begin{minipage}{\linewidth} 
\begin{algorithm}[H]
\caption{Wasserstein Barycenter Transport}\label{alg:wbt}
\begin{algorithmic}[1]
\Require $\mathcal{Q} = 
\{Q_{k}\}_{k=1}^{K}$, $Q_{k} \in \mathcal{P}_{2}(\mathcal{X} \times\mathcal{Y})$, $Q_{T} \in \mathcal{P}_{2}(\mathcal{X})$, $\lambda \in \Delta_{K}$
\Ensure Labeled target domain data $\{ T_{P\rightarrow Q_{T}}(x_{i}^{(P)}), y_{i}^{(P)} \}_{i=1}^{(P)}$
\For{$\tau \gets 1$ to $n_{\text{iter}}$}
    \State Compute $P^{\star} = \text{Bar}(\lambda, \mathcal{Q})$
    \State Compute $\gamma^{\star}$ between $P^{\star}$ and $Q_{T}$ (eq.~\ref{eq:__ot})
    \State Transport $(x_{i}^{(P)}, y_{i}^{(P)})$ (eq.~\ref{eq:__barymap})
\EndFor
\end{algorithmic}
\end{algorithm}
\end{minipage}%
}
\end{wrapfigure}
\noindent\textbf{Wasserstein Barycenter Transport $(p=2)$.}~\cite{montesuma2021wasserstein} uses a two-step procedure for finding $P$. First, one solves,
\begin{align*}
    P^{\star} = \argmin{P \in \mathcal{P}_{1,\text{ac}}(\Omega)}\sum_{k=1}^{K}\lambda_{k}\mathbb{W}_{2}(Q_{k},P)^{2},
\end{align*}
which is the $2-$Wasserstein barycenter problem. Then, one aligns the barycenter with the target domain through \gls{ot}. This is done through the barycenter mapping,
\begin{align}
    \gamma^{\star} &= \argmin{\gamma\in\Gamma(P,\hat{Q}_{T})}\sum_{i=1}^{n}\sum_{j=1}^{n_{T}}\gamma_{ij}\lVert x_{i}^{(P)} - x_{j}^{(Q_{T})} \rVert_{2}^{2},\label{eq:__ot}
\end{align}
and,
\begin{align}
    T_{P\rightarrow Q_{T}}(x_{i}^{(P)}) &= n_{T}\sum_{j=1}^{n_{T}}\gamma_{ij}^{\star}x_{j}^{(Q_{T})},\label{eq:__barymap}
\end{align}
which is the minimizer of $P\mapsto \mathbb{W}_{2}(P, \hat{Q}_{T})^{2}$. We provide an algorithm for this strategy in Algorithm~\ref{alg:wbt}. Note that since this procedure generates a labeled dataset consisting of data that follow the target domain measure, we can train a classifier with the pairs $\{ T_{P\rightarrow Q_{T}}(x_{i}^{(P)}), y_{i}^{(P)} \}$.

\subsection{Adapting Neural Network Barycenter Solvers to DA}\label{appx:da-unsup-wbt}

Here, we cover \emph{how} we adapted previous neural network-based barycenters to perform domain adaptation. There is one main challenge with using unlabeled barycenters for domain adaptation: we cannot directly move the barycenter to the target domain, as we would end up with more unlabeled data. A straightforward solution to this problem is to move the available labeled data that we have, that is, the source domain data. For this scenario, the barycenter domain works as a \emph{pivot domain}, as we transport the source to the target through the composition,
\begin{align}
    T_{Q_{k}\rightarrow Q_{T}}(x_{i}^{(Q_{k})}) = T_{P\rightarrow Q_{T}}(T_{Q_{k}\rightarrow P}(x_{i}^{(Q_{k})})).\label{eq:composition-barymap}
\end{align}
In comparison, by computing a \emph{labeled barycenter}, we synthesize data in $P^{\star}$.

As we demonstrated in Tables~\ref{tab:office31} through~\ref{tab:isruc}, this is generally beneficial for domain adaptation. In this way, we can acquire a labeled dataset in the target domain through $\{\{T_{P\rightarrow Q_{T}}(T_{Q_{k} \rightarrow P}(x_{i}^{(Q_{k})})), y_{i}^{(Q_{k})}  \}_{i=1}^{n_{k}}\}_{k=1}^{K}$. To isolate the effect of the quality of the computed barycenters, we estimate the mappings as we did in Equations~\ref{eq:__ot} and~\ref{eq:__barymap}, that is,
\begin{align}
    T_{Q_{k} \rightarrow P}(x_{i}^{(Q_{k})}) = n\sum_{j=1}^{n}\gamma_{k,i,j}^{\star}x_{j}^{(P)} \quad T_{P\rightarrow Q_{T}}(x_{i}^{(P)}) = n_{T}\sum_{j=1}^{n_{T}}\gamma_{ij}^{\star}x_{j}^{(Q_{T})},\label{eq:composition-barymap-detailed}
\end{align}
where $\gamma_{k}^{\star}$ is the \gls{ot} plan from $Q_{k}$ to $P$, and $\gamma^{\star}$ is the \gls{ot} plan from $P$ to $Q_{T}$.


\section{Additional Experiments}\label{appx:additional-details-da}

In total, we compare five domain adaptation benchmarks, divided into computer vision, neuroscience, and chemical engineering. An overview of the benchmarks is presented in Table 1 in our main paper. Here, we give further details for each of the benchmarks. For \textbf{computer vision}, we use the Office 31~\cite{saenko2010adapting} and Office Home~\cite{venkateswara2017deep} benchmarks. For neuroscience, we use the BCI-CIV-2a~\cite{brunner2008bci} and ISRUC~\cite{khalighi2013automatic} benchmarks. For chemical engineering, we use the TEP benchmark~\cite{montesuma2024benchmarking}.

\subsection{Evaluation Protocol and Backbones}\label{appx:details-running}

As we describe below, all benchmarks deal with multi-class classification. We measure performance using the standard classification accuracy,
\begin{align*}
    \text{accuracy}(y, \hat{y}) = \dfrac{1}{n}\sum_{i=1}^{n}\delta(y_{i} - \hat{y}_{i})\times 100\%.
\end{align*}
For the Office 31, Office Home, Tennessee Eastman Process, and BCI-CIV-2a, we adopt the so-called \emph{leave-one-domain-out} evaluation. For example, given three domains $\{A, B, C\}$, we perform $3$ experiments, leaving one domain as the target domain at each time. Then, one of these experiments would be $\{A, B\} \rightarrow C$, where $\{A, B\}$ are the sources, and $C$ is the target.

As we discuss in the main paper, we perform adaptation over embeddings. To that end, we use the same experimental setting as~\cite{montesuma2024lighter} for the Office 31, Office Home, and TEP benchmarks. More specifically, we train a ResNet 50 and ResNet 101 for the Office 31 and Office Home benchmarks, and a CNN for the TEP benchmark. For this latter backbone, we refer readers to~\cite{montesuma2024benchmarking} for further details. For neuroscience benchmarks, we follow the experimental protocol of~\cite{wang2024cbramod}, where we fine-tune CBraMod, a transformer-based foundation model, on available source domain data. We use the exact same training settings as these authors.

\subsection{Fine-Grained Results}\label{appx:fine-grained-results}

\begin{wrapfigure}{r}{0.45\linewidth}
    \centering
    \resizebox{\linewidth}{!}{\begin{tabular}{lcccc>{\columncolor[gray]{0.9}}c}
        \toprule
        Algorithm & $\mathcal{X} \times \mathcal{Y}$ & A & D & W & Avg. $\uparrow$ \\
        \midrule
        ResNet50 & - & 67.50 & 95.00 & 96.83 & 86.40\\
        \midrule
        WJDOT~\cite{turrisi2022multi} & - & 67.77 & 97.32 & 95.32 & 86.80 \\
        DaDiL-E~\cite{montesuma2023multi} & - & 70.55 & 100.00 & 98.83 & 89.79\\
        DaDiL-R~\cite{montesuma2023multi} & - & 70.90 & 100.00 & 98.83 & 89.91\\
        GMM-DaDiL~\cite{montesuma2024lighter} & - & 72.47 & 100.00 & 99.41 & 90.63 \\
        \midrule
        Discrete~\cite{cuturi2014fast} & \xmark & 70.55 & 96.43 & 96.49 & 87.82\\
        NormFlow~\cite{visentin2025computing} & \xmark & 65.67 & 97.32 & 94.73 & 85.91\\
        CW2B~\cite{korotin2021continuous} & \xmark & 69.16 & 94.64 & 95.32 & 86.37\\
        NOT~\cite{kolesov2024estimating} & \xmark & 69.33 & 92.85 & 96.49 & 86.22\\
        U-NOT~\cite{gazdieva2024robust} & \xmark & 68.29 & 97.32 & 95.32 & 86.97\\
        WGF (ours) & \xmark & 69.16 & 98.21 & 96.49 & 87.96\\
        \midrule
        Discrete~\cite{montesuma2023multi} & \cmark & 67.94 & 98.21 & 97.66 & 87.93\\
        GMM~\cite{montesuma2024lighter} & \cmark & 70.13 & 99.11 & 96.49 & 88.54 \\
        WGF (ours) & \cmark & 70.03 & 99.11 & 99.41 & 89.52\\
        \bottomrule
    \end{tabular}}
    \captionof{table}{Classification accuracy per domain for the Office 31 benchmark.}
    \label{tab:office31}
\end{wrapfigure}
\noindent\textbf{Office 31}~\cite{saenko2010adapting} contains, in total, 4652 images among 31 classes of objects. These images are divided into 3 different domains: \emph{Amazon}, \emph{dSLR}, and \emph{Webcam}. Images in the \emph{Amazon} domain represent products, usually with an homogeneous background, medium resolution, and studio lightning. There are 2817 images in this domain. The \emph{dSLR} domain contains high-resolution images obtained through a digital SLR camera in an office environment. There are 498 images in this domain. Finally, the \emph{Webcam} domain contains images of objects captured with a webcam, also in an office environment. There are 795 images in this domain. As analyzed in~\cite{ringwald2021adaptiope}, this benchmark is one of the most commonly used in the literature. Despite its moderate size, it presents challenges of its own. For instance, since the \emph{Amazon} domain images were collected automatically, they has label noise.

We show our results in the Office 31 benchmark in Table~\ref{tab:office31}. Overall, dictionary learning methods such as DaDiL~\cite{montesuma2023multi} or GMM-DaDiL~\cite{montesuma2024lighter} remain state-of-the-art for this benchmark. However, our gradient flow algorithm is able to improve previous methods, closing the gap between these two kinds of algorithms. In comparison with \emph{unsupervised barycenters}, using labels in the ground-cost ($\mathcal{X} \times \mathcal{Y}$ = \cmark) has a clear advantage.

\begin{wrapfigure}{r}{0.45\linewidth}
\centering
        \resizebox{\linewidth}{!}{\begin{tabular}{lccccc>{\columncolor[gray]{0.9}}c}
        \toprule
        Algorithm & $\mathcal{X} \times \mathcal{Y}$ & Ar & Cl & Pr & Rw & Avg. $\uparrow$ \\
        \midrule
        ResNet101 & - & 72.90 & 62.20 & 83.70 & 85.00 & 75.95\\
        \midrule
        WJDOT~\cite{turrisi2022multi} & - & 74.28 & 63.80 & 83.78 & 84.52 & 76.59\\
        DaDiL-E~\cite{montesuma2023multi} & - & 77.16 & 64.95 & 85.47 & 84.97 & 78.14\\
        DaDiL-R~\cite{montesuma2023multi} & - & 75.92 & 64.83 & 85.36 & 85.32 & 77.86\\
        GMM-DaDiL~\cite{montesuma2024lighter} & - & 77.16 & 66.21 & 86.15 & 85.32 & 78.81\\
        \midrule
        Discrete~\cite{cuturi2014fast} & \xmark & 76.13 & 63.91 & 83.11 & 85.55 & 77.17\\
        NormFlow~\cite{visentin2025computing} & \xmark & 77.16 & 62.54 & 83.10 & 84.63 & 76.86\\
        CW2B~\cite{korotin2021continuous} & \xmark & 74.69 & 63.46 & 82.54 & 85.09 & 76.44\\
        NOT~\cite{kolesov2024estimating} & \xmark & 72.22 & 63.57 & 82.77 & 82.91 & 75.36\\
        U-NOT~\cite{gazdieva2024robust} & \xmark & 75.30 & 64.94 & 82.88 & 84.28 & 76.85\\
        WGF (ours) & \xmark & 74.89 & 64.60 & 83.89 & 85.21 & 77.56\\
        \midrule
        Discrete~\cite{montesuma2023multi} & \cmark & 76.54 & 63.91 & 83.78 & 84.74 & 77.25\\
        GMM~\cite{montesuma2024lighter} & \cmark & 75.31 & 64.26 & 86.71 & 85.21 & 77.87\\
        WGF (ours) & \cmark & 76.13 & 65.52 & 85.81 & 86.24 & 78.42\\
        \bottomrule
    \end{tabular}}
        \captionof{table}{Classification accuracy per domain for the Office-Home benchmark.}
\label{tab:office_home}
\end{wrapfigure}
\noindent\textbf{Office-Home}~\cite{venkateswara2017deep} contains 15,500 images among 65 classes of objects. There are four domains: \emph{Art}, \emph{Clipart}, \emph{Product}, and \emph{Real-World}. The \emph{Art} domain contains artistic depictions of objects, such as sketches, paintings, and ornamentation. It has 2427 images. The \emph{Clipart} domain is a collection of graphic art representing the objects. This domain has 4365 images. \emph{Product} contains images of objects without a background. It has 4439 images. \emph{Real-World} contains images of objects captured with a camera. It contains 4357 images. In this sense, the \emph{Product} domain is similar to the \emph{Amazon} domain in the Office 31 benchmark, and the \emph{Real-World} is similar to \emph{dSLR} or \emph{Webcam}. In comparison with the \emph{Office31} benchmark, the \emph{Office-Home} benchmark adds another factor of domain variation: style (e.g., \emph{Art} vs. \emph{Real-World}). We show our results in Table~\ref{tab:office_home}. In comparison with the \emph{Office 31} benchmark, our methods get even closer to the state-of-the-art. Again, using labels in the ground cost ($\mathcal{X} \times \mathcal{Y}$ = \cmark) offers a clear advantage.

\begin{remark}
    We follow previous works on visual domain adaptation, especially~\cite{saenko2010adapting} and~\cite{venkateswara2017deep}, and report the classification accuracy on fixed partitions of the target domains. For that reason, Tables~\ref{tab:office31} and~\ref{tab:office_home} do not have confidence intervals. While it would be possible to run these experiments with random partitions (e.g., $k-$fold cross validation), this would mean that our results would not be comparable to previous established results in~\cite{montesuma2024lighter}, for instance.
\end{remark}

\noindent\textbf{Tennessee-Eastman Process (TEP)}~\citep{reinartz2021extended,montesuma2024benchmarking} is a benchmark in chemical engineering. This benchmark comprises a set of simulations of a chemical reaction that occurs inside a reactor. In total, 34 variables are measured over time. As a result, we have a set of time series representing how the reaction progresses over each simulation. In some of these simulations, one of 28 possible faults can be introduced. The goal is, based on the readings, to determine \emph{which kind of fault}, or its absence, has occurred. As a result, there are 29 classes. Each simulation is also associated with a production mode, which drastically changes the statistical properties of the data generation process, thus introducing a distribution shift. As a result, each domain corresponds to one of these modes of operation. We refer readers to~\cite{montesuma2024benchmarking} and~\cite{reinartz2021extended} for more details.

\begin{table}[ht]
    \centering
    \resizebox{\linewidth}{!}{\begin{tabular}{lccccccc>{\columncolor[gray]{0.9}}c}
    \toprule
    Algorithm & $\mathcal{X} \times \mathcal{Y}$ & {Mode 1} & {Mode 2} & {Mode 3} & {Mode 4} & {Mode 5} & {Mode 6} & {Avg. $\uparrow$} \\
    \midrule
    CNN$^{\dag}$ & - & 80.82 $\pm$ 0.96 & 63.69 $\pm$ 1.71 & 87.47 $\pm$ 0.99 & 79.96 $\pm$ 1.07 & 74.44 $\pm$ 1.52 & 84.53 $\pm$ 1.12 & 78.48\\
    \midrule
    WJDOT~\cite{turrisi2022multi} & - & 89.06 $\pm$ 1.34 & 75.60 $\pm$ 1.84 & {89.99 $\pm$ 0.86} & {89.38 $\pm$ 0.77} & 85.32 $\pm$ 1.29 & 87.43 $\pm$ 1.23 & 86.13\\
    DaDiL-E$^{\ddag}$~\cite{montesuma2023multi} & - & 90.45 $\pm$ 1.02 & {77.08 $\pm$ 1.21} & 86.79 $\pm$ 2.14 & 89.01 $\pm$ 1.35 & 84.04 $\pm$ 3.16 & 87.85 $\pm$ 1.06 &  85.87\\
    DaDiL-R$^{\ddag}$~\cite{montesuma2023multi} & - & 91.97 $\pm$ 1.22 & {77.15 $\pm$ 1.32} & 85.41 $\pm$ 1.69 & {89.39 $\pm$ 1.03} & 84.49 $\pm$ 1.95 & {88.44 $\pm$ 1.29} & {86.14}\\
    GMM-DaDiL~\cite{montesuma2024lighter} & - & 91.72 $\pm$ 1.41 & 76.41 $\pm$ 1.89 & {89.68 $\pm$ 1.49} & 89.18 $\pm$ 1.17 & {86.05 $\pm$ 1.46} & {88.02 $\pm$ 1.12} & {86.85}\\
    \midrule
    Discrete~\cite{cuturi2014fast} & \xmark & 90.38 $\pm$ 1.29 & 71.17 $\pm$ 1.26 & 85.27 $\pm$ 1.09 & 87.29 $\pm$ 0.99 & 83.83 $\pm$ 1.74 & 84.94 $\pm$ 1.87 & 83.81\\
    NormFlow~\cite{visentin2025computing} & \xmark & 87.06 $\pm$ 0.57 & 66.25 $\pm$ 1.11 & 88.58 $\pm$ 1.00 & 89.35 $\pm$ 0.34 & 80.99 $\pm$ 2.21 & 85.08 $\pm$ 1.45 & 82.89\\
    CW2B~\cite{korotin2021continuous} & \xmark & 91.72 $\pm$ 0.79 & 73.71 $\pm$ 1.42 & 88.37 $\pm$ 1.31 & 89.32 $\pm$ 1.14 & 84.66 $\pm$ 1.75 & 87.19 $\pm$ 1.24 & 85.83\\
    NOT~\cite{kolesov2024estimating} & \xmark & 90.31 $\pm$ 1.08 & 72.37 $\pm$ 0.94 & 87.99 $\pm$ 1.31 & 89.28 $\pm$ 0.81 & 84.35 $\pm$ 1.68 & 85.29 $\pm$ 2.02 & 84.93\\
    U-NOT~\cite{gazdieva2024robust} & \xmark & 90.82 $\pm$ 0.65 & 72.68 $\pm$ 1.58 & 88.65 $\pm$ 0.95 & 89.00 $\pm$ 1.28 & 85.22 $\pm$ 1.07 & 86.19 $\pm$ 1.70 & 85.43\\
    WGF (ours) & \xmark & 91.51 $\pm$ 1.32& 73.77 $\pm$ 1.24 & 87.44 $\pm$ 1.16 & 89.38 $\pm$ 0.62 & 84.87 $\pm$ 1.75 & 86.08 $\pm$ 2.23 & 85.51\\
    \midrule
    Discrete~\cite{montesuma2023multi} & \cmark & {92.38 $\pm$ 0.66} & 73.74 $\pm$ 1.07 & 88.89 $\pm$ 0.85 & {89.38 $\pm$ 1.26} & 85.53 $\pm$ 1.35 & 86.60 $\pm$ 1.63 & 86.09\\
    GMM~\cite{montesuma2024lighter} & \cmark & {92.23 $\pm$ 0.70} & 71.81 $\pm$ 1.78 & 84.72 $\pm$ 1.92 & 89.28 $\pm$ 1.55 & {87.51 $\pm$ 1.73} & 82.49 $\pm$ 1.81 & 84.67\\
    WGF (ours) & \cmark & 92.34 $\pm$ 1.05 & 76.10 $\pm$ 1.56 & 89.78 $\pm$ 0.90 & 89.32 $\pm$ 1.03 & 85.92 $\pm$ 1.35 & 87.74 $\pm$ 1.52 & 86.87\\
    \bottomrule
    \end{tabular}}
    \caption{Classification accuracy per domain on the Tennessee Eastman Process benchmark.}
    \label{tab:tep}
\end{table}

We report our results in Table~\ref{tab:tep}. For this benchmark, the gap between unlabeled ($\mathcal{X}\times\mathcal{Y}$ = \xmark) and labeled ($\mathcal{X}\times\mathcal{Y}$ = \cmark) barycenter methods is closer. For instance, CW2B~\cite{korotin2021continuous} achieves 85.83 for the average target domain performance versus 86.09 for the labeled discrete barycenter. Furthermore, it surpasses the \gls{gmm} method of~\cite{montesuma2024lighter}. In comparison, our methods establish a new state-of-the-art for barycenter-based domain adaptation in this benchmark, having similar performance to GMM-DaDiL~\cite{montesuma2024lighter}.

\begin{wrapfigure}{r}{0.6\linewidth}
\centering
\resizebox{\linewidth}{!}{\begin{tabular}{lcccccccccc>{\columncolor[gray]{0.9}}c}
\toprule
Algorithm & $\mathcal{X} \times \mathcal{Y}$ & 1 & 2 & 3 & 4 & 5 & 6 & 7 & 8 & 9 & Avg. $\uparrow$ \\
\midrule
CBraMod$^{\dag}$ & - & 52.22 & 40.45 & 64.23 & 40.41 & 47.74 & 48.34 & 53.64 & 50.52 & 55.03 & 50.30\\
\midrule
DaDiL-R$^{\ddag}$~\cite{montesuma2023multi} & - & 58.68 & 40.27 & 66.14 & 41.45 & 59.54 & 51.38 & 53.12 & 57.46 & 55.21 & 53.69\\
GMM-DaDiL~\cite{montesuma2024lighter} & - & 60.59 & 42.36 & 70.31 & 48.12 & 61.11 & 50.34 & 62.15 & 58.85 & 60.00 & 57.10\\
\midrule
Discrete~\cite{cuturi2014fast}& \xmark & 60.93 & 43.40 & 69.96 & 50.00 & 61.28 & 50.69 & 61.63 & 61.45 & 57.11 & 57.38\\
NormFlow~\cite{visentin2025computing} & \xmark & 57.11 & 43.22 & 67.53 & 47.71 & 60.24 & 51.04 & 56.25 & 60.76 & 55.20 & 55.45\\
CW2B~\cite{korotin2021continuous} & \xmark & 60.00 & 43.57 & 70.13 & 48.95 & 60.93 & 51.21 & 61.28 & 61.63 & 57.46 & 57.25\\
NOT~\cite{kolesov2024estimating} & \xmark & 61.28 & 43.22 & 70.31 & 47.91 & 60.59 & 51.21 & 61.46 & 61.80 & 57.29 & 57.23\\
U-NOT~\cite{gazdieva2024robust} & \xmark & 60.59 & 44.09 & 69.44 & 47.70 & 61.11 & 50.34 & 61.97 & 61.11 & 58.15 & 57.17\\
WGF (ours) & \xmark & 60.93 & 43.75 & 69.96 & 49.16 & 60.76 & 51.38 & 61.97 & 63.54 & 57.64 & 57.68\\
\midrule
Discrete~\cite{montesuma2023multi} & \cmark & 60.76 & 42.53 & 70.66 & 48.54 & 61.28 & 50.69 & 61.63 & 61.28 & 59.54 & 57.43\\
GMM~\cite{montesuma2024lighter} & \cmark & 60.59 & 42.53 & 69.79 & 47.08 & 61.45 & 50.34 & 62.32 & 59.54 & 59.72 & 57.04\\
WGF (ours) & \cmark & 60.93 & 46.18 & 68.92 & 50.21 & 62.32 & 50.69 & 61.63 & 61.11 & 57.11 & 57.68\\
\bottomrule
\end{tabular}}
\captionof{table}{Classification accuracy per domain on the BCI-CIV-2a benchmark.}
\label{tab:bci_civ}
\end{wrapfigure}
\noindent\textbf{BCI-CIV-2a}~\citep{brunner2008bci} contains \gls{eeg} data from nine subjects. In brief, an \gls{eeg} measures the electrical activity of the brain over time. As such, like the TEP benchmark, this benchmark has a collection of time series data. The BCI-CIV-2a data is concerned with motor imagery tasks, that is, imagining the movement of the left hand, right hand, both feet, and tongue. These correspond to classes 1, 2, 3, and 4. Therefore, this benchmark contains four classes. Our goal here is to perform \emph{cross-subject} adaptation; namely, we use data from a given set of source subjects, and try to predict on a target subject. Therefore, each domain is a subject, so there are nine domains in total. There are, on average, $560$ samples per subject, with a total of $5088$ samples.

We show our results in Table~\ref{tab:bci_civ}. For cross-subject studies, we highlight that, besides improving individual domain performance, it is desirable to not degrade over the source-only performance. This is indeed the case for almost all methods. For instance, DaDiL~\cite{montesuma2023multi} degrades performance of subjects 2 and 7 by a small margin. For this benchmark, the difference between unlabeled ($\mathcal{X}\times\mathcal{Y}$ = \xmark) and labeled ($\mathcal{X}\times\mathcal{Y}$ = \cmark) is not as marked as before. We hypothesize that this benchmark falls into the covariate shift hypothesis, namely, $Q_{S}(X) \neq Q_{T}(X)$, rather than the more general joint shift $Q_{S}(X, Y) \neq Q_{T}(X, Y)$.

\begin{wrapfigure}{r}{0.6\linewidth}
\centering
\resizebox{\linewidth}{!}{\begin{tabular}{lccccccccccc>{\columncolor[gray]{0.9}}c}
\toprule
Algorithm & $\mathcal{X} \times \mathcal{Y}$ & 1 & 2 & 3 & 4 & 5 & 6 & 7 & 8 & 9 & 10 & \scriptsize{Avg. $\uparrow$} \\
\midrule
CBraMod$^{\dag}$ & - & 78.97 & 80.35 & 70.46 & 85.59 & 80.69 & 72.85 & 77.00 & 73.29 & 83.41 & 68.95 & 76.63\\
\midrule
WJDOT~\cite{turrisi2022multi} & - & 80.41 & 78.25 & 65.69 & 82.85 & 81.74 & 77.74 & 76.33 & 68.75 & 86.95 & 70.81 & 76.95\\ 
DaDiL-E~\cite{montesuma2023multi} & - & 76.12 & 77.90 & 65.93 & 83.21 & 81.28 & 68.45 & 75.66 & 72.61 & 79.75 & 65.93 & 75.14\\
DaDiL-R~\cite{montesuma2023multi} & - & 78.06 & 78.13 & 66.97 & 84.64 & 81.51 & 72.02 & 76.33 & 72.84 & 80.85 & 67.55 & 75.89\\
GMM-DaDiL~\cite{montesuma2024lighter} & - & 76.63 & 78.95 & 66.51 & 82.73 & 77.44 & 73.81 & 77.00 & 71.70 & 83.05 & 66.86 & 75.47\\
\midrule
Discrete~\cite{cuturi2014fast} & \xmark & 78.77 & 80.58 & 69.76 & 85.83 & 79.06 & 73.45 & 78.33 & 72.50 & 84.14 & 68.48 & 77.09\\
NormFlow~\cite{visentin2025computing} & \xmark & 81.12 & 82.90 & 75.58 & 84.28 & 80.34 & 78.69 & 75.88 & 70.34 & 80.48 & 70.81 & 78.05\\
CW2B~\cite{korotin2021continuous} & \xmark & 79.08 & 78.25 & 69.07 & 86.54 & 78.48 & 70.35 & 76.88 & 70.90 & 82.80 & 66.04 & 75.84\\
NOT~\cite{kolesov2024estimating} & \xmark & 78.57 & 79.88 & 67.32 & 85.59 & 78.72 & 73.21 & 77.77 & 72.50 & 83.41 & 67.21 & 76.44\\
U-NOT~\cite{gazdieva2024robust} & \xmark & 80.20 & 83.48 & 73.72 & 87.50 & 76.51 & 72.62 & 76.66 & 70.22 & 78.17 & 71.51 & 77.06\\
WGF (ours) & \xmark & 79.69 & 81.86 & 70.58 & 86.66 & 78.14 & 77.74 & 78.77 & 72.16 & 85.97 & 70.00 & 78.16 \\
\midrule
Discrete~\cite{montesuma2023multi} & \cmark & 78.87 & 80.69 & 72.09 & 86.07 & 78.60 & 78.45 & 78.00 & 72.95 & 85.48 & 70.81 & 78.20\\
GMM~\cite{montesuma2024lighter} & \cmark & 75.20 & 75.46 & 69.88 & 80.71 & 78.60 & 71.90 & 76.33 & 68.41 & 78.17 & 68.14 & 74.28\\
WGF (ours) & \cmark & 80.10 & 80.81 & 76.62 & 85.95 & 76.86 & 80.35 & 76.55 & 77.72 & 84.87 & 73.95 & 80.02 \\
\bottomrule
\end{tabular}}
\captionof{table}{Classification accuracy per domain on the ISRUC benchmark.}
\label{tab:isruc}
\end{wrapfigure}
\noindent\textbf{ISRUC}~\cite{khalighi2013automatic} is a benchmark for EEG data for sleep staging. In brief, sleep staging is the classification of physiological data, such as EEG, recorded during sleep into different sleep stages. In this context, the ISRUC benchmark is the most large scale benchmark used in this paper. This study comprises 100 subjects, with a total of 89240 samples (approximately 900 samples per domain). The sleep staging problem in this benchmark has five classes: awake, non-rapid eye movement sleep (subdivided into N1, N2, N3), and rapid eye movement sleep. 

Due to the large number of subjects in this benchmark, we adopt a slightly different evaluation protocol. Instead of using the \emph{leave-one-domain-out} strategy, we fix the first $90$ subjects as \emph{training subjects} and evaluate the performance of adapting to the remaining $10$ subjects. This means that we compute Wasserstein barycenters of $K = 90$ input measures. Overall, our \gls{wgf} method has a clear advantage over all other methods, with unsupervised \gls{wgf} (78.16\% average target accuracy) being the best over unlabeled barycenters, and labeled barycenters as well (80.02\% average test accuracy).


\subsection{Complexity Analysis and Additional Running Time}\label{appx:complexity}

We start by analyzing the complexity of \emph{evaluating} the functionals used in our work. We recall that,
\begin{align*}
    \mathbb{V}(P) = \dfrac{1}{n}\sum_{i=1}^{n}V(z_{i}^{(P)}), \quad   \mathbb{U}(P) = \dfrac{1}{n^{2}}\sum_{i=1}^{n}\sum_{j=1}^{n}U(z_{i}^{(P)}, z_{j}^{(P)}),
\end{align*}
where, assuming that the complexity of evaluating $V$ and $U$ is $\mathcal{O}(1)$, the complexity of evaluating $\mathbb{V}$ is $\mathcal{O}(n)$ and for $\mathbb{U}$ is $\mathcal{O}(n^{2})$. The complexity of $\mathbb{B}$ is usually the bottleneck, as it involves computing $K$ \gls{ot} problems. For exact \gls{ot}, this means we need to solve $K$ \gls{ot} problems of size $m \times n$, where $m$ is the batch size and $n$ is the barycenter support size. These \gls{ot} problems need to be solved sequentially, as parallelizing $K$ independent linear programs is non-trivial. In practice, we use the network simplex solver of~\cite{bonneel2011displacement}, wrapped in Python by~\cite{flamary2021pot}. This algorithm has worst-case complexity of $\mathcal{O}(n^{3})$ but in practice, runs in $\mathcal{O}(n^{2})$. Assuming $n \gg m$, the complexity becomes
\begin{align*}
    \text{Exact barycenter functional complexity} = \mathcal{O}(K \times m \times n),
\end{align*}
which scales linearly with the barycenter support size $n$. Though this might appear contradictory to the usual $\mathcal{O}(n^{3})$ complexity of \gls{ot}, we highlight that we are keeping $m$ constant as we scale the barycenter support size $n$. The same analysis can be done for entropic \gls{ot},
\begin{align*}
    \text{Entropic barycenter functional complexity} = \mathcal{O}(K \times L \times m \times n),
\end{align*}
where, instead of a factor $m^{2}$, we have $L\times m$, where $L > 0$ is the number of Sinkhorn iterations. Both of these approaches scale linearly with the number of samples in the barycenter support. However, there are two important advantages of using entropic \gls{ot}. First, we can use GPU acceleration for computing the \gls{ot} variables. Second, as shown in Section~\ref{sec:minibatch}, we can vectorize the $K$ \gls{ot} problems. While this does not alleviate the $K$ term in the complexity, it does lead to further speedups. These factors explain the difference in running time between exact and entropic \gls{ot}.

\begin{figure}[ht]
    \centering
    \begin{subfigure}{0.32\linewidth}
        \includegraphics[width=\linewidth]{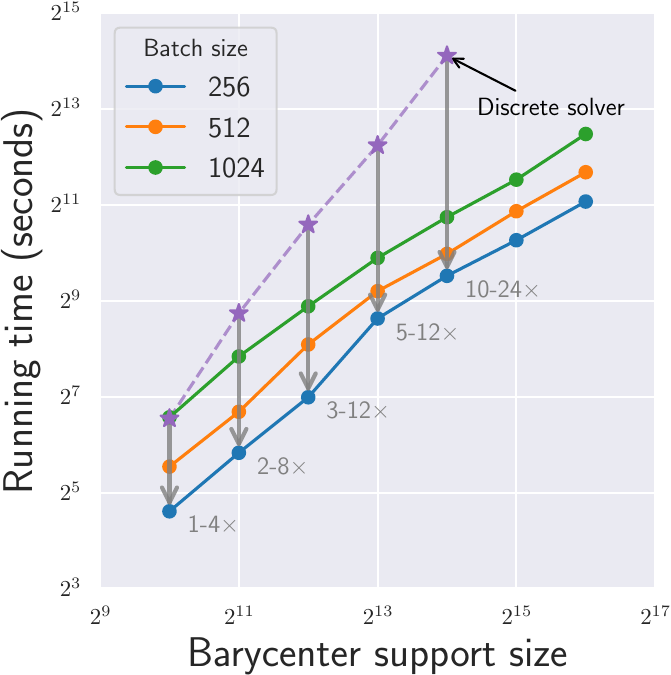}
        \caption{Scaling with $n$ ($\epsilon = 0$)}
    \end{subfigure}\hfill
    \begin{subfigure}{0.32\linewidth}
        \includegraphics[width=\linewidth]{Figures/EntropicOTTimings.pdf}
        \caption{Scaling with $n$ ($\epsilon > 0$)}
    \end{subfigure}\hfill
    \begin{subfigure}{0.32\linewidth}
        \includegraphics[width=\linewidth]{Figures/EntropicAndExact.pdf}
        \caption{Scaling with $\epsilon$ ($m = 256$)}
    \end{subfigure}
    \caption{Running time analysis in the Swiss roll example. In (a) and (b), we show that our \gls{wgf} approach improves the scaling of the discrete solver of~\cite{cuturi2014fast}. In (c), we isolate the speedups of GPU parallelism and Sinkhorn vectorization.}
    \label{fig:appendix-timings}
\end{figure}


Next, we perform a running time analysis of the tested methods on the ISRUC benchmark, which contains the largest number of samples and domains. For all neural net methods, we fix a batch size of $256$ samples. Overall, our proposed methods run significantly faster than previous methods. For instance, in comparison with the discrete solver of~\cite{cuturi2014fast}, our algorithm is approximately seven times faster \emph{per iteration}. It is noteworthy that some neural net methods are relatively faster per iteration (e.g., NormFlow). However, we empirically verified that these methods sometimes need a large number of iterations to converge to a good barycenter (e.g., 5000 iterations).

\begin{table}[ht]
    \centering
\resizebox{\linewidth}{!}{
    \begin{tabular}{cccccc}
        \toprule
         Method & Complexity per iteration & Parameter Complexity & Running time per iteration (s) & \# Iterations & Running Time (s) \\
         \midrule
         Discrete~\cite{cuturi2014fast,montesuma2023multi} & $K n^{3}\log n$ & $n$ & 4.28 & 50 & 214.37\\
         NormFlow~\cite{visentin2025computing} & N/A & $|f|$ & 0.17 & 5000 & 878.32\\
         CW2B~\cite{korotin2021continuous} & N/A & $K(|f| + |g|)$ & 2.18 & 1000 & 2181.00 \\
         NOT~\cite{kolesov2024estimating} & N/A & $K(|f| + |g|)$ & 41.63 & 25 &  1040.75 \\
         U-NOT~\cite{gazdieva2024robust} & N/A & $K(|f| + |g|)$ & 13.29 & 100 &  1329.77 \\
         \midrule
         WGF (ours, $\epsilon = 0$) & $Km^{2}n$ & $n$ & 0.05 & 200 & 10.43 \\
         WGF (ours, $\epsilon = 10^{-3}$) & $KLmn$ & $n$ & 0.03 & 200 & 7.00\\
         \bottomrule
    \end{tabular}
}
    \caption{Complexity and running time (in seconds) of different barycenter calculation strategies on the ISRUC benchmark. $n_{b}$ denotes the batch size, $n$ number of samples, $K$ number of input measures, $d$ number of dimensions. For NormFlow, $f$ denotes the neural net implementing the normalizing flow. For CW2B, NOT and U-NOT, $f$ and $g$ denote the neural networks approximating the Kantorovich potentials. $|f|$ and $|g|$ denote the number of their parameters.}
    \label{tab:Running time}
\end{table}

\subsection{Effect of Different Functionals}\label{appx:functional-effect}

\begin{figure}[ht]
    \centering
    \begin{subfigure}{0.16\linewidth}
        \includegraphics[width=\linewidth]{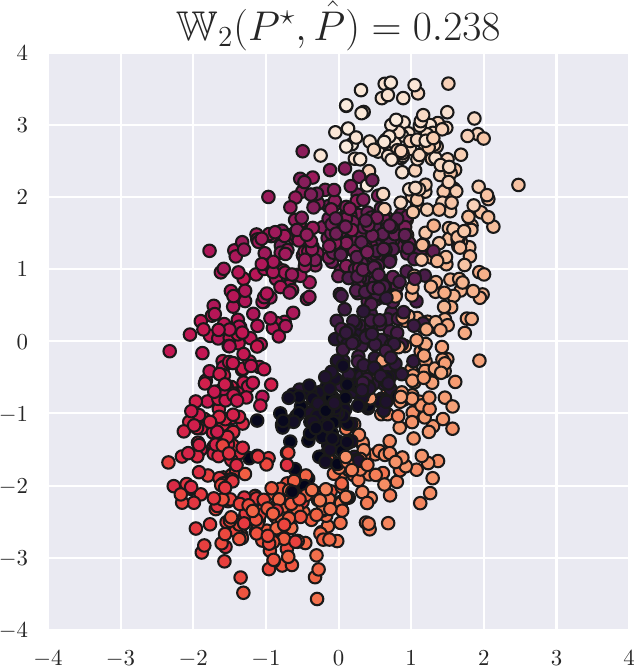}
        \caption{$(10^{-3}, 10^{-3})$}
    \end{subfigure}\hfill
    \begin{subfigure}{0.16\linewidth}
        \includegraphics[width=\linewidth]{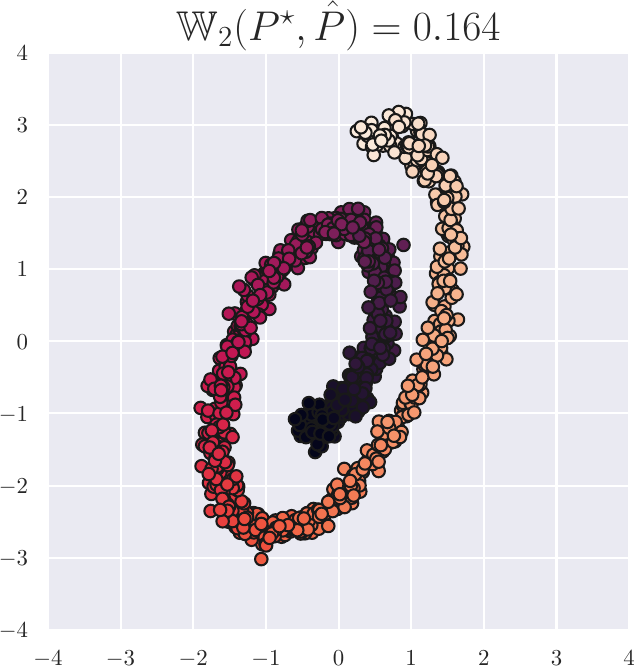}
        \caption{$(10^{-4}, 10^{-3})$}
    \end{subfigure}\hfill
    \begin{subfigure}{0.16\linewidth}
        \includegraphics[width=\linewidth]{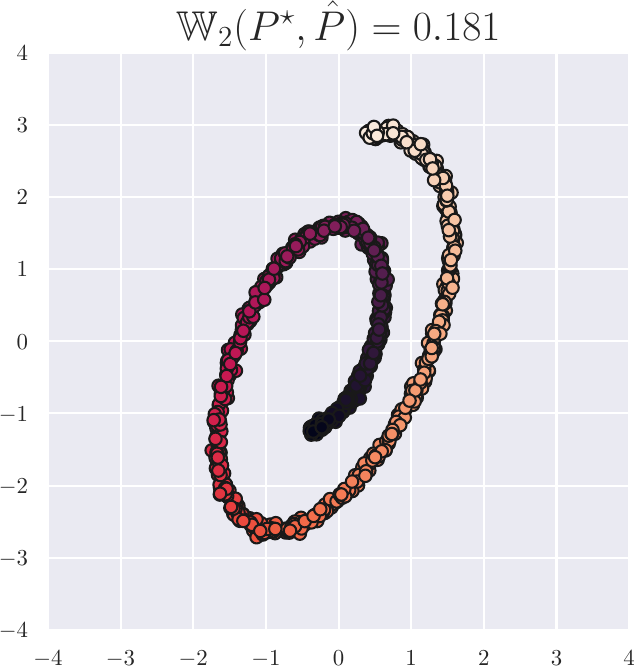}
        \caption{$(10^{-5}, 10^{-3})$}
    \end{subfigure}
    \begin{subfigure}{0.16\linewidth}
        \includegraphics[width=\linewidth]{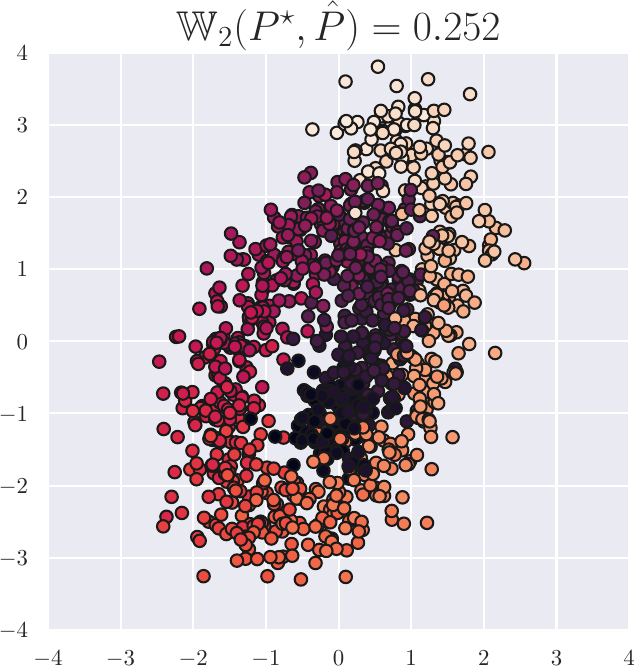}
        \caption{$(10^{-3}, 0.0)$}
    \end{subfigure}\hfill
    \begin{subfigure}{0.16\linewidth}
        \includegraphics[width=\linewidth]{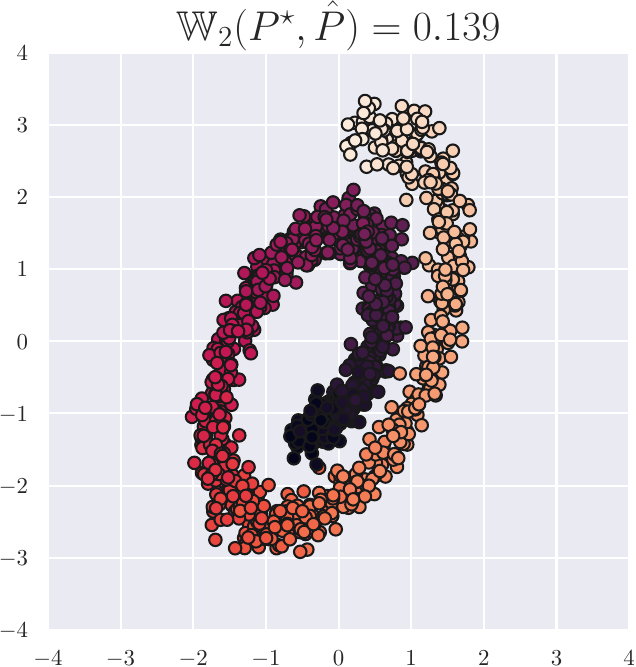}
        \caption{$(10^{-4}, 0.0)$}
    \end{subfigure}\hfill
    \begin{subfigure}{0.16\linewidth}
        \includegraphics[width=\linewidth]{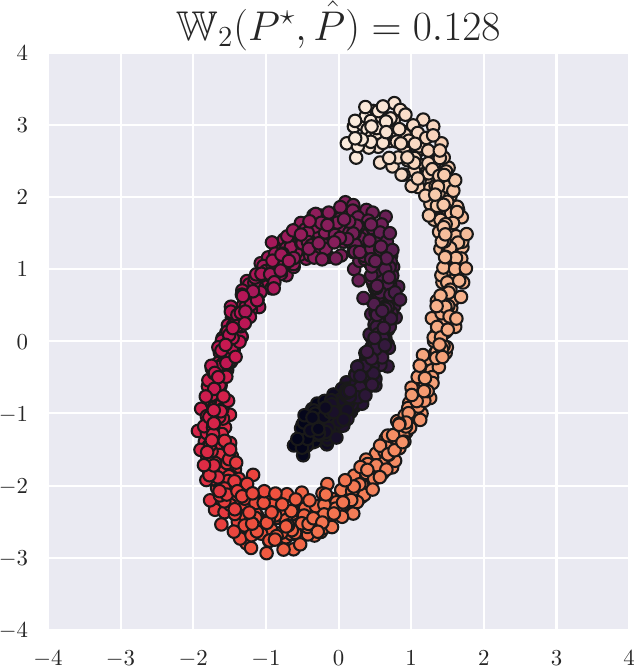}
        \caption{$(10^{-5}, 0.0)$}
    \end{subfigure}\hfill
    \caption{Wasserstein barycenter as a function of $(\eta_{G},\epsilon)$, i.e., diffusion coefficient $\xi$, and entropy regularization $\epsilon$.}
    \label{fig:comparison-barycenters}
\end{figure}

\noindent\textbf{On the effect of $\mathbb{G}$.} The internal energy functional~\citep{chizat2025doubly} controls the smoothness of $\mathbb{G}$, and acts, in the case of the entropy functional $G(s) = s\log s$, as a diffusion term. In Figure~\ref{fig:comparison-barycenters}, we show the effect of varying $\xi$ (cf. Equation~\ref{eq:langevin}) in the final barycenter. Overall, an aggressive diffusion coefficient leads to fuzzy barycenters (cf. Figure~\ref{fig:comparison-barycenters} (a)), while a coefficient that is too small leads to low variance measures (cf. Figure~\ref{fig:comparison-barycenters} (c)). This is, in fact, an artifact of entropic \gls{ot}. While a large diffusion coefficient does lead to fuzzy barycenters with exact \gls{ot} (cf. Figure~\ref{fig:comparison-barycenters} (d)), using a small coefficient does not degrade the quality of the computed barycenter.

\begin{figure}[h]
   \centering
   \begin{subfigure}{0.24\linewidth}
       \includegraphics[width=\linewidth]{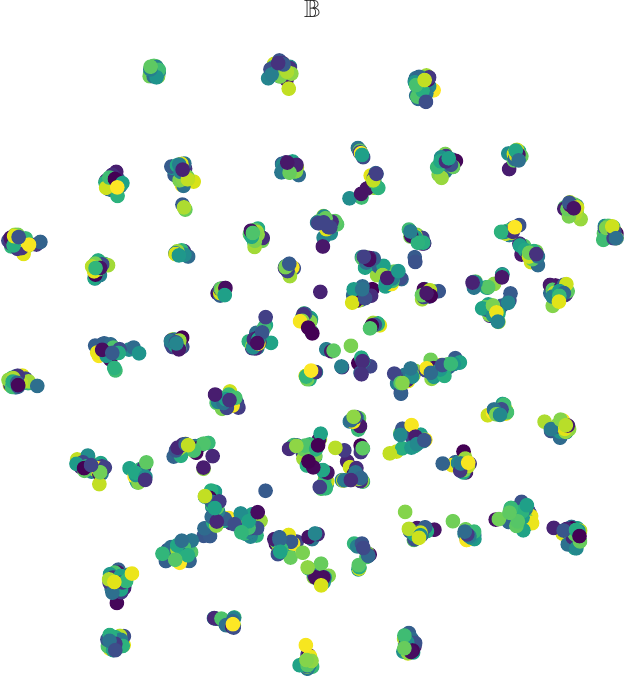}
       \caption{Unsupervised}
   \end{subfigure}\hfill
   \begin{subfigure}{0.24\linewidth}
       \includegraphics[width=\linewidth]{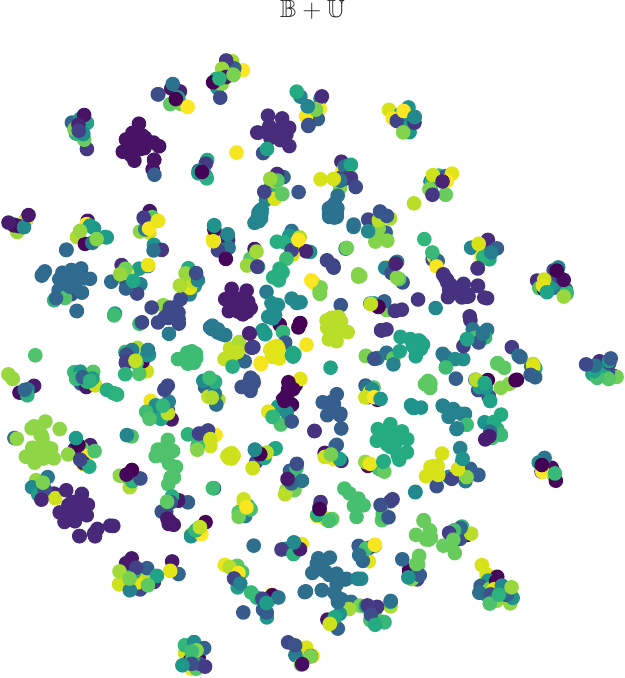}
       \caption{Unsupervised}
   \end{subfigure}\hfill 
   \begin{subfigure}{0.24\linewidth}
       \includegraphics[width=\linewidth]{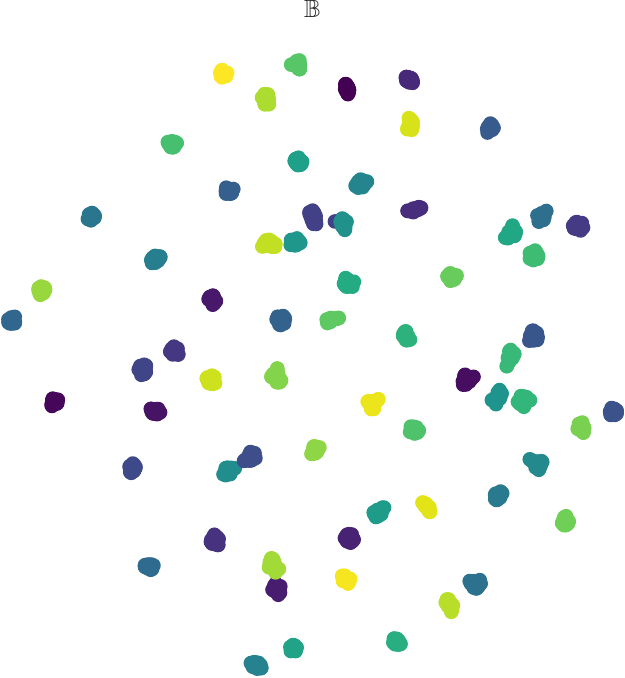}
       \caption{Supervised}
   \end{subfigure}\hfill
   \begin{subfigure}{0.24\linewidth}
       \includegraphics[width=\linewidth]{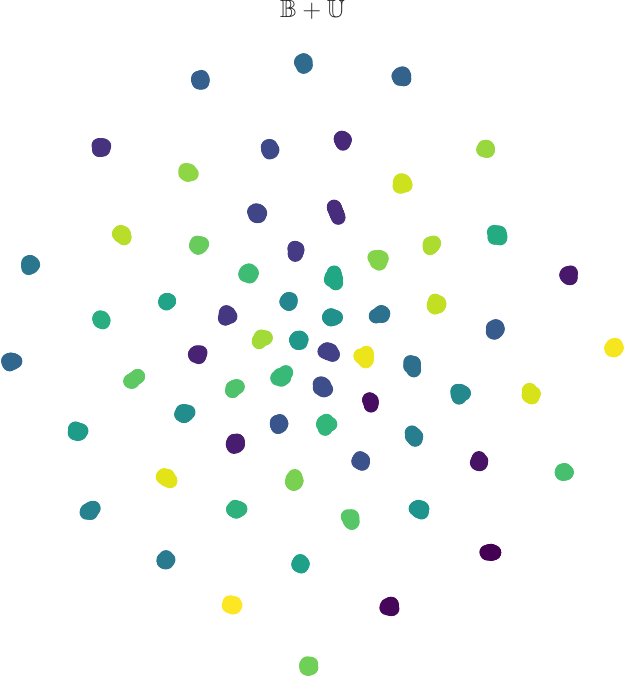}
       \caption{Supervised}
   \end{subfigure}\hfill
   \caption{t-SNE embeddings of unsupervised (a, b) and supervised (c, d) barycenters. While it is possible to group some samples according to their pseudo-labels based on $\mathbb{U}$, the task is in general difficult. For the supervised case, the interaction energy is able to make classes more separable (d, in comparison to c).}
   \label{fig:vis}
\end{figure}

\noindent\textbf{On the effect of $\mathbb{U}$.} We now visualize the effect of adding the repulsion interaction energy functional in the Office-Home benchmark. For computer vision experiments, we use ResNets~\citep{he2016deep} as the backbone, which produce $2048-$dimensional embeddings. For that reason, we use the cosine distance $d(x, x') = 1 - \text{cossim}(x, x')$ in Equation~\ref{eq:repulsion} and we fix $\text{margin} = 1$, which encourages class clusters to live in orthogonal subspaces of the latent space. Overall, adding $\mathbb{U}$ leads to well-separated classes.

\begin{wrapfigure}{r}{0.37\linewidth}
    \centering
    \includegraphics[width=\linewidth]{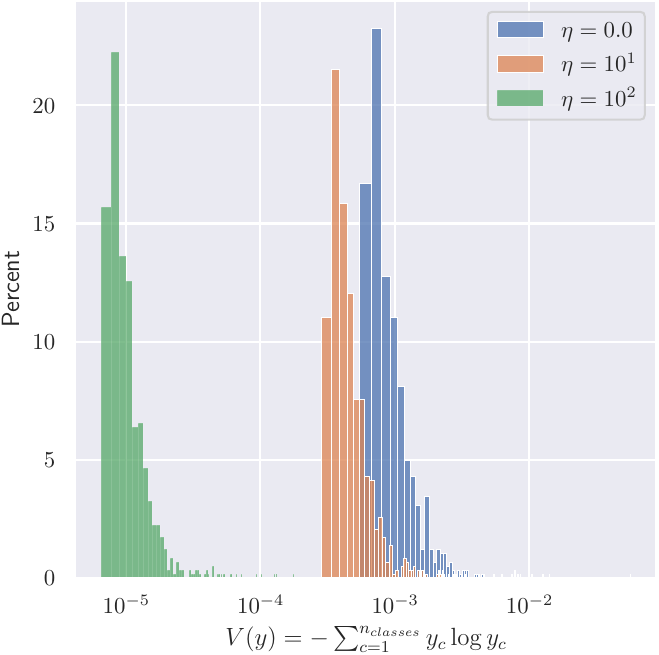}
    \caption{Distribution of label entropy values as a function of potential energy scaling $\eta$.}
    \label{fig:EntropyDistribution}
\end{wrapfigure}
In Figure~\ref{fig:vis}, we show a comparison between unsupervised and supervised regularized barycenters. In (a), unsupervised, unregularized barycenters do not have any label structure, since there is no label supervision. In (b), we show that by adding the interaction energy functional, we can cluster samples with the same assigned label --- even without input measure label supervision. In (c) and (d), we show the supervised case, in which labels in the input measures are available. In (c), classes are well clustered, but some of them overlap in the embedding space. Adding the interaction energy functional $\mathbb{U}$ effectively pushes these apart, as shown in (d). Overall, we conclude that \emph{enforcing well-separated class clusters is not straightforward without the inductive bias coming from the labels}.

\noindent\textbf{On the effect of $\mathbb{V}$.} In Figure~\ref{fig:EntropyDistribution}, we show the distribution of label entropy values (cf. Equation~\ref{eq:repulsion}) as a function of the potential scaling $\eta$,
\begin{equation*}
    \mathbb{V}(P) = \frac{\eta}{n}\sum_{i=1}^{n}V(y_{i}^{(P)}).
\end{equation*}
We see that increasing the weight of this functional leads to increasingly sharp labels.

\subsection{Hyperparameter Setting}\label{appx:hyperparam}

In this section, we detail the hyperparameters of our methods. For both flows, we have three main hyperparameters,
\begin{itemize}
    \item number of samples in the barycenter support, $n$,
    \item learning rate $\alpha$ for the gradient flow,
    \item number of iterations $n_{\text{iter}}$.
\end{itemize}
Additionally, for the empirical flow, we have the batch size $m$ for the input measures. We show the complete list of hyperparameters in Table~\ref{tab:hyper-parameters}. In addition to these hyperparameters, we also have the choice of regularizing functionals, $\mathbb{V}$ and $\mathbb{U}$. 

\begin{table}[ht]
    \centering
    \begin{tabular}{lccccccccccc}
        \toprule
        Benchmark & $n$ & $m$ & $\alpha$ & $n_{\text{iter}}$ & $\epsilon$ & $\mathbb{V}$ & $\eta_{V}$ & $\mathbb{U}$ & $\eta_{U}$ & margin & $\eta_{G}$\\
        \midrule
        Supervised $(\mathcal{X}\times\mathcal{Y})$\\
        Office31 & 620 & 620 & 1.0 & 100 & $10^{-3}$ & Entropy & $10^{2}$  & Repulsion & $10^{3}$ & $1.0$ & $10^{-2}$ \\
        Office-Home & 1300 & 325 & 1.0 & 100 & $10^{-3}$ & - & - & Repulsion & $10^{5}$ & 1.0 & $10^{-4}$ \\
        TEP & 870 & 145 & 1.0 & 50 & $0.0$ & Entropy & $10^{2}$ & Repulsion & $10^{3}$ & 1.0 & 0.0 \\
        BCI-CIV-2a & 40 & 80 & 1.0 & 50 & $10^{-2}$ & Entropy & $10^{2}$ & Repulsion & $10^{5}$ & 1.0 & 0.0\\
        ISRUC & 200 & 100 & 0.1 & 200 & 0.0 & - & - & - & - & - & -\\
        \midrule
        Unsupervised $(\mathcal{X})$\\
        Office31 & 620 & 465 & 1.0 & 100 & $10^{-2}$ & - & - & Repulsion & $10^{5}$ & 1.0 & $10^{-3}$\\
        Office-Home & 1950 & 650 & 1.0 & 100 & 0.0 & - & - & Repulsion & $10^{6}$ & 1.0 & $10^{-2}$\\
        TEP & 580 & 145 & 1.0 & 50 & 0.0 & - & - & - & - & - & -\\
        BCI-CIV-2a & 80 & 40 & 1.0 & 50 & $10^{-2}$ & - & - & - & - & - & -\\
        ISRUC & 250 & 50 & 0.1 & 200 & $10^{-3}$ & - & - & - & - & - & -\\
        \bottomrule
    \end{tabular} 
    \caption{Hyperparameters used in each benchmark. $n$ denotes number of samples, $m$ denotes the batch size, $\alpha$ denotes learning rate, $n_{\text{iter}}$ denotes number of iterations.}
    \label{tab:hyper-parameters}
\end{table}

We offer some additional reasoning on how to choose these parameters. The number of samples controls the complexity of the calculated barycenter. We express these values in terms of the number of classes. For instance, $n = 1550$ in the Office 31 benchmark translates to $n = 31 \times 50$ samples, that is, $50$ samples per class. The same reasoning applies to the batch size $m$. Furthermore, our method is fairly robust to the choice of learning rate. For instance, as we discussed in Section~\ref{appx:calculus}, the fixed-point iterations in the discrete barycenter are equivalent to setting $\alpha = n/2$, where $n$ is the number of samples in the Wasserstein barycenter. This generally strikes a balance on how fast the algorithm converges.

\subsection{Ablation}\label{sec:ablation}

Table~\ref{tab:ablation} ablates the effect of the functionals $\mathbb{G},\mathbb{V}, \mathbb{U}$ in the Office-Home benchmark. First, our \gls{wgf} consistently outperforms the discrete baseline (77.17 vs. 77.53 and 77.24 vs. 78.42 in the unlabeled and labeled cases). Second, $\mathbb{V}$ acts as a stabilizer with high entropic regularization ($\epsilon = 10^{-2}$). While the entropic regularizer blurs the barycentric measure, $\mathbb{V}$ has a sharpening effect. Third, our \gls{wgf} is more sensitive to the entropic regularization $\epsilon$, particularly when used as a loss function (our case) rather than for \gls{ot} plan calculations, as in~\cite{cuturi2014fast}. This latter point highlights the importance of using a sharpnening functional. We explore the effects of $\mathbb{G}$, $\mathbb{V}$, and $\mathbb{U}$ in further detail in Appendix~\ref{appx:functional-effect}.


\begin{table}[ht]
    \centering
\begin{tabular}{l c ccccccccc}
\toprule
Method & $\epsilon$ & $\mathbb{B}$ & $\mathbb{B}{+}\mathbb{U}$ & $\mathbb{B}{+}\mathbb{V}$ & $\mathbb{B}{+}\mathbb{V}{+}\mathbb{U}$ & $\mathbb{B}{+}\mathbb{G}$ & $\mathbb{B}{+}\mathbb{G}{+}\mathbb{V}$ & $\mathbb{B}{+}\mathbb{G}{+}\mathbb{U}$ & $\mathbb{B}{+}\mathbb{G}{+}\mathbb{V}{+}\mathbb{U}$ \\
\midrule
\multicolumn{10}{l}{\textit{Unlabeled setting ($\mathcal{X}$ only)}} \\
Discrete   & 0.0   & 77.17 & --- & --- & --- & --- & --- & --- & --- \\
Discrete   & 0.01  & 76.61 & --- & --- & --- & --- & --- & --- & --- \\
Discrete   & 0.001 & 76.56 & --- & --- & --- & --- & --- & --- & --- \\
WGF (ours) & 0.0   & 77.41 & 76.83 & 77.41 & 77.13 & 76.88 & 76.88 & 77.53 & 77.53\\
WGF (ours) & 0.01  & 75.54 & 74.42 & 75.26 & 75.26 & 74.55 & 74.55 & 75.00 & 75.00\\
WGF (ours) & 0.001 & 77.15 & 76.88 & 76.94 & 76.94 & 77.28 & 77.28 & 77.26 & 77.26\\
\midrule
\multicolumn{10}{l}{\textit{Labeled setting ($\mathcal{X} \times \mathcal{Y}$)}} \\
Discrete   & 0.0   & 77.24 & --- & --- & --- & --- & --- & --- & --- \\
Discrete   & 0.01  & 76.73 & --- & --- & --- & --- & --- & --- & --- \\
Discrete   & 0.001 & 76.64 & --- & --- & --- & --- & --- & --- & --- \\
WGF (ours) & 0.0   & 77.97 & 78.12 & 78.07 & 78.25 & 78.11 & 78.11 & 78.19 & 78.19\\
WGF (ours) & 0.01  & 77.59 & 77.87 & 77.83 & 77.83 & 77.83 & 77.88 & 77.89 & 77.94\\
WGF (ours) & 0.001 & 78.16 & 78.16 & 78.15 & 78.16 & 77.99 & 78.07 & \textbf{78.42} & 78.37\\
\bottomrule
\end{tabular}
    \caption{Ablation on the contribution of each energy functional on domain adaptation performance.}
    \label{tab:ablation}
\end{table}

\end{document}